\def\year{2022}\relax
\documentclass[letterpaper]{article} 
\usepackage{aaai22}  
\usepackage{times}  
\usepackage{helvet}  
\usepackage{courier}  
\usepackage[hyphens]{url}  
\usepackage{graphicx} 
\usepackage{changepage}
\usepackage{enumerate}
\usepackage{subcaption}
\usepackage{amsmath,amssymb,amsfonts,amsthm}
\usepackage{xcolor}
\usepackage{booktabs}
\usepackage{enumitem}
\usepackage{mathtools}
\usepackage{array}
\usepackage{multirow}
\usepackage{colortbl}
\usepackage{hyperref}

\usepackage{colortbl}
\definecolor{LightCyan}{rgb}{0.88,1,1}
\usepackage[titletoc]{appendix}
\usepackage{placeins}

\usepackage{natbib}  
\usepackage{caption} 
\DeclareCaptionStyle{ruled}{labelfont=normalfont,labelsep=colon,strut=off} 
\usepackage{algorithm}
\usepackage{algorithmic}

\usepackage{newfloat}
\usepackage{listings}
\floatstyle{ruled}
\newfloat{listing}{tb}{lst}{}
\floatname{listing}{Listing}
\nocopyright

\definecolor{LightCyan}{rgb}{0.88,1,1}

\DeclareMathOperator*{\argmax}{argmax}

\usepackage{stackengine}

\newcommand{\OSb}{$\mathcal{O_{\overline{S}}}$}
\newcommand{\OS}{$\mathcal{O_{S}}$}

\newcommand{\POS}{$\mathcal{P}(\mathcal{O_{S}})$}
\newcommand{\POSC}{$\mathcal{P}(\mathcal{O_{S}})_{\perp}$}

\newcommand{\ROCRSM}{$\mathcal{R_{OC}} \cup \mathcal{R_{SM}}$}

\title{Behind the Curtain: Learning Occluded Shapes for 3D Object Detection}
\author {
    Qiangeng Xu,
    Yiqi Zhong,
    Ulrich Neumann
}
\affiliations {
    University of Southern California\\
    qiangenx@usc.edu, yiqizhon@usc.edu, uneumann@usc.edu
}
\begin{document}

\makeatletter
\let\@oldmaketitle\@maketitle
\renewcommand{\@maketitle}{\@oldmaketitle

\begin{center}
    \begin{adjustwidth}{0pt}{0pt}
        \centering
        \includegraphics[width=0.95\textwidth]{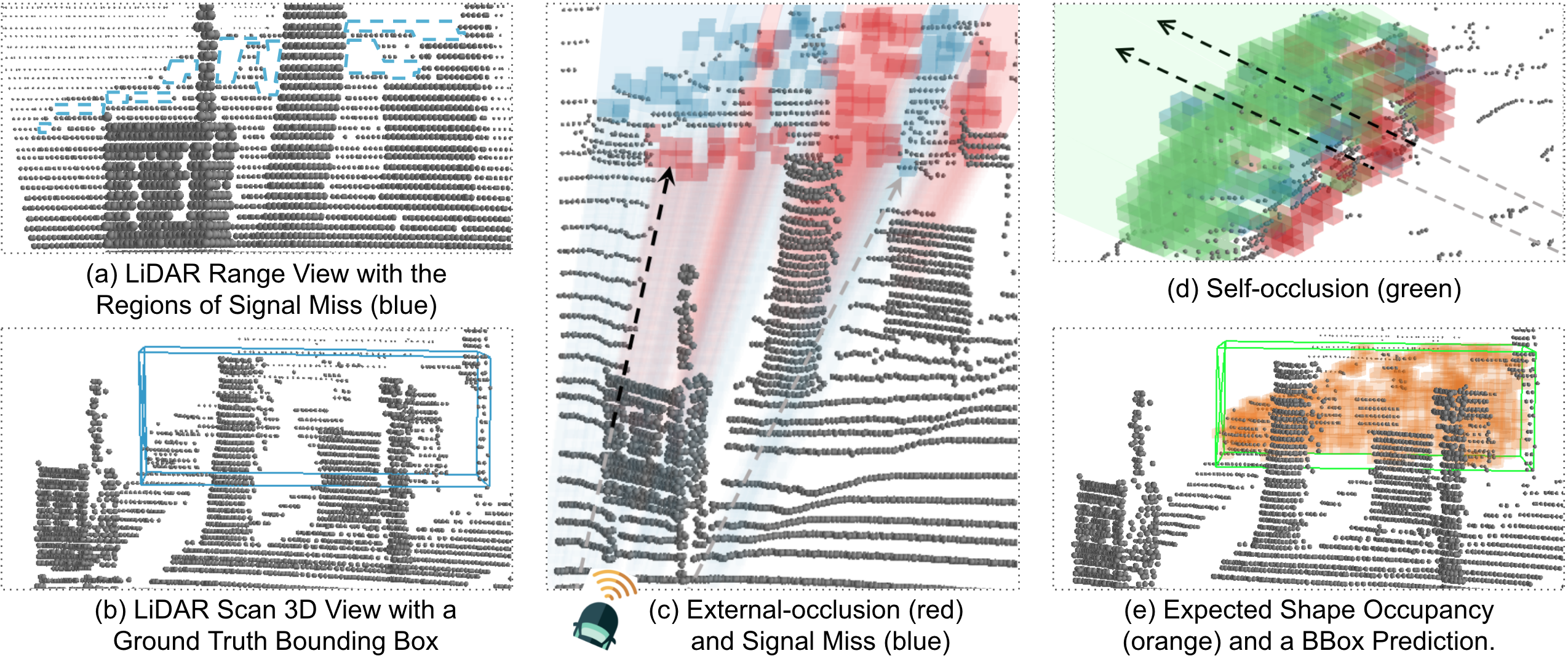}
        \captionsetup{aboveskip=5pt}
        \captionof{figure}{In a LiDAR scan (a) and (b), locating an object is difficult when its shape is largely missing. We discover three causes of shape miss: external-occlusion (red regions in (c)), signal miss (blue regions in (c)), and self-occlusion (green regions in (d)). BtcDet learns the occupancy probability of complete object shapes (e) and achieves the state-of-the-art detection performance.}
        \label{fig:teaser}
    \end{adjustwidth}
\end{center}
 }
\makeatother
\maketitle
    
\begin{abstract}
    Advances in LiDAR sensors provide rich 3D data that supports 3D scene understanding. However, due to occlusion and signal miss, LiDAR point clouds are in practice 2.5D as they cover only partial underlying shapes, which poses a fundamental challenge to 3D perception. To tackle the challenge, we present a novel LiDAR-based 3D object detection model, dubbed \textit{Behind the Curtain Detector (BtcDet)}, which learns the object shape priors and estimates the complete object shapes that are partially occluded (curtained) in point clouds. BtcDet first identifies the regions that are affected by occlusion and signal miss. In these regions, our model predicts the probability of occupancy that indicates if a region contains object shapes. Integrated with this probability map, BtcDet can generate high-quality 3D proposals. Finally, the probability of occupancy is also integrated into a proposal refinement module to generate the final bounding boxes. Extensive experiments on the KITTI Dataset \cite{geiger2013vision} and the Waymo Open Dataset \cite{sun2019scalability} demonstrate the effectiveness of BtcDet. Particularly, for the 3D detection of both cars and cyclists on the KITTI benchmark, BtcDet surpasses all of the published state-of-the-art methods by remarkable margins. Code is released  \footnote{\href{https://github.com/Xharlie/BtcDet}{https://github.com/Xharlie/BtcDet}}.
\end{abstract}
\section{Introduction}
    \label{sec:intro}
    \begin{figure*}[!hbt]
        \begin{adjustwidth}{-0pt}{0pt}
            \begin{subfigure}{0.45\linewidth}
                \flushleft
                \includegraphics[width=0.99\linewidth]{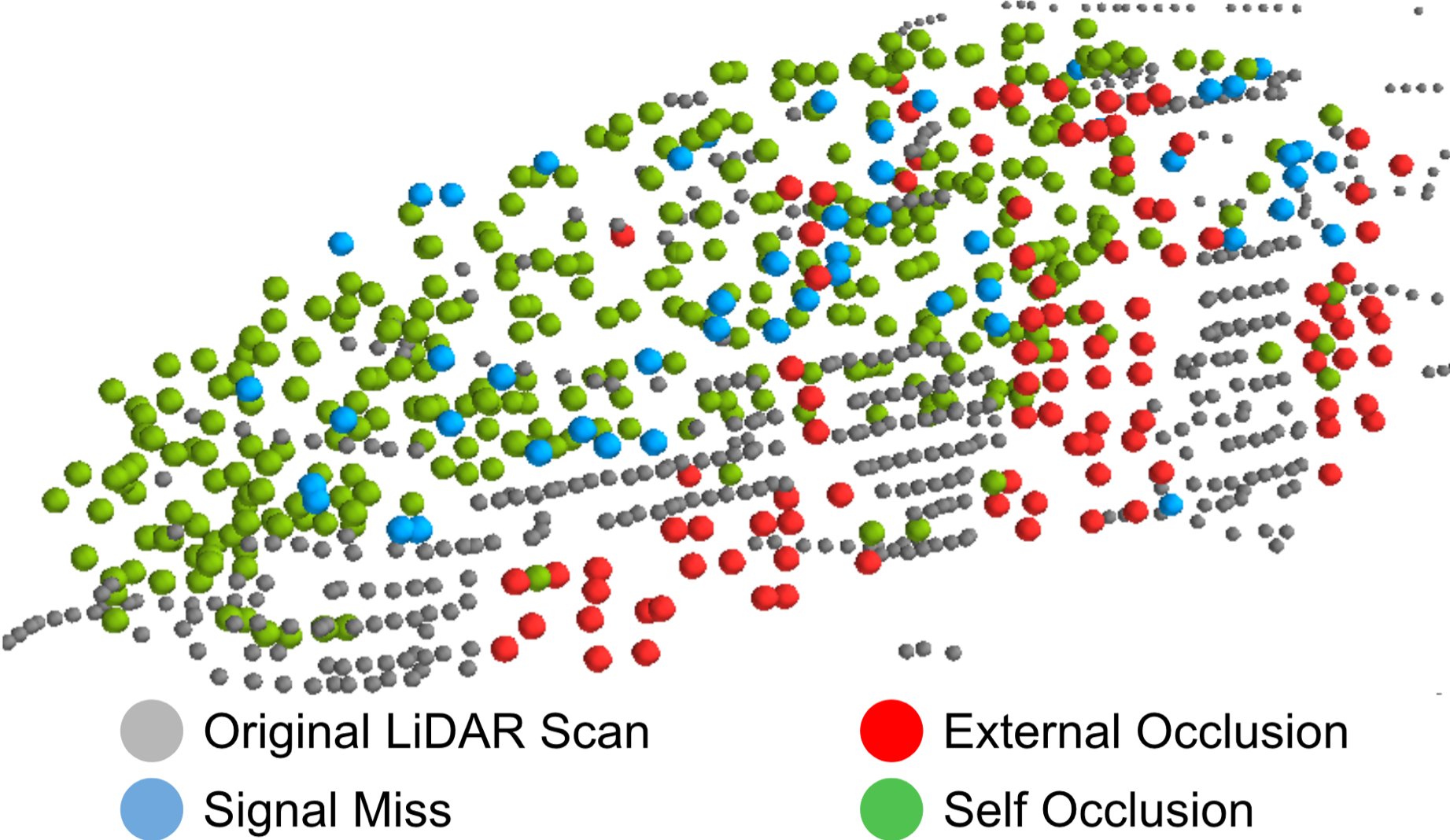}
                \caption{The points to recover different shape miss regions.}
            \end{subfigure} \hfill
            \begin{subfigure}{0.55\linewidth}
                \flushright
                \includegraphics[width=0.99\linewidth]{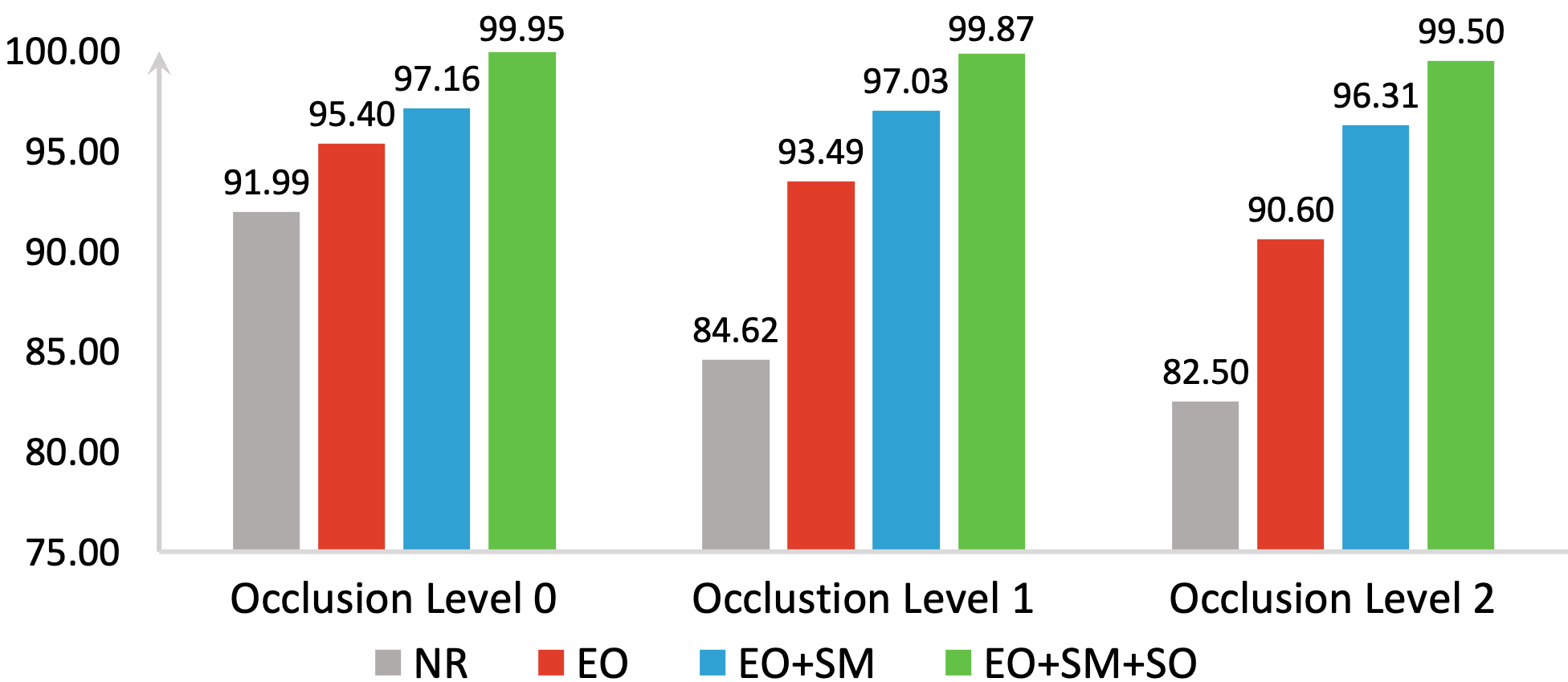} 
                \caption{The 3D Average Precisions with shape miss recovery.}
            \end{subfigure}
        \end{adjustwidth}
        \caption{The impact of the three types of shape miss. (b) shows PV-RCNN's \cite{shi2020pv} car 3D detection APs with different occlusion levels on the KITTI \cite{geiger2013vision} \textit{val} split. NR means no shape miss recovery. EO, SM, and SO indicate adding car points in the regions of external-occlusion, signal miss and self-occlusion, respectively, as visualized in (a).}
        \label{fig:recover}
    \end{figure*}
    With high-fidelity, the point clouds acquired by LiDAR sensors significantly improved autonomous agents's ability to understand 3D scenes. LiDAR-based models achieved state-of-the-art performance on 3D object classification \cite{xu2020grid}, visual odometry \cite{pan2021mulls}, and 3D object detection \cite{shi2020pv}. Despite being widely used in these 3D applications, LiDAR frames are technically 2.5D. After hitting the first object, a laser beam will return and leave the shapes behind the occluder missing from the point cloud. 
    
    To locate a severely occluded object (e.g., the car in Figure \ref{fig:teaser}(b)), a detector has to recognize the underlying object shapes even when most of its parts are missing. Since shape miss inevitably affects object perception, it is important to answer two questions: 
    \begin{itemize}[leftmargin=10pt] 
            \item \textit{What are the causes of shape miss in point clouds?}
            \item \textit{What is the impact of shape miss on 3D object detection?}
    \end{itemize}
    
    \subsection{Causes of Shape Miss}
        \label{sec:causes} 
        To answer the first question, we study the objects in KITTI \cite{geiger2013vision} and discover three causes of shape miss. 
        
        \vspace{5pt}
        \noindent\textbf{External-occlusion.} As visualized in Figure \ref{fig:teaser}(c), occluders block the laser beams from reaching the red frustums behind them. In this situation, the external-occlusion is formed, which causes the shape miss located at the red voxels.
        
        \vspace{5pt}
        \noindent\textbf{Signal miss.} As Figure \ref{fig:teaser}(c) illustrates, certain materials and reflection angles prevent laser beams from returning to the sensor after hitting some regions of the car (blue voxels). After projected to range view, the affected blue frustums in Figure \ref{fig:teaser}(c) appear as the empty pixels in Figure \ref{fig:teaser}(a).
        
        \vspace{5pt}
        \noindent\textbf{Self-occlusion.} LiDAR data is 2.5D by nature. As shown in Figure \ref{fig:teaser}(d), for a same object, its parts on the far side (the green voxels) are occluded by the parts on the near side. The shape miss resulting from self-occlusion inevitably happens to every object in LiDAR scans. 
        
    \subsection{Impact of Shape Miss}   
        \label{sec:impact} 
        To analyze the impact of shape miss on 3D object detection, we evaluate the car detection results of the scenarios where we recover certain types of shape miss on each object by borrowing points from similar objects (see the details of finding similar objects and filling points in Sec. 3.1). 
        
        In each scenario, after resolving certain shape miss in both the \textit{train} and \textit{val} split of KITTI \cite{geiger2013vision}, we train and evaluate a popular detector PV-RCNN \cite{shi2020pv}. The four scenarios are:  
        \begin{itemize}[topsep=0pt,leftmargin=10pt] 
            \item \textbf{NR:} Using the original data without shape miss recovery.
            \item \textbf{EO:} Recovering the shape miss caused by external-occlusion (adding the red points in Figure \ref{fig:recover}(a)).
            \item \textbf{EO+SM:} Recovering the shape miss caused by external-occlusion and signal miss (adding the red and blue points in Figure \ref{fig:recover}(a)). 
            \item \textbf{EO+SM+SO:} Recovering all the shape miss (adding the red, blue and green points in Figure \ref{fig:recover}(a)).  
        \end{itemize}
        \vspace{3pt}
        We report detection results on cars with three occlusion levels (level labels are provided by the dataset). As shown in Figure \ref{fig:recover}(b), without recovery (NR), it is more difficult to detect objects with higher occlusion levels. Recovering shapes miss will reduce the performance gaps between objects with different levels of occlusion .
        If all shape miss are resolved (EO+SM+SO), the performance gaps are eliminated and almost all objects can be effectively detected (APs $>$ $99\%$).
    \subsection{The Proposed Method}
       The above experiment manually resolves the shape miss by filling points into the labeled bounding boxes and significantly improve the detection results. However, during test time, how do we resolve shape miss without knowing bounding box labels? 
       
       In this paper, we propose \textit{Behind the Curtain Detector (BtcDet)}. To the best of our knowledge, BtcDet is the first 3D object detector that targets the object shapes affected by occlusion. With the knowledge of shape priors, BtcDet estimates the occupancy of complete object shapes in the regions affected by occlusion and signal miss. After being integrated into the detection pipeline, the occupancy estimation benefits both region proposal generation and proposal refinement. Eventually, BtcDet surpasses all of the state-of-the-art methods published to date by remarkable margins.
\begin{figure*} 
    \vspace{-5pt}
    \begin{adjustwidth}{-0pt}{-0pt}
        \centering
        \includegraphics[width=0.95\linewidth]{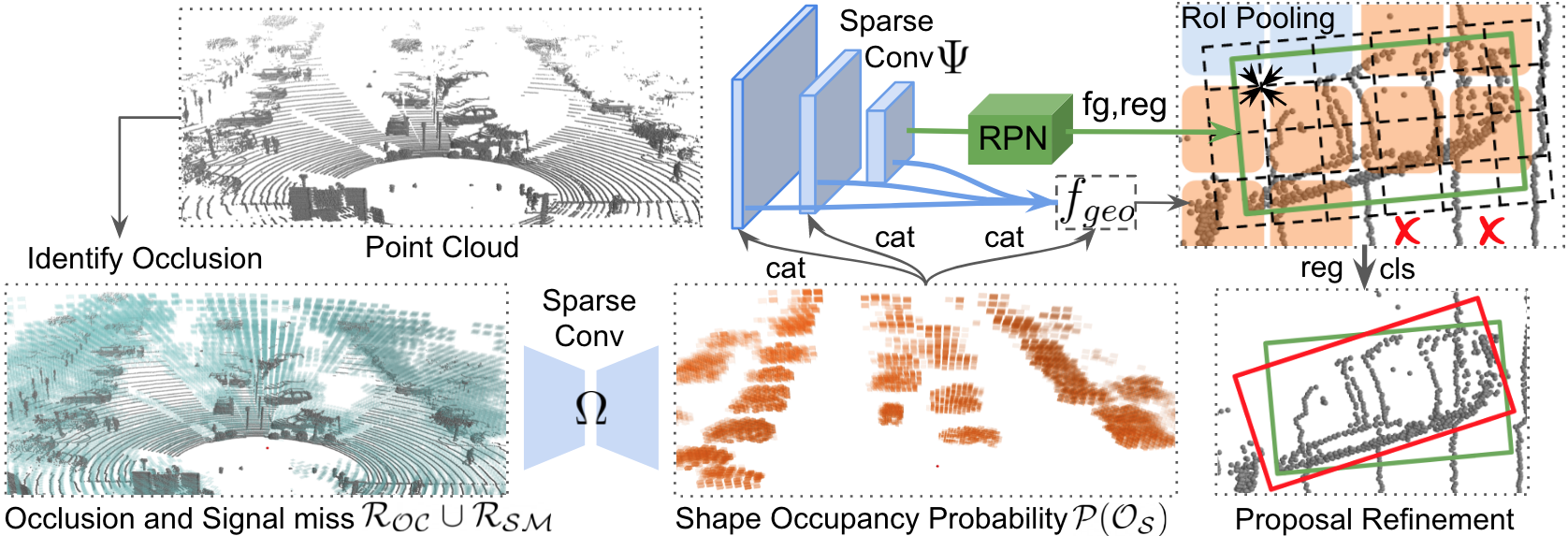}
        \caption{The detection pipeline. BtcDet first identifies the regions of occlusion and signal miss \ROCRSM\. In these regions, BtcDet estimates the shape occupancy probability \POS\ (the orangex voxels have \POS\ $> 0.3$). When the backbone network $\Psi$ extracts detection features from the point cloud, \POS\ is concatenated with $\Psi$'s intermediate feature maps. Then, a RPN network takes the output and generates 3D proposals. For each proposal (e.g., the green box), BtcDet pools the local geometric features $f_{geo}$ to the nearby grids and finally generate the final bounding box prediction (the red box) and the confidence score.}
        \label{fig:detection}
    \end{adjustwidth}
\end{figure*}
\section{Related Work}
    \label{sec:related}
    \vspace{5pt}
	\noindent\textbf{LiDAR-based 3D object detectors.} 
    Voxel-based methods divide point clouds by voxel grids to extract features \cite{zhou2018voxelnet}. Some of them also use sparse convolution to improve model efficiency, e.g., SECOND\cite{yan2018second}. Point-based methods such as PointRCNN~\cite{shi2019pointrcnn} generate proposals directly from points. STD~\cite{yang2019std} applies sparse to dense refinement and VoteNet~\cite{qi2019deep} votes the proposal centers from point clusters. These models are supervised on the ground truth bounding boxes without explicit consideration for the object shapes.
    
    \vspace{5pt}
	\noindent\textbf{Learning shapes for 3D object detection.}
    Bounding box prediction requires models to understand object shapes. Some detectors learn the shape related statistics as an auxiliary task. PartA$^2$~\cite{9018080} learns object part locations. SA-SSD and AssociateDet~\cite{he2020sassd,du2020associate} use auxiliary networks to preserve structural information. Studies~\cite{li2021sienet,yan2020sparse,najibi2020dops,xu2021spg} such as SPG conduct point cloud completion to improve object detection. These models are shape-aware but overlook the impact of occlusion on object shapes.
    
  
    \vspace{5pt}
	\noindent\textbf{Occlusion handling in computer vision.} The negative impact of occlusion on various computer vision tasks, including tracking \cite{liu2018context}, image-based pedestrian detection \cite{zhang2018occlusion}, image-based car detection \cite{reddy2019occlusion} and semantic part detection \cite{saleh2021occlusion}, is acknowledged. Efforts addressing occlusion include the amodal instance segmentation \cite{follmann2019learning}, the Multi-Level Coding that predicts the presence of occlusion \cite{qi2019amodal}. These studies, although focus on 2D images, demonstrate the benefits of modeling occlusion to solving visual tasks. Point cloud visibility is addressed in \cite{Hu_2020_CVPR} and is used in multi-frame detection and data augmentation. This method, however, does not learn and explore the visibility's influence on object shapes. Our proposed BtcDet is the first 3D object detector that learns occluded shapes in point cloud data. We compare \cite{Hu_2020_CVPR}'s approach with ours in Sec. \ref{sec:ablation}. 
	
\section{Behind the Curtain Detector}
    \label{sec:method}
    Let $\Theta$ denote the parameters of a detector, $\{p_1, p_2, ..., p_N\}$ denote the LiDAR point cloud, $\mathcal{X}, \mathcal{D}, {\mathcal{S}_{ob}}, {\mathcal{S}_{oc}}$ denote the estimated box center, the box dimension, the observed objects shapes and the occluded object shapes, respectively. Most LiDAR-based 3D object detectors \cite{yi2020segvoxelnet,chen2020object,shi2020point} only supervise the bounding box prediction. These models have
    \begin{adjustwidth}{0pt}{0pt}
        \begin{align}
            \Theta_{MLE}=\argmax_\Theta P(\mathcal{X}, \mathcal{D} \ | \ \{p_1, p_2, ..., p_N\}, \Theta), \label{prob:bbxonly}
        \end{align}
    \end{adjustwidth}
    while structure-aware models \cite{9018080,he2020sassd,du2020associate} also supervise ${\mathcal{S}_{ob}}$'s statistics so that
    \begin{adjustwidth}{0pt}{0pt}
        \begin{align}
            \Theta_{MLE}=\argmax_\Theta P(\mathcal{X}, \mathcal{D}, \mathcal{S}_{ob} \ | \ \{p_1, p_2, ..., p_N\}, \Theta). \label{prob:bbxshape}
        \end{align}
    \end{adjustwidth}
    
    None of the above studies explicitly model the complete object shapes $\mathcal{S} = \mathcal{S}_{ob} \cup \mathcal{S}_{oc}$, while the experiments in Sec. \ref{sec:impact} show the improvements if $\mathcal{S}$ is obtained. BtcDet estimates $\mathcal{S}$ by predicting the shape occupancy \OS\ for regions of interest. After that, BtcDet conducts object detection conditioned on the estimated probability of occupancy \POS. The optimization objectives can be described as follows:
    \begin{adjustwidth}{0pt}{0pt}
        \begin{align}
            &\argmax_{\Theta} P(\mathcal{O_S} \ | \ \{p_1, p_2, ..., p_N\}, \mathcal{R_{SM}}, \mathcal{R_{OC}}, \Theta), \label{prob:occ} \\
            &\argmax_\Theta P(\mathcal{X}, \mathcal{D} \ | \ \{p_1, p_2, ..., p_N\}, \mathcal{P}(\mathcal{O_{S}}), \Theta). \label{prob:det}
        \end{align}
    \end{adjustwidth}
    
    \vspace{5pt}
	\noindent\textbf{Model overview.} As illustrated in Figure \ref{fig:detection}, BtcDet first identifies the regions of occlusion $\mathcal{R_{OC}}$ and signal miss $\mathcal{R_{SM}}$, and then, let a shape occupancy network $\Omega$ estimate the probability of object shape occupancy \POS. The training process is described in Sec. \ref{sec:learnshape}. 
    
    Next, BtcDet extracts the point cloud 3D features by a backbone network $\Psi$. The features are sent to a Region Proposal Network (RPN) to generate 3D proposals. To leverage the occupancy estimation, the sparse tensor \POS\ is concatenated with the feature maps of $\Psi$. (See Sec. \ref{sec:agg}.) 
    
    Finally, BtcDet applies the proposal refinement. The local geometric features $f_{geo}$ are composed of \POS\ and the multi-scale features from $\Psi$. For each region proposal, we construct local grids covering the proposal box. BtcDet pools the local geometric features $f_{geo}$ onto the local grids, aggregates the grid features, and generates the final bounding box predictions. (See Sec. \ref{sec:refine}.) 
    
    \subsection{Learning Shapes in Occlusion}
        \label{sec:learnshape}
        \begin{figure*}[!htb]
            \begin{adjustwidth}{-0pt}{-0pt}
                \centering
                \includegraphics[width=1.00\linewidth]{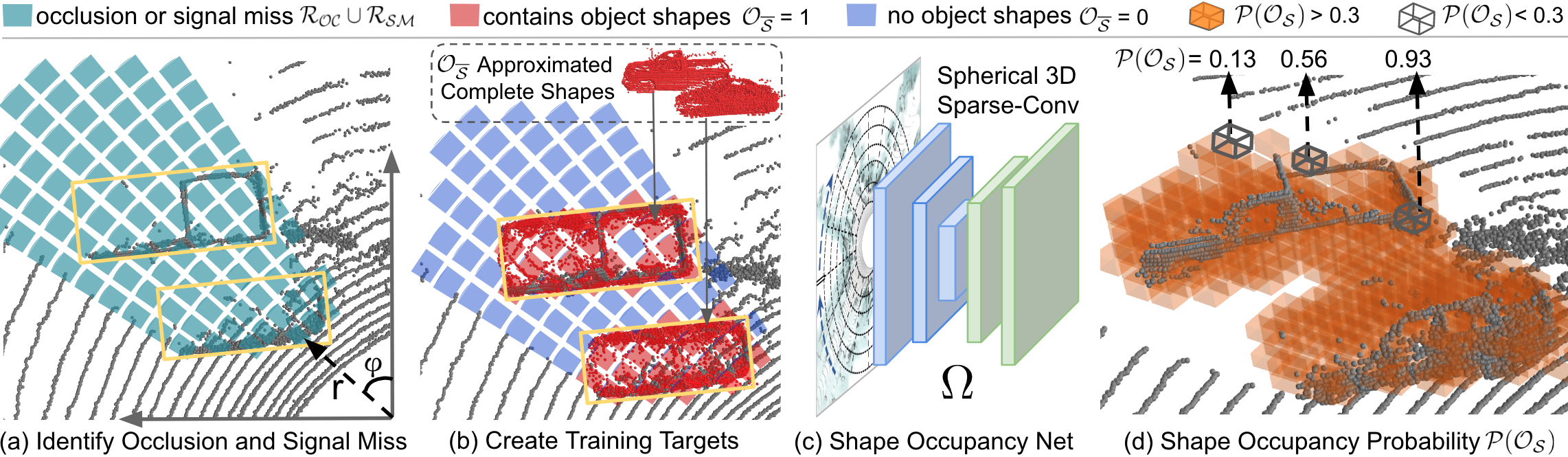}
                \caption{Learning Occluded Shapes. (a) The regions of occlusion or signal miss \ROCRSM\ can be identified after the spherical voxelization for the point cloud. (b) To label the occupancy \OSb\ (1 or 0), We place the approximated complete object shapes $\overline{S}$ (red points) in the corresponding boxes. (c) A shape occupancy network $\Omega$ predicts the shape occupancy probability \POS\ for voxels in \ROCRSM\ , supervised by \OSb. (d) Voxels are colored orange if it has a prediction \POS\ $> 0.3$. }
                \label{fig:occ}
            \end{adjustwidth}
        \end{figure*}
        \vspace{5pt}
		\noindent\textbf{Approximate the complete object shapes for ground truth labels.} Occlusion and signal miss preclude the knowledge of the complete object shapes $S$. However, we can assemble the approximated complete shapes $\overline{\mathcal{S}}$, based on two assumptions:
        \begin{itemize}[leftmargin=10pt] 
            \item Most foreground objects resemble a limited number of shape prototypes, e.g., pedestrians share a few body types.
            \item Foreground objects, especially vehicles and cyclists, are roughly symmetric. 
        \end{itemize}
        We use the labeled bounding boxes to query points belonging to the objects. For cars and cyclists, we mirror the object points against the middle section plane of the bounding box.
        
        A heuristic $\mathcal{H}(A,B)$ is created to evaluate if a source object $B$ covers most parts of a target object $A$ and provides points that can fill $A$'s shape miss. To approximate $A$'s complete shape, we select the top 3 source objects $B_1,B_2,B_3$ with the best scores. The final approximation $\overline{\mathcal{S}}$ consists of $A$'s original points and the points of $B_1,B_2,B_3$ that fill $A$'s shape miss. The target objects are the occluded object in the current training frame, while the source objects are other objects of the same class in the detection training set. Both can be extracted by the ground truth bounding boxes. Please find details of $\mathcal{H}(A,B)$ in Appendix B and more visualization of assembling $\overline{\mathcal{S}}$ in Appendix G. 
        
        \begin{figure}[!htb]
            \begin{adjustwidth}{-0pt}{-0pt}
                \centering
                \includegraphics[width=1.0\linewidth]{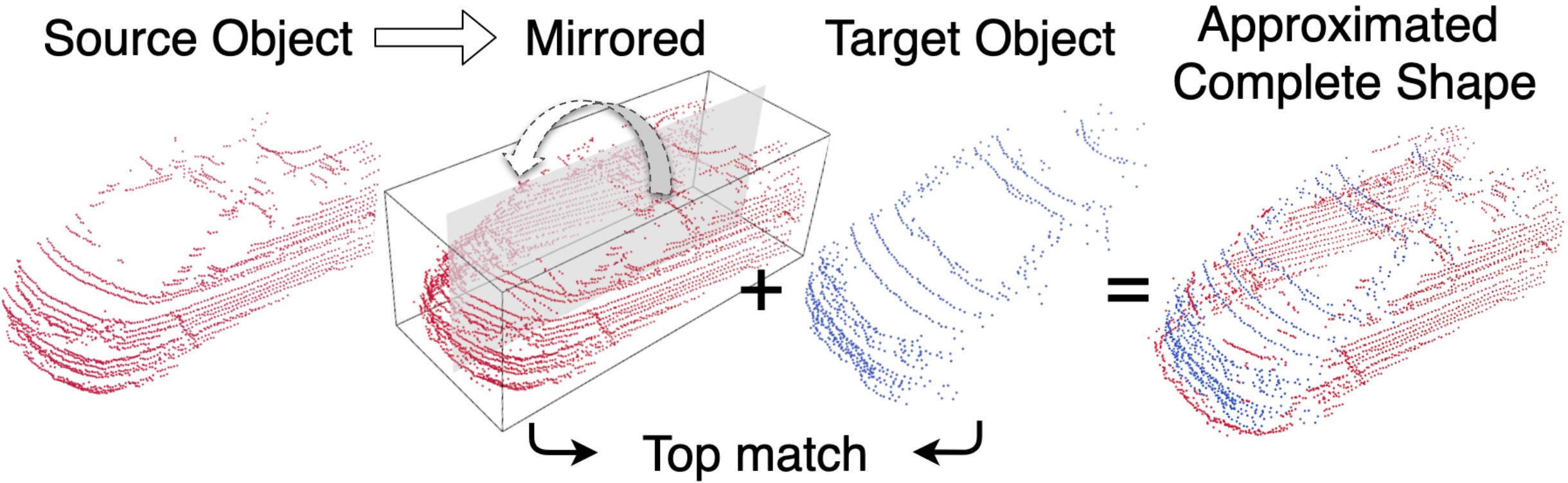}
                \caption{Assemble the approximated complete shape $\overline{\mathcal{S}}$ for an object (blue) by using points from top match objects.}
                \label{fig:compose_template}
            \end{adjustwidth}
        \end{figure}
        
        \vspace{5pt}
		\noindent\textbf{Identify \ROCRSM\ in the spherical coordinate system.} According to our analysis in Sec. \ref{sec:causes}, 
        ``shape miss'' only exists in the occluded regions $\mathcal{R_{OC}}$ and the regions with signal miss $\mathcal{R_{SM}}$ (see Figure \ref{fig:teaser}(c) and (d)).
        Therefore, we need to identify \ROCRSM\ before learning to estimate shapes.
        
        In real-world scenarios, there exists at most one point in the tetrahedron frustum of a range image pixel. When the laser is stopped at a point, the entire frustum behind the point is occluded. We propose to voxelize the point cloud using an evenly spaced spherical grid so that the occluded regions can be accurately formed by the spherical voxels behind any LiDAR point.
        As shown in Figure \ref{fig:occ}(a), each point ($x,y,z$) is transformed to the spherical coordinate system as ($r,\phi,\theta$):
        \begin{adjustwidth}{0pt}{0pt}
            \begin{align}
                r &= \sqrt{(x^2 +  y^2 + z^2)}, \ \ \ \phi = arctan2(y, x), \label{eq:coordinate} \\ \nonumber
                \theta &= arctan2(z, \sqrt{x^2 + y^2}).
            \end{align}
        \end{adjustwidth}
        $\mathcal{R_{OC}}$ includes nonempty spherical voxels and the empty voxels behind these voxels. In Figure \ref{fig:teaser}(a), the dashed lines mark the potential areas of signal miss. In range view, we can find pixels on the borders between the areas having LiDAR signals and the areas of no signal. $\mathcal{R_{SM}}$ is formed by the spherical voxels that project to these pixels.
        
        \vspace{5pt}
		\noindent\textbf{Create training targets.} In \ROCRSM\ , we predict the probability \POS\ for voxels if they contain points of $\overline{\mathcal{S}}$. As illustrated in \ref{fig:occ}(b), $\overline{\mathcal{S}}$ are placed at the locations of the corresponding objects. We set \OSb\ $=1$ for the spherical voxels that contain $\overline{\mathcal{S}}$, and \OSb\ $=0$ for the others. \OSb\ is used as the ground truth label to approximate the occupancy  \OS\ of the complete object shape.
        Estimating occupancy has two advantages over generating points:
        \begin{itemize}[leftmargin=10pt]
            \item $\overline{\mathcal{S}}$ is assembled by multiple objects. The shape details approximated by the borrowed points are inaccurate and the point density of different objects is inconsistent. The occupancy \OSb\ avoids these issues after rasterization. 
            \item The plausibility issue of point generation can be avoided.
        \end{itemize}
    	\vspace{5pt}
		\noindent\textbf{Estimate the shape occupancy.}
        In \ROCRSM, we encode each nonempty spherical voxel with the average properties of the points inside (x,y,z,feats), then, send them to a shape occupancy network $\Omega$. The network consists of two down-sampling sparse-conv layers and two up-sampling inverse-convolution layers. Each layer also includes several sub-manifold sparse-convs \cite{graham2017submanifold} (see Appendix D). The spherical sparse 3D convolutions are similar to the ones in the Cartesian coordinate, except that the voxels are indexed along ($r,\phi,\theta$). The output \POS\ is supervised by the sigmoid cross-entropy Focal Loss \cite{lin2017focal}:
        \begin{adjustwidth}{0pt}{0pt}
            \begin{align}
                \mathcal{L}_{focal}(p_v) &= -(1-p_v)^{\gamma}log(p_v), \label{loss:focal} \\ \nonumber
                \text{where} & \  p_v =
                \begin{cases}
                  \text{\POS\ } & \text{if \OSb\ = 1 at voxel \ } v \\
                  1 - \text{\POS\ } & \text{otherwise,} 
                \end{cases}  \\
                \mathcal{L}_{shape} &= \frac{\sum_{v \in \mathcal{R_{OC}} \cup \mathcal{R_{SM}}} w_v \cdot \mathcal{L}_{focal}(p_v)}{|\mathcal{R_{OC}} \cup \mathcal{R_{SM}}|}, \label{loss:shape} \\ \nonumber
                \text{where} &  \  w_v =
                \begin{cases}
                  \delta & \text{if} \ v \in \text{regions of shape miss} \\
                  1 & \text{otherwise.} 
                \end{cases}
            \end{align}
        \end{adjustwidth}
        Since $\overline{\mathcal{S}}$ borrows points from other objects in the shape miss regions, we assign them a weighting factor $\delta$, where $\delta<1$.
        
    \subsection{Shape Occupancy Probability Integration}
    \label{sec:agg}
        Trained with the customized supervision, $\Omega$ learns the shape priors of partially observed objects and generates \POS. To benefit detection, \POS\ is transformed from the spherical coordinate to the Cartesian coordinate and fused with $\Psi$, a sparse 3D convolutional network that extracts detection features in the Cartesian coordinate.. 
        
        For example, a spherical voxel has a center ($r,\phi,\theta$) which is transformed as $x = r cos\theta cos\phi, \ y = r  cos\theta  sin\phi, \ z = r  sin\theta$.
        Assume $x,y,z$ is inside a Cartesian voxel $v^{i,j,k}$. Since several spherical voxels  can be mapped to $v^{i,j,k}$, $v^{i,j,k}$ takes the max value of these voxels $SV(v^{i,j,k})$:
        \begin{adjustwidth}{0pt}{0pt}
            \begin{align}
                \mathcal{P}(\mathcal{O_{S}})_{v^{ijk}} = max (\{\mathcal{P}(\mathcal{O_{S}})_{sv}:sv \in SV(v^{i,j,k}) \}).
            \end{align}
        \end{adjustwidth}
        The occupancy probability of these Cartesian voxels forms a sparse tensor map $\mathcal{P}(\mathcal{O_{S}})_{\perp} = \{\mathcal{P}(\mathcal{O_{S}})_{v}\}$, which is, then, down-sampled by max-poolings into multiple scales and concatenated with $\Psi$'s intermediate feature maps:
        \begin{adjustwidth}{0pt}{0pt}
            \begin{align}
                f_{\Psi_{i}}^{in} = \big[ f_{\Psi_{i-1}}^{out}, \ maxpool^{\ i-1}_{\times 2} ( \mathcal{P}(\mathcal{O_{S}})_{\perp} ) \big],
                \label{eq:psi_fuse} 
            \end{align}
        \end{adjustwidth}  
       where $f_{\Psi_{i}}^{in}$, $f_{\Psi_{i-1}}^{out}$ and $maxpool^{\ i-1}_{\times 2}(\cdot)$ denote the input features of $\Psi$'s $i$th layer, the output features of $\Psi$'s $i-1$th layer, and applying stride-$2$ maxpooling $i-1$ times, respectively.
       
       The Region Proposal Network (RPN) takes the output features of $\Psi$ and generates 3D proposals. Each proposal includes  $(x_p,y_p,z_p), (l_p,w_p,h_p), \theta_p, p_p$, namely, center location, proposal box size, heading and proposal confidence. 
       
    \subsection{Occlusion-Aware Proposal Refinement}
    \label{sec:refine}
        \vspace{5pt}
		\noindent\textbf{Local geometry features.} BtcDet's refinement module further exploits the benefit of the shape occupancy. To obtain accurate final bounding boxes, BtcDet needs to look at the local geometries around the proposals. Therefore, we construct a local feature map $f_{geo}$ by fusing multiple levels of $\Psi$'s features. In addition, we also fuse \POSC\ into $f_{geo}$ to bring awareness to the shape miss in the local regions. $\mathcal{P}(\mathcal{O_{S}})_{\perp}$ provides two benefits for proposal refinement:
        \begin{itemize}[leftmargin=10pt]
            \item \POSC\ only has values in \ROCRSM\ so that it can help the box regression avoid the regions outside \ROCRSM, e.g., the regions with cross marks in Figure \ref{fig:detection}. 
            \item The estimated occupancy indicates the existence of unobserved object shapes, especially for empty regions with high \POS\ , e.g., some orange regions in Figure \ref{fig:detection}.
        \end{itemize}
        $f_{geo}$ is a sparse 3D tensor map with spatial resolution of $400\times352\times5$. The process for producing $f_{geo}$ is described in Appendix D. 
        
        \vspace{5pt}
		\noindent\textbf{RoI pooling.} On each proposal, we construct local grids which have the same heading of the proposal. To expand the receptive field, we set a size factor $\mu$ so that:
        \begin{adjustwidth}{0pt}{0pt}
            \begin{align}
                w_{grid} = \mu \cdot w_p, \ \ \  l_{grid} = \mu \cdot l_p, \ \ \ h_{grid} = \mu \cdot h_p.
                \label{eq:proportion_grid}
            \end{align}
        \end{adjustwidth}
        The grid has a dimension of $12\times4\times2$. We pool the nearby features $f_{geo}$ onto the nearby grids through trilinear-interpolation (see Figure \ref{fig:detection}) and aggregates them by sparse 3D convolutions. After that, the refinement module predicts an IoU-related class confidence score and the residues between the 3D proposal boxes and the ground truth bounding boxes, following \cite{yan2018second,shi2020pv}.
    \subsection{Total Loss}
        The RPN loss $\mathcal{L}_{rpn}$ and the proposal refinement loss $\mathcal{L}_{pr}$ follow the most popular design among detectors \cite{shi2020pv,yan2018second}. The total loss is:
	    \begin{align}
            \mathcal{L}_{total} = 0.3 \mathcal{L}_{shape} + \mathcal{L}_{rpn} + \mathcal{L}_{pr}.
        \end{align}
        More details of the losses and the network architectures can be found in Appendix C and D.
        
\section{Experiments}
    \label{sec:experiment}
    In this section, we describe the implementation details of BtcDet and compare BtcDet with state-of-the-art detectors on two datasets: the KITTI Dataset~\cite{geiger2013vision} and the Waymo Open Dataset~\cite{sun2019scalability}. We also conduct ablation studies to demonstrate the effectiveness of the shape occupancy and the feature integration strategies. More detection results can be found in the Appendix F. The quantitative and qualitative evaluations of the occupancy estimation can be found in the Appendix E and H.
    
    \vspace{5pt}
    \noindent\textbf{Datasets.} The \textit{KITTI Dataset} includes 7481 LiDAR frames for training and 7518 LiDAR frames for testing. We follow~\cite{REF:Multiview3D_2017} to divide the training data into a \textit{train} split of 3712 frames and a \textit{val} split of 3769 frames. The \textit{Waymo Open Dataset} (WOD) consists of 798 segments of 158361 LiDAR frames for training and 202 segments of 40077 LiDAR frames for validation. The KITTI Dataset only provides LiDAR point clouds in 3D, while the WOD also provides LiDAR range images.
    
	\vspace{5pt}
    \noindent\textbf{Implementation and training details.} BtcDet transforms the point locations ($x,y,z$) to ($r,\phi,\theta$) for the KITTI Dataset, while directly extracting ($r,\phi,\theta$) from the range images for the WOD. On the KITTI Dataset, we use a spherical voxel size of ($0.32m, 0.52^{\circ}, 0.42^{\circ}$) within the range [$2.24m, 70.72m$] for $r$, [$-40.69^{\circ}, 40.69^{\circ}$] for $\phi$ and [$-16.60^{\circ}, 4.00^{\circ}$] for $\theta$. On the WOD, we use a spherical voxel size of ($0.32m, 0.81^{\circ}, 0.31^{\circ}$) within the range [$2.94m, 74.00m$] for $r$, [$-180^{\circ}, 180^{\circ}$] for $\phi$ and [$-33.80^{\circ}, 6.00^{\circ}$] for $\theta$. Determined by grid search, we set $\gamma = 2$ in Eq.\ref{loss:focal}, $\delta = 0.2$ in Eq.\ref{loss:shape} and $\mu = 1.05$ in Eq.\ref{eq:proportion_grid}. 
    
    In all of our experiments, we train our models with a batch size of 8 on 4 GTX 1080 Ti GPUs. On the KITTI Dataset, we train BtcDet for 40 epochs, while on the WOD, we train BtcDet for 30 epochs. The BtcDet is end-to-end optimized by the ADAM optimizer \cite{kingma2014adam} from scratch. We applies the widely adopted data augmentations \cite{shi2020pv,deng2020voxel,lang2019pointpillars,yang20203dssd,ye2020hvnet}, which includes flipping, scaling, rotation and the ground-truth augmentation.
    \subsection{Evaluation on the KITTI Dataset}
        \begin{table*} 
        \begin{adjustwidth}{-0pt}{0pt}
        \setlength\extrarowheight{1.0pt}
        \centering
        {\small
            \begin{tabular}{l|ccc|ccc|ccc|c}
            \hline
            \multirow{2}{*}{Method} & \multicolumn{3}{c|}{Car 3D $AP_{R40}$}                & \multicolumn{3}{c|}{Ped. 3D $AP_{R40}$} & \multicolumn{3}{c|}{Cyc. 3D $AP_{R40}$}              &  3D $AP_{R11}$     \\
                                    & Easy           & Mod.           & Hard           & Easy       & Mod.       & Hard      & Easy           & Mod.           & Hard           & Car Mod.           \\ \hline
            PointPillars \cite{lang2019pointpillars}           & 87.75          & 78.39          & 75.18          & 57.30      & 51.41      & 46.87     & 81.57          & 62.94          & 58.98          & 77.28          \\
            SECOND \cite{yan2018second}                 & 90.97          & 79.94          & 77.09          & 58.01      & 51.88      & 47.05     & 78.50          & 56.74          & 52.83          & 76.48          \\
            SA-SSD \cite{he2020sassd}                 & 92.23          & 84.30          & 81.36          & -          & -          & -         & -              & -              & -              & 79.91          \\
            PV-RCNN \cite{shi2020pv}                & 92.57          & 84.83          & 82.69          & 64.26      & 56.67      & 51.91     & 88.88          & 71.95          & 66.78          & 83.90          \\
            Voxel R-CNN \cite{deng2020voxel}             & 92.38          & 85.29          & 82.86          & -          & -          & -         & -              & -              & -              & 84.52 \\ \hline
            BtcDet (Ours)           & \textbf{93.15} & \textbf{86.28} & \textbf{83.86} & \textbf{69.39} & \textbf{61.19}      & \textbf{55.86}     & \textbf{91.45} & \textbf{74.70} & \textbf{70.08} & \textbf{86.57} \\ \hline
            \end{tabular}
         }
	    \end{adjustwidth}
	    \caption{Comparison on the KITTI \textit{val} set, evaluated by the 3D Average Precision (AP) under 40 recall thresholds (R40). The 3D APs on under 11 recall thresholds are also reported for the moderate car objects.}
        \label{tb:kitti_val}
	\end{table*}
	\begin{table*}[!hbt]
            \begin{adjustwidth}{0pt}{0pt}
                \setlength\extrarowheight{1pt}
                \setlength\tabcolsep{5pt}
                \centering
                {\small
                \begin{tabular}{l|lc|cccc|cccc}
                \hline
                \multirow{2}{*}{\normalsize Method}  & \multirow{2}{*}{\normalsize Reference} & \multirow{2}{*}{\normalsize Modality} & \multicolumn{4}{c|}{\normalsize Car 3D $AP_{R40}$}                      & \multicolumn{4}{c}{\normalsize Cyc. 3D $AP_{R40}$}  \\       &              &             & Easy           & Mod.           & Hard           & mAP            & Easy           & Mod.           & Hard           & mAP            \\ \hline
                \small EPNet \cite{huang2020epnet}                    & \small ECCV 2020                  & \small LiDAR+RGB                & 89.81          & 79.28          & 74.59          & 81.23          & -              & -              & -              & -              \\
                \small 3D-CVF \cite{yoo20203d}                   & \small ECCV 2020                  & \small LiDAR+RGB                & 89.20          & 80.05          & 73.11          & 80.79          & -              & -              & -              & -              \\ \hline
                \small PointPillars \cite{lang2019pointpillars}             & \small CVPR 2019                  & \small LiDAR                     & 82.58          & 74.31          & 68.99          & 75.29          & 77.10          & 58.65          & 51.92          & 62.56          \\
                \small STD \cite{yang2019std}                      & \small ICCV 2019                  & \small LiDAR                     & 87.95          & 79.71          & 75.09          & 80.92          & 78.69          & 61.59          & 55.30          & 65.19          \\
                \small HotSpotNet \cite{chen2020object}               & \small ECCV 2020                  & \small LiDAR                     & 87.60          & 78.31          & 73.34          & 79.75          & 82.59          & 65.95          & 59.00          & 69.18          \\
                \small PartA$^2$ \cite{9018080} & \small TPAMI 2020                 & \small LiDAR                     & 87.81          & 78.49          & 73.51          & 79.94          & 79.17          & 63.52          & 56.93          & 66.54          \\
                \small 3DSSD \cite{yang20203dssd}                    & \small CVPR 2020                  & \small LiDAR                     & 88.36          & 79.57          & 74.55          & 80.83          & 82.48          & 64.10          & 56.90          & 67.83          \\
                \small SA-SSD \cite{he2020sassd}                   & \small CVPR 2020                  & \small LiDAR                     & 88.75          & 79.79          & 74.16          & 80.90          & -              & -              & -              & -              \\
                \small Asso-3Ddet \cite{du2020associate}               & \small CVPR 2020                  & \small LiDAR                     & 85.99          & 77.40          & 70.53          & 77.97          & -              & -              & -              & -              \\
                \small PV-RCNN \cite{shi2020pv}                   & \small CVPR 2020                  & \small LiDAR                     & 90.25          & 81.43          & 76.82          & 82.83          & 78.60          & 63.71          & 57.65          & 66.65          \\
                \small Voxel R-CNN \cite{deng2020voxel}              & \small AAAI 2021                  & \small LiDAR                     & \textbf{90.90} & 81.62          & 77.06          & 83.19          & -              & -              & -              & -              \\
                \small CIA-SSD \cite{zheng2020ciassd}                  & \small AAAI 2021                  & \small LiDAR                     & 89.59          & 80.28          & 72.87          & 80.91          & -              & -              & -              & -              \\
                \small TANet \cite{liu2020tanet}                    & \small AAAI 2021                  & \small LiDAR                     & 83.81          & 75.38          & 67.66          & 75.62          & 73.84          & 59.86          & 53.46          & 62.39          \\ \hline
                BtcDet (Ours)            & \small AAAI 2022      & \small LiDAR                     & 90.64          & \textbf{82.86} & \textbf{78.09} & \textbf{83.86} & \textbf{82.81} & \textbf{68.68} & \textbf{61.81} & \textbf{71.10} 
                    \\ 
                \rowcolor{LightCyan} \textit{Improvement}	& \multicolumn{1}{c}{-}   &   -   & \textit{-0.26}	    & \textit{+1.24}    &\textit{+0.94} & \textit{+0.67} &   \textit{+0.33}        &	\textit{+2.73} & \textit{+2.81} & \textit{+1.92} \\ \hline
                \end{tabular}
                }
        	    \caption{Comparison on the KITTI \textit{test} set, evaluated by the 3D Average Precision (AP) of 40 recall thresholds (R40) on the KITTI server. BtcDet surpasses all the leader board front runners that are associated with publications released before our submission. The mAPs are averaged over the APs of easy, moderate, and hard objects. Please find more results in Appendix F.} 
    	        \label{tb:kitti_test}
    	    \end{adjustwidth}
		\end{table*}
		
        We evaluate BtcDet on the KITTI \textit{val} split after training it on the \textit{train} split. To evaluate the model on the KITTI test set, we train BtcDet on $80\%$ of all \textit{train}+\textit{val} data and hold out the remaining 20\% data for validation. Following the protocol in \cite{geiger2013vision}, results are evaluated by the Average Precision (AP) with an IoU threshold of 0.7 for cars and 0.5 for pedestrians and cyclists.
        
    	\vspace{5pt}
        \noindent\textbf{KITTI validation set.} As summarized in Table \ref{tb:kitti_val}, we compare BtcDet with the state-of-the-art LiDAR-based 3D object detectors on cars, pedestrians and cyclists using the AP under 40 recall thresholds (R40). We reference the R40 APs of SA-SSD, PV-RCNN and Voxel R-CNN to their papers, the R40 APs of SECOND to \cite{pang2020clocs} and the R40 APs of PointRCNN and PointPillars to the results of the officially released code. We also report the published 3D APs under 11 recall thresholds (R11) for the moderate car objects. 
        On all object classes and difficulty levels, BtcDet outperforms models that only supervise bounding boxes (Eq.\ref{prob:bbxonly}) as well as structure-aware models (Eq.\ref{prob:bbxshape}). Specifically, BtcDet outperforms other models by $2.05\%$ 3D R11 AP on the moderate car objects,  which makes it the first detector that reaches above $86\%$ on this primary metric.
        
		\vspace{5pt}
		\noindent\textbf{KITTI test set.} As shown in Table \ref{tb:kitti_test}, we compare BtcDet with the front runners on the KITTI test leader board. Besides the official metrics, we also report the mAPs that average over the APs of easy, moderate, and hard objects. As of May. 4th, 2021, compared with all the models associated with publications, BtcDet \textbf{surpasses} them on car and cyclist detection by big margins. Those methods include the models that take inputs of both LiDAR and RGB images and the ones taking LiDAR input only. We also list more comparisons and the results in Appendix F. 

    \vspace{-5pt}
    \subsection{Evaluation on the Waymo Open Dataset}
        \begin{table*}[!hbt]
            \begin{adjustwidth}{-10pt}{-10pt}
                \setlength\extrarowheight{1.0pt}
                \setlength\tabcolsep{4.5pt}
                \centering
                {\small
                    \begin{tabular}{l|ccccc|cccccc}
                    \cline{1-11}
                    \multicolumn{1}{c|}{} & \multicolumn{4}{c}{LEVEL\_1 \ 3D mAP} & mAPH  & \multicolumn{4}{c}{LEVEL\_2 \ 3D mAP} &  mAPH  &  \\
                    Method                & Overall  & 0-30m & 30-50m & 50m-Inf & Overall & Overall  & 0-30m & 30-50m & 50m-Inf & Overall &  \\ \cline{1-11}
                    PointPillar \cite{lang2019pointpillars}           & 56.62    & 81.01 & 51.75  & 27.94   & -       & -        & -     & -      & -       & -       &  \\
                    MVF \cite{zhou2020end}                  & 62.93    & 86.30 & 60.02  & 36.02   & -       & -        & -     & -      & -       & -       &  \\
                    SECOND \cite{yan2018second}               & 72.27    & -     & -      & -       & 71.69   & 63.85    & -     & -      & -       & 63.33   &  \\
                    Pillar-OD \cite{wang2020pillar}      & 69.80    & 88.53 & 66.50  & 42.93   & -       & -        & -     & -      & -       & -       &  \\
                    AFDet \cite{ge2020afdet}        & 63.69    & 87.38 & 62.19  & 29.27   & -       & -        & -     & -      & -       & -       &  \\
                    PV-RCNN \cite{shi2020pv}        & 70.30    & 91.92 & 69.21  & 42.17   & 69.69   & 65.36    & 91.58 & 65.13  & 36.46   & 64.79   &  \\
                    Voxel R-CNN \cite{deng2020voxel}       & 75.59    & 92.49 & 74.09  & 53.15   & -       & 66.59    & 91.74 & 67.89  & 40.80   & -       &  \\ \cline{1-11}
                    BtcDet (ours)  	 &  \textbf{78.58}  &   \textbf{96.11} &  \textbf{77.64}  &  \textbf{54.45}  &  \textbf{78.06} &  \textbf{70.10}	&    \textbf{95.99}	  &   \textbf{70.56}	 &   \textbf{43.87}  &    \textbf{69.61}   &  \\ \cline{1-11}
                    \end{tabular}
                }
    	    \end{adjustwidth}
    	    \caption{Comparison for vehicle detection on the Waymo Open Dataset validation set.}
	        \label{tb:waymo}
		\end{table*}
		
        We also compare BtcDet with other models on the Waymo Open Dataset (WOD). We report both 3D mean Average Precision (mAP) and 3D mAP weighted by Heading (mAPH) for vehicle detection. The official metrics also include separate mAPs for objects belonging to different distance ranges. Two difficulty levels are also introduced, where the LEVEL\_1 mAP calculates for objects that have more than 5 points and the LEVEL\_2 mAP calculates for objects that have more than 1 point.
        
        As shown in Table \ref{tb:waymo}, BtcDet outperforms these state-of-the-art detectors on all distance ranges and all difficulty levels by big margins. BtcDet outperforms other detectors on the LEVEL\_1 3D mAP by $2.99\%$ and the LEVEL\_2 3D mAP by $3.51\%$. In general, BtcDet brings more improvement on the LEVEL\_2 objects, since objects with fewer points usually suffer more from occlusion and signal miss. These strong results on WOD, one of the largest published LiDAR datasets, manifest BtcDet's ability to generalize.
        
    \subsection{Ablation Studies}
        \label{sec:ablation}
         We conduct ablation studies to demonstrate the effectiveness of the shape occupancy and the feature integration strategies. All model variants are trained on the KITTI \textit{train} split and evaluated on the \textit{val} split.
        \begin{table}[!htb]
            \begin{adjustwidth}{-0pt}{-0pt}
            \centering
                {\small
                    \begin{tabular}{lccc}
                        \hline
                        \begin{tabular}[c]{@{}l@{}}Model \\ Variant\end{tabular}             & \begin{tabular}[c]{@{}c@{}}Learned \\ Features\end{tabular} & \begin{tabular}[c]{@{}c@{}}Integrated \\ Features\end{tabular}      & \begin{tabular}[c]{@{}c@{}}3D $AP_{R11}$\\ Car Mod.\end{tabular} \\ 
                        \hline
                        BtcDet$_1$(base) & $-$   & $-$   & 83.71  \\
                        BtcDet$_2$  & $-$ &  \ROCRSM  & 84.01 \\
                        BtcDet$_3$  & \POS$_{\perp}$ &  \POS$_{\perp}$  & 86.03 \\
                        BtcDet$_4$   & \POS$_{\circledcirc}$ &  $\mathfrak{1}$(\POS$_{\perp}$ $\geq 0.5$) & 85.59 \\\hline\hline
                        \textbf{BtcDet (main)} &  \POS$_{\circledcirc}$  & \POS$_{\perp}$ & \textbf{86.57} \\ \hline 
                    \end{tabular}
                }
                \captionsetup{margin={1pt,0pt}}
                \caption{Ablation studies on the learned features (Sec. \ref{sec:learnshape}) and the features fused into $\Psi$ and $f_{geo}$ (Sec. \ref{sec:agg}). BtcDet$_2$ directly use a binary map that labels \ROCRSM. $\circledcirc$ and $\perp$ indicate the spherical and the Cartesian coordinate. The ``$\mathfrak{1}$'' operator converts float values to binary codes with a threshold of 0.5. All variants share the same architecture.}
                \label{tb:ab_feat}
            \end{adjustwidth}
        \end{table}

        \vspace{5pt}
		\noindent\textbf{Shape Features.} As shown in Table \ref{tb:ab_feat}, we conduct ablation studies by controlling the shape features learned by $\Omega$ and the features used in the integration. All the model variants share the same architecture and integration strategies. 

        Similarly to \cite{Hu_2020_CVPR}, \textbf{BtcDet$_2$} directly fuses the binary map of. \ROCRSM\ into the detection pipeline. Although the binary map provides the information of occlusion, the improvement is limited since that the regions with code 1 are mostly background regions and less informative. 
        
        \textbf{BtcDet$_3$} learns \POS$_{\perp}$ directly. The network $\Omega$ predicts probability for Cartesian voxels. One Cartesian voxel will cover multiple spherical voxels when being close to the sensor, and will cover a small portion of a spherical voxel when being located at a remote distance. Therefore, the occlusion regions are misrepresented in the Cartesian coordinate. 
        
        \textbf{BtcDet$_4$} convert the probability to hard occupancy, which cannot inform the downstream branch if a region is less likely or more likely to contain object shapes. 
        
        These experiments demonstrate the effectiveness of our choices for shape features, which help the main model improve $2.86$ AP over the baseline \textbf{BtcDet$_1$}.
        
        \begin{table}[!htb]
            \begin{adjustwidth}{-0pt}{-0pt}
            \centering
            \setlength\tabcolsep{2.5pt}
                {\small
                    \begin{tabular}{lcccc}
                        \hline
                        \begin{tabular}[l]{@{}l@{}}Model \\ Variant\end{tabular}             & \begin{tabular}[c]{@{}c@{}}Integrate \\ Layers of $\Psi$ \end{tabular} & \begin{tabular}[c]{@{}c@{}}Integrate  \\ $f_{geo}$ \end{tabular}      &
                        \begin{tabular}[c]{@{}c@{}}Proposal bbox\\ 3D $AP_{R11}$ \end{tabular} & \begin{tabular}[c]{@{}c@{}}Final bbox\\ 3D $AP_{R11}$ \end{tabular} \\ 
                        \hline
                        BtcDet$_1$(base)  & $-$ &  $-$  & 77.75 &  83.71 \\
                        BtcDet$_5$  & $-$ & $\checkmark$ & 77.73 & 84.50 \\
                        BtcDet$_6$  & 1,2 &  $-$  & \textbf{78.97} &  85.72 \\
                        BtcDet$_7$  & 1 &  $\checkmark$  & 78.54 & 85.73 \\
                        BtcDet$_8$  & 1,2,3 &  $\checkmark$  & 78.76 & 86.11 \\\hline\hline
                        \textbf{BtcDet (main)} & 1,2 & $\checkmark$ & 78.93 & \textbf{86.57} \\ \hline 
                    \end{tabular}
                }
                \captionsetup{margin={1pt,0pt}}
                \caption{Ablation studies on which layers of $\Psi$ are fused with \POS$_\perp$ (Eq. \ref{eq:psi_fuse}) and whether to fuse \POS$_\perp$ into $f_{geo}$. We evaluate on the KITTI's moderate car objects and show the 3D $AP_{R11}$ of the proposal and final bounding box.}
                \label{tb:ab_fuse}
            \end{adjustwidth}
        \end{table}
        
    	\vspace{5pt}
		\noindent\textbf{Integration strategies.} We conduct ablation studies by choosing different layers of $\Psi$ to concatenate with \POS$_{\perp}$ and whether to use \POS$_{\perp}$ to form $f_{geo}$. The former mostly affects the proposal generation, while the latter affects proposal refinement. 
		
		In Table \ref{tb:ab_fuse}, the experiment on \textbf{BtcDet$_5$} shows that we can improve the final prediction AP by $0.8$ if we only integrate \POS$_{\perp}$ for proposal refinement. On the other hand, the experiment on \textbf{BtcDet$_6$} shows the integration with $\Psi$ alone can improve the AP by $1.2$ for proposal box and final bounding box prediction AP by $2.0$ over the baseline. 
		
		The comparisons of  \textbf{BtcDet$_7$}, \textbf{BtcDet$_8$} and BtcDet (main) demonstrates integrating \POS$_\perp$ with $\Psi$'s first two layers is the best choice. Since \POS\ is a low level feature while the third layer of $\Psi$ would contain high level features, we observe a regression when BtcDet$_8$ also concatenates \POS$_\perp$ with $\Psi$'s third layer.
		
		These experiments demonstrate both the integration with $\Psi$ and the integration to form $f_{geo}$ can bring improvement independently. When working together, two integrations finally help BtcDet surpass all the state-of-the-art models.
        
\section{Conclusion and Future Work}
    \label{sec:conclusion}
    In this paper, we analyze the impact of shape miss on 3D object detection, which is attributed to occlusion and signal miss in point cloud data. To solve this problem, we propose Behind the Curtain Detector (BtcDet), the first 3D object detector that targets this fundamental challenge. A training method is designed to learn the underlying shape priors. BtcDet can faithfully estimate the complete object shape occupancy for regions affected by occlusion and signal miss. After the integration with the probability estimation, both the proposal generation and refinement are significantly improved. In the experiments on the KITTI Dataset and the Waymo Open Dataset, BtcDet surpasses all the published state-of-the-art methods by remarkable margins. Ablation studies further manifest the effectiveness of the shape features and the integration strategies. Although our work successfully demonstrates the benefits of learning occluded shapes, there is still room to improve the model efficiency. Designing models that expedite occlusion identification and shape learning can be a promising future direction. 

\begin{appendices}

    \twocolumn[{%
        \renewcommand\twocolumn[1][]{#1}%
        \begin{center}
            \centering
            \LARGE \textbf{\appendixname}
            \vspace{30pt}
        \end{center}%
    }]

    \section{ Data and Code License}
        The datasets we use for experiments are the KITTI Dataset~\cite{geiger2013vision} and the Waymo Open Dataset~\cite{sun2019scalability}. Both of them are well-known and licensed for academic research. 
        
        We license our code under ``Apache License 2.0''. The code will be released.

    \section{Heuristic for Source Object Selection}
        To approximate the complete object shapes for a target object $A$, a heuristic $\mathcal{H}(A,B)$ is created to evaluate if a source object $B$ covers most of $A$ and can provides points in the regions of $A$'s shape miss.
        The lower the score, the better a object $B$ is for $A$. The heuristic is:
        \begin{adjustwidth}{0pt}{0pt}
            \begin{align}
                \mathcal{H}(A,B) &= \sum_{x \in P_A} \min_{y \in P_B} ||x-y|| - \alpha IoU(\mathcal{D}_A, \mathcal{D}_B) \label{eq:template} \\ \nonumber
                &+ \beta / \big|\{x: x \in Vox(P_B), x \notin Vox(P_A)\}\big|,
            \end{align}
        \end{adjustwidth}
        where $P_A$ and $P_B$ are the object point sets and $D_A$ and $D_B$ are their bounding boxes. 
        \begin{itemize}[leftmargin=10pt]
            \item The first term $\sum_{x \in P_A} \min_{y \in P_B} ||x-y||$ measures if $A$'s points are well covered by $B$'s points (half Chamfer Distance). 
            \item The second term $\alpha IoU(\mathcal{D}_A, \mathcal{D}_B)$ measures the similarity of their bounding box size.
            \item The third term $\beta / \big|\{x: x \in Vox(P_B), x \notin Vox(P_A)\}\big|$ measures the number of extra voxels that B can add to A.
        \end{itemize}

    \section{Training Target and Loss}
	    \subsection{Region Proposal Network (RPN)}
	        We follow the most popular RPN design of anchor-based 3D detection models \cite{lang2019pointpillars,yan2018second,shi2020pv,deng2020voxel}.
	        
	        To generate region proposals, for each class, we first set anchor size as the size of the average 3D objects, and set anchor orientations at $0^{\circ}$ and $90^{\circ}$. Then, we 
    	    We adopt the box encoding for RPN, which is introduced in \cite{lang2019pointpillars,yan2018second}:
            \begin{adjustwidth}{0pt}{0pt}
                \begin{align}
                    x_t = \frac{x_g - x_a}{d_a}, \ \ y_t &= \frac{y_g - y_a}{d_a}, \ \ z_t = \frac{z_g - z_a}{h_a}, \nonumber \\ 
                    where \ \ d_a &= \sqrt{({l_a}^2 + {w_a}^2)} \ ;  \label{eq:encodingsxyz} 
                    \\ 
                    w_t = log(\frac{w_g}{w_a}), \ \ l_t &= log(\frac{l_g}{l_a}), \ \ h_t = log(\frac{h_g}{h_a}), \nonumber \\ 
                    \theta_t &= \theta_g - \theta_a, 
                    \label{eq:encodingswlh}
                \end{align}
            \end{adjustwidth}
            where $x,y,z$ are the box centers, $w,l,h$ and $\theta$ are width, length, height and yaw angle of the boxes, respectively. The subscripts $t,a,g$ denote encoded value, anchor and ground truth, respectively. 
            
            Car (KITTI) or vehicle (WOD) anchors are assigned to ground-truth objects if their IoUs are above 0.6 ($f_g = 1$) or treated as in background if their IoUs are less than 0.45 ($f_g = 0$). The anchors with IoUs in between are ignored in training. For pedestrians and cyclists, the foreground object matching threshold is 0.5 and the background matching threshold is 0.35.
            
            To deal with adversarial angle problem (the orientation at $0$ or $\pi$), we follow \cite{yan2018second} and set the regression loss for orientation as:
            \begin{align}
                \mathcal{L}_{rpn}^{\theta} = SmoothL_1(sin(\theta_p - \theta_t)),
                \label{eq:angleloss}
            \end{align}
            where ``p'' indicates the predicted value. Since the above loss treats opposite directions indistinguishably, a direction classifier is also used and supervised by a softmax loss $\mathcal{L}_{dir}$.
            
            We use Focal Loss \cite{lin2017focal} as the classification loss:
            \begin{align}
                \mathcal{L}_{rpn}^{cls} &= \mathcal{L}_{focal}(p_t) = -\alpha_t(1-p_t)^{\gamma}log(p_t), 
                 \label{loss:focal_appendix} 
                 \\ 
                \text{where} \ \ \ p_t &=
                \begin{cases}
                  p_p & \text{if the box are assigned} \nonumber \\ 
                      & \text{to a foreground object} \ f_g = 1 \nonumber \\ 
                  1 - p_p  & \text{otherwise,} \nonumber
                \end{cases}  \nonumber
            \end{align}
            in which $p_p$ is the predicted foreground score. The parameters of the focal loss are $\alpha=0.25$ and $\gamma=2$.
            The total loss of RPN is:
            \begin{align}
                \mathcal{L}_{rpn} &= \cfrac{1}{N_a} \ \sum_{i}^{N_a} \Big[\mathcal{L}_{rpn}^{cls}
                + \mathfrak{1}(f_g \geq 1)  
                \label{eq:rpn}\\ 
                &\cdot [2 (\mathcal{L}_{rpn}^{\theta} + \mathcal{L}_{rpn}^{reg}) + 0.2 \mathcal{L}_{rpn}^{dir}] \Big], \nonumber
            \end{align}
            where $N_a$ is the number of sampled anchors, $\mathfrak{1}(f_g \geq 1)$ means the regression losses are only applied on the foreground anchors, $\mathcal{L}_{rpn}^{reg}$ is the $SmoothL_1$ regression loss on the encoded x,y,z,w,l,h as described in Eq.\ref{eq:encodingsxyz} and Eq.\ref{eq:encodingswlh} and $\mathcal{L}_{rpn}^{dir}$ is the direction classification loss for predicting the angle bin.
        
	    \subsection{Proposal Refinement}
    	    Following \cite{jiang2018acquisition,li2019gs3d,yan2018second,9018080,shi2020pv,deng2020voxel}, there are two branches in the proposal refinement module, one for class confidence score and another for box regression. 
    	    We follow \cite{jiang2018acquisition,shi2020pv,li2019gs3d,shi2019points} and adopt the 3D IoU weighted RoI confidence training target for each RoI:
    	    \begin{align}
                y_g &=
                \begin{cases}
                    1 &  \ \text{if} \ IoU > 0.75,  \\
                    2 \cdot IoU - 0.5 & \text{if} \ 0.25 <  \ IoU \leq 0.75, \\
                    0  & \text{if} \  \ IoU \leq 0.25, \\
                \end{cases} 
            \end{align}
            
            To conduct regression for bounding box refinement, We adopt the state-of-the-art residual-based box encoding functions:
            \begin{adjustwidth}{0pt}{0pt}
                \begin{align}
                    x_r = \frac{x_g - x_p}{d_p}, \ \ y_r &= \frac{y_g - y_p}{d_p}, \ \ z_r = \frac{z_g - z_p}{h_p}, \nonumber \\ 
                    \ with \ \ d_p &= \sqrt{({l_p}^2 + {w_p}^2)} \ ; \label{eq:encodings1}
                    \\
                    w_r = log(\frac{w_g}{w_p}), \ \ l_r &= log(\frac{l_g}{l_p}), \ \ h_r = log(\frac{h_g}{h_p}), \nonumber \\ 
                    \ \ \theta_r &= \theta_g - \theta_p,
                    \label{eq:encodings2}
                \end{align}
            \end{adjustwidth}
            where $x,y,z$ are the box centers, $w,l,h$ and $\theta$ are the width, length, height and yaw angle of the boxes, respectively. The subscripts $r,p,g$ denote residue, 3D proposal and ground truth, respectively. 
        
            The total proposal refinement loss are: 
            \begin{align}
                \mathcal{L}_{pr} = &\cfrac{1}{N_p} \ \sum_{i}^{N_p} \Big[ \mathcal{L}_{pr}^{cls} + \label{eq:conf_trgts}
                \\
                &\mathfrak{1}(IoU \geq 0.55)\cdot (\mathcal{L}_{pr}^{\theta} + \mathcal{L}_{pr}^{reg})\Big],
                \nonumber
            \end{align}
            where $N_p$ is the number of sampled proposal, $\mathcal{L}_{pr}^{cls}$ is the binary cross entropy loss using the training targets Eq.\ref{eq:conf_trgts}, $\mathfrak{1}(IoU \geq 0.55)$ means we only apply regression loss on positive proposals, $\mathcal{L}_{pr}^{\theta}$ and $\mathcal{L}_{pr}^{reg}$ are similar to the corresponding regression losses in the RPN.
	    
	    \subsection{Total Loss} The total loss is the combination of $\mathcal{L}_{shape}$ introduced in Section 3.1 of the main paper, the RPN loss $\mathcal{L}_{rpn}$ and the proposal refinement loss $\mathcal{L}_{pr}$:
	    \begin{align}
            \mathcal{L}_{total} = 0.3 \mathcal{L}_{shape} + \mathcal{L}_{rpn} + \mathcal{L}_{pr},
            \label{eq:total} 
        \end{align}
	    where we conduct grid search and find the weighting factor of 0.3 helps BtcDet achieve the best results.
            
	\section{Network Architecture}
        In this section, we describe the network architecture of the shape occupancy network, the detection feature backbone network, the region proposal network, and the proposal refinement network of BtcDet.
	    \subsection{Shape Occupancy Network $\Omega$}
\begin{figure*}[!htb]
                \begin{adjustwidth}{-0pt}{-0pt}
                    \centering
                    \includegraphics[width=1.0\linewidth]{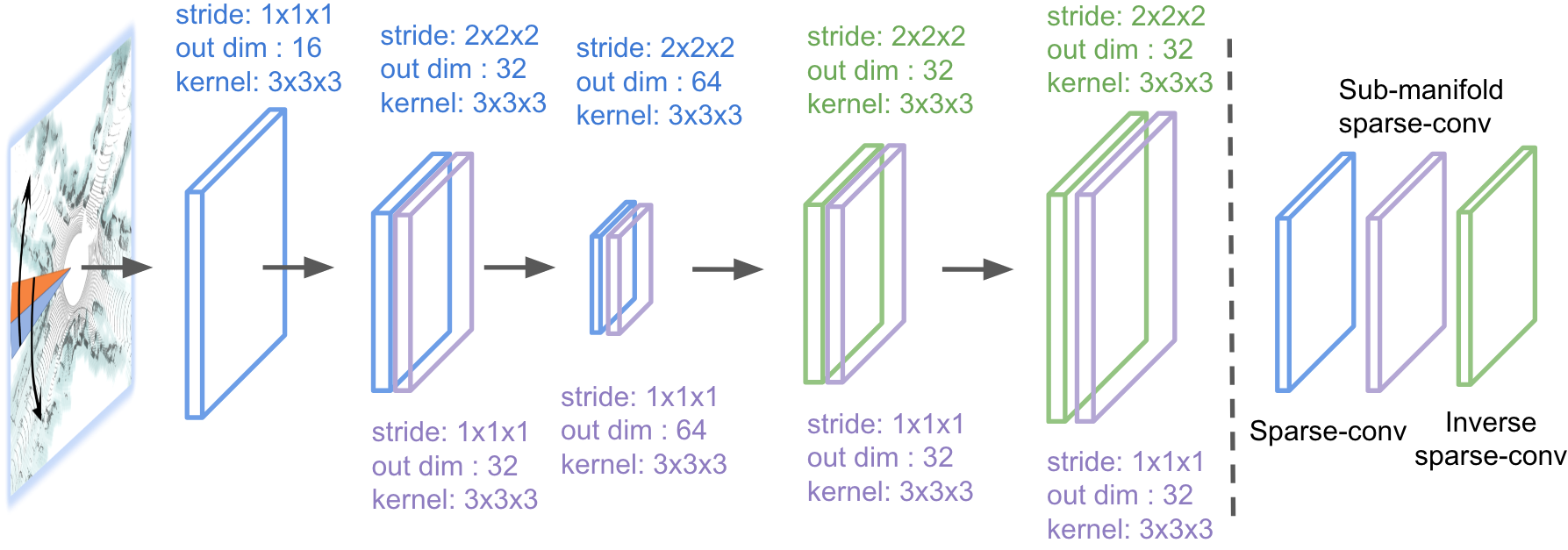}
                    \caption{The network architecture of the shape occupancy network. The blue layers are regular sparse 3D convolutions~\cite{graham2015sparse}, the purple layers are sub-manifold sparse 3D convolutions~\cite{graham2017submanifold}, while the green layers are inverse sparse 3D convolutions (spatial up-sampling).}
                    \label{fig:occ_arch}
                \end{adjustwidth}
            \end{figure*}

    	    As visualized in Figure \ref{fig:occ_arch}, we use a lightweight spherical sparse 3D convolution network with five sparse-conv layers. Two of them are down-sampling and two of them are up-sampling layers, each consists of a regular sparse-conv of stride 2 following by a sub-manifold sparse-conv \cite{graham2017submanifold}. The dimensions of these layers' output features are 16, 32, 64, 32, and 32, respectively.
        \subsection{Detection Backbone Network $\Psi$}
\begin{figure*}[!htb]
                \begin{adjustwidth}{-0pt}{-0pt}
                    \centering
                    \includegraphics[width=1.0\linewidth]{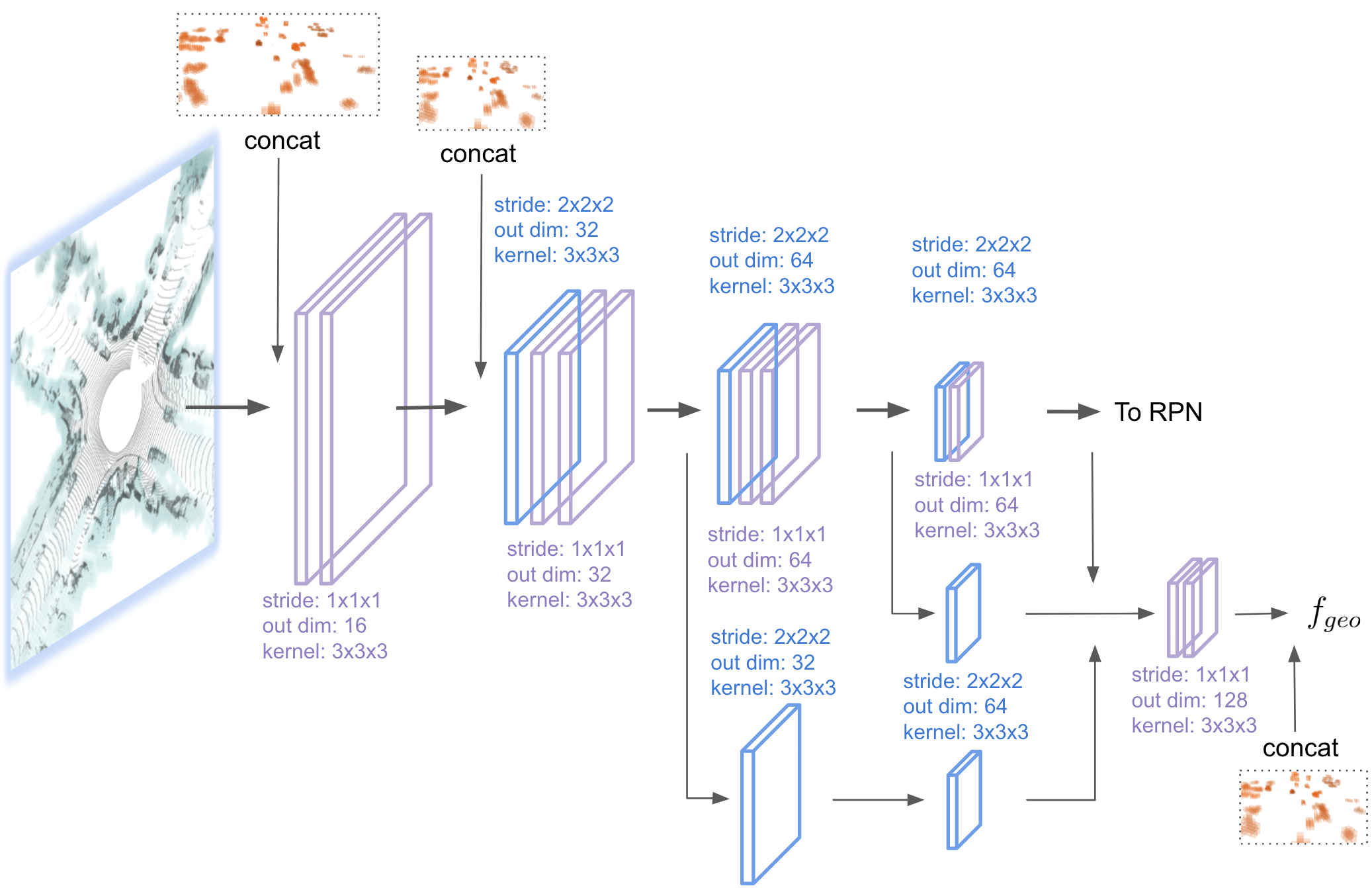}
                \end{adjustwidth}
                \caption{The architecture of the detection feature backbone network. The blue layers are the regular sparse 3D convolutions~\cite{graham2015sparse} and the purple layers are sub-manifold sparse 3D convolutions~\cite{graham2017submanifold}.}
                \label{fig:det_extraction_arch}
            \end{figure*}

            The backbone of the detection feature extraction network follows \cite{shi2020pv,deng2020voxel} but has thinner network layers. The point cloud is voxelized into Cartesian voxels where the features of each occupied voxel are the mean of the points' xyz and features (e.g., intensity). Besides, the sparse probability tensor of object occupancy in the spherical coordinate has been transformed to the  Cartesian coordinate \POS$_\perp$, so that two channels from \POS$_\perp$ can be concatenated with layers of $\Psi$. One channel holds the occupancy probability \POS\ and the other holds the binary code if \POS\ exists in a voxel. As visualized in Figure \ref{fig:det_extraction_arch}, three down-sampling layers down-sample the features to 8$\times$ smaller, which are fed into the region proposal network. The feature maps of the second, the third, and the final layer are further integrated with \POS$_\perp$ to form a local geometric feature $f_{geo}$, which supports the proposal refinement.
            
        \subsection{Region Proposal Network}
            We stack the input features to bird-eye view 2D features. Then, a thinner version of the 2D convolution networks in \cite{lang2019pointpillars,yan2018second} propagates the features and output residues of 2 anchors per class per grid on the output feature maps. Instead of dimensions of 256 as in \cite{shi2020pv,deng2020voxel}, the intermediate feature maps in our 2D convolution networks has feature dimension of 128.
        
        \subsection{Proposal Refinement Network}
            \begin{figure*}[!htb]
                \begin{adjustwidth}{-0pt}{-0pt}
                    \centering
                    \includegraphics[width=1.0\linewidth]{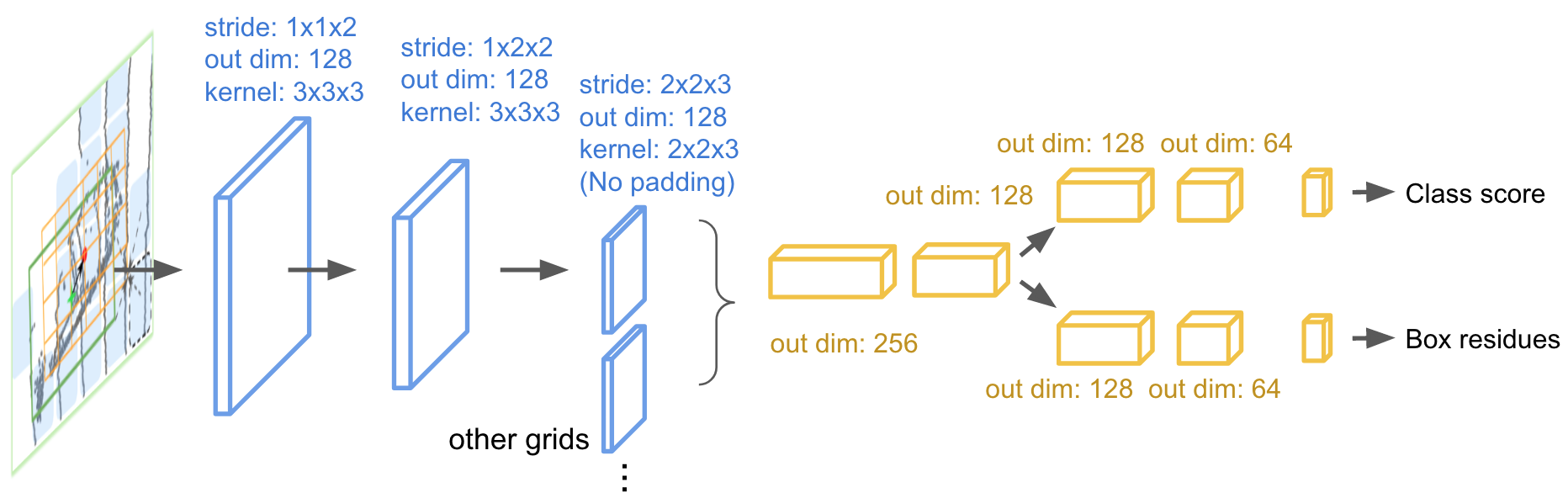}
                    \caption{The architecture of the proposal refinement network in BtcDet's detection pipeline. The blue layers are the regular sparse 3D convolutions~\cite{graham2015sparse} and the yellow layers are fully-connected layers. The stride numbers correspond to Z, Y, X axis. In addition of the above network, We also aggregate the nearby points of the proposal.}
                    \label{fig:refine_arch}
                \end{adjustwidth}
            \end{figure*}
            A local grid of a region proposal has a grid size of (2, 4, 12) along the locally orientated axis of Z, Y, X. As shown in Figure \ref{fig:refine_arch}, the aggregation network of the pooled local geometric features consist of three layers with the strides of (1,1,2), (1,2,2), (2,2,3). The first two layers have zero paddings, while the last layer does not. After that, we send them to several fully connected layers. 
            
            We have $3\times3\times3$ this kind of local grids for each proposal box. The center of a local grid ($x_{grid},y_{grid},z_{grid}$) have a shift away from the proposal center ($x_p,y_p,z_p$) by distances ($\Delta_x \in \{\pm \lambda, 0\} \times w_p, \Delta_y = \{\pm \lambda, 0\} \times l_p, \Delta_z = \{\pm \lambda, 0\} \times h_p$), where $w_p,l_p,h_p$ is the width, length and height of the proposal box. We find $\lambda = 0.25$ achieves the best results. The proposal refinement network aggregates results from all these shifted local grids and outputs the residues regression and the class confidence score, which lead to the final bounding box predictions.
            
    \section{Occupancy Estimation of Complete Object Shapes}
        \label{sec:occupancy}
        We show the evaluation of the occupancy estimation in Table \ref{tb:aa}. The results are averaged among all voxels in the regions of \ROCRSM. A prediction is considered positive if \POS\ $ > $ threshold. The metrics we evaluate are precision, recall, F1 score, accuracy, and object coverage. The object coverage is the percentage of all bounding boxes that contain at least one positive voxel (\POS\ $ > $ threshold). We show the measures on three thresholds of $0.3, 0.5$, and $0.7$. The accuracy results under all thresholds are very high (>99\%) since the classes are extremely imbalanced. However, no matter under which threshold, we can achieve relatively high object coverage, which means the estimation is faithful enough for RPN and other downstream networks to rely on.
        
        \begin{table*}[htb!]
            \centering
    		\begin{tabular}{c|ccccc}
                \hline
                 Threshold & Precision &  Recall  & F1 Score  &  Accuracy  &  Object Coverage  \\ \hline
                 0.3 & 39.8 \%  & 94.2 \%   & 55.1 \% & 99.6 \% & 95.6 \%  \\
                 0.5 & 60.0 \%  & 81.9 \%   & 68.3 \% & 99.7 \% & 94.0 \%  \\ 
                 0.7 & 80.9 \%  & 47.0 \%   & 58.6 \% & 99.8 \% & 84.0 \%  \\\hline
            \end{tabular}
    		\caption{The results of occupancy estimation. The model is trained on the KITTI's train split and then evaluated on the KITTI's val split. The precision, recall, F1 score, accuracy and object coverage are evaluated by setting \POS\ $ > $ the corresponding thresholds. Those metrics are evaluated considering all voxels in \ROCRSM. The object coverage is the the percentage of all bounding boxes that contain at least one voxel that is predicted as positive.}
    		\label{tb:aa}
    	\end{table*}

    \section{More Comparison Results on the KITTI Test Set} 
        We show more results of comparisons between BtcDet and other state-of-the-art detectors in Table \ref{tb:more_kitti_test}. Because the average point number in pedestrians is smaller than other objects, the shape estimation is sensitive to a few observed points. Therefore, if the point number distribution of pedestrians in the test split is different, our model may not be able to provide an accurate shape occupancy estimation. As a result, BtcDet's pedestrian detection on KITTI's test split does not perform as well as on KITTI's val split. We consider improving the results with this situation in our future works.
        \begin{table*}[!hbt]
            \begin{adjustwidth}{0pt}{0pt}
                \setlength\extrarowheight{1pt}
                \setlength\tabcolsep{5pt}
                \centering
                {\small
                \begin{tabular}{l|cccc|cccc|cccc}
                \hline
                \multirow{2}{*}{\normalsize Method}  & \multicolumn{4}{c|}{\normalsize Ped. 3D $AP_{R40}$} & \multicolumn{4}{c|}{\normalsize Car 3D $AP_{R40}$}                      & \multicolumn{4}{c}{\normalsize Cyc. 3D $AP_{R40}$}  \\      & Easy           & Mod.           & Hard           & mAP            & Easy           & Mod.           & Hard           & mAP            & Easy           & Mod.           & Hard           & mAP            \\ \hline
                \small F-PointNet \cite{qi2018frustum}               &  50.53    & 42.15             & 38.08     &    43.59     & 82.19          & 69.79          & 60.59          & 70.86          & 72.27          & 56.12          & 49.01          & 59.13          \\
                \small AVOD-FPN \cite{ku2018joint}                  & 50.46                  & 42.27 & 39.04    &    43.92    & 83.07          & 71.76          & 65.73          & 73.52          & 63.76          & 50.55          & 44.93          & 53.08          \\
                \small F-ConvNet \cite{wang2019frustum}                & 52.16    &      43.38    &   38.8 &   44.78    & 87.36          & 76.39          & 66.69          & 76.81          & 81.98          & 65.07          & 56.54          & 67.86          \\
                \small Uber-MMF \cite{liang2019multi}              & - & - & - & -                        & 88.40          & 77.43          & 70.22          & 78.68          & -              & -              &      -          & -              \\
                \small EPNet \cite{huang2020epnet}                    & 52.79   &   44.38   &   41.29                   & 46.15                 & 89.81          & 79.28          & 74.59          & 81.23          & -              & -              & -              & -              \\
                \small CLOCsPVCas \cite{pang2020clocs}             &    -   &   -   &   -   &   -        & 88.94          & 80.67          & 77.15          & 82.25          & -              & -              & -              & -              \\
                \small 3D-CVF \cite{yoo20203d}                   &  -   &   -   &   -   &   - & 89.20          & 80.05          & 73.11          & 80.79          & -              & -              & -              & -              \\ \hline
                \small SECOND \cite{yan2018second}                   &  48.73    &  40.57   &  37.77           &       42.36           & 83.34          & 72.55          & 65.82          & 73.90          & 71.33          & 52.08          & 45.83          & 56.41          \\
                \small PointPillars \cite{lang2019pointpillars}  & 51.45   &  41.92 &	38.89 & 	44.09      & 82.58          & 74.31          & 68.99          & 75.29          & 77.10          & 58.65          & 51.92          & 62.56          \\
                \small PointRCNN \cite{shi2019pointrcnn}                & 47.98	&  39.37 & 	36.01 & 	41.12    & 86.96          & 76.50          & 71.39          & 78.28          & 74.96          & 58.82          & 52.53          & 62.10          \\
                \small 3D Iou Loss \cite{zhou2019iou}              &  -   &   -   &   -   &   -                     & 86.16          & 75.64          & 70.70          & 77.50          & -              & -              & -              & -              \\
                \small Fast PointRCNN \cite{chen2019fast}          &  -   &   -   &   -   &   -            & 85.29          & 77.40          & 70.24          & 77.64          & -              & -              & -              & -              \\
                \small STD \cite{yang2019std}                      &  53.29	& 42.47	& 38.35	& 44.70 & 87.95          & 79.71          & 75.09          & 80.92          & 78.69          & 61.59          & 55.30          & 65.19          \\
                \small SegVoxelNet \cite{yi2020segvoxelnet}              & -   &   -   &   -   &   -           & 86.04          & 76.13          & 70.76          & 77.64          & -              & -              & -              & -              \\
                \small VoxelFPN \cite{kuang2020voxel}             &  -   &   -   &   -   &   -          & 85.63          & 76.70          & 69.44          & 77.26          & -              & -              & -              & -              \\
                \small HotSpotNet \cite{chen2020object}               &  53.10	& \textbf{45.37} &	\textbf{41.47} &	\textbf{46.65}     & 87.60          & 78.31          & 73.34          & 79.75          & 82.59          & 65.95          & 59.00          & 69.18          \\
                \small PartA$^2$ \cite{9018080} & 53.10	&   43.35 	&   40.06	   &   45.50  & 87.81          & 78.49          & 73.51          & 79.94          & 79.17          & 63.52          & 56.93          & 66.54          \\
                \small SERCNN \cite{Zhou_2020_CVPR}                   &  -   &   -   &   -   &   -     & 87.74          & 78.96          & 74.14          & 80.28          & -              & -              & -              & -              \\
                \small Point-GNN \cite{shi2020point}                &  51.92 &	43.77 &	40.14 &	45.28      & 88.33          & 79.47          & 72.29          & 80.03          & 78.60          & 63.48          & 57.08          & 66.39          \\
                \small 3DSSD \cite{yang20203dssd}                    & 50.64  & 	43.09	  & 39.65    &  44.46         & 88.36          & 79.57          & 74.55          & 80.83          & 82.48          & 64.10          & 56.90          & 67.83          \\
                \small SA-SSD \cite{he2020sassd}                   &    -   &   -   &   -   &   -     & 88.75          & 79.79          & 74.16          & 80.90          & -              & -              & -              & -              \\
                \small Asso-3Ddet \cite{du2020associate}               &  -   &   -   &   -   &   -      & 85.99          & 77.40          & 70.53          & 77.97          & -              & -              & -              & -              \\
                \small PV-RCNN \cite{shi2020pv}                   & 52.17  & 	43.29  & 	40.29  & 	45.25     & 90.25          & 81.43          & 76.82          & 82.83          & 78.60          & 63.71          & 57.65          & 66.65          \\
                \small Voxel R-CNN \cite{deng2020voxel}              &  -   &   -   &   -   &   -       & \textbf{90.90} & 81.62          & 77.06          & 83.19          & -              & -              & -              & -              \\
                \small CIA-SSD \cite{zheng2020ciassd}                  & -   &   -   &   -   &   -          & 89.59          & 80.28          & 72.87          & 80.91          & -              & -              & -              & -              \\
                \small TANet \cite{liu2020tanet}                    &   \textbf{53.72}   &  	44.34	&   40.49	&   46.18     & 83.81          & 75.38          & 67.66          & 75.62          & 73.84          & 59.86          & 53.46          & 62.39          \\ \hline
                BtcDet (Ours)            &   47.80 & 41.63 & 39.30  &  42.91         & 90.64          & \textbf{82.86} & \textbf{78.09} & \textbf{83.86} & \textbf{82.81} & \textbf{68.68} & \textbf{61.81} & \textbf{71.10} 
                    \\ 
                \hline
                \end{tabular}
                }
                \captionsetup{aboveskip = 5pt}
                \captionsetup{belowskip = -10pt}
        	    \caption{Comparison results for all three classes of objects on the KITTI \textit{test} set, evaluated by the 3D Average Precision (AP) of 40 recall thresholds (R40) on the KITTI server. The mAPs are averaged over the APs of easy, moderate, and hard objects.}
    	        \label{tb:more_kitti_test}
    	    \end{adjustwidth}
		\end{table*}
		
    \section{More Visualization for the Complete Object Shape Approximation}
\begin{figure*}[!htb]
            \vspace{-20pt}
            \begin{adjustwidth}{-10pt}{-10pt}
                \begin{subfigure}{0.5\linewidth}
                    \flushleft
                    \includegraphics[width=0.95\linewidth]{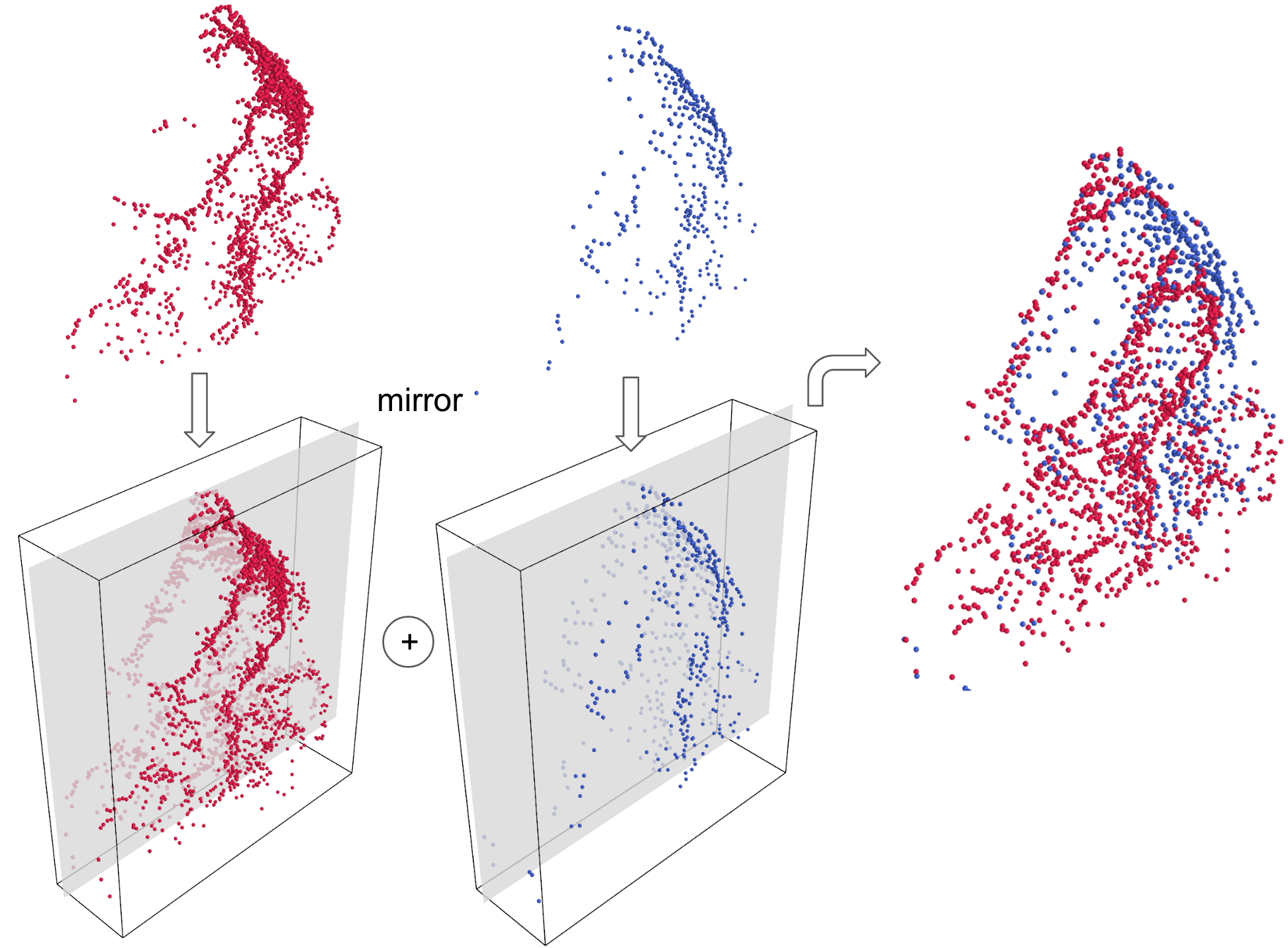}
                \end{subfigure}
                \begin{subfigure}{0.5\linewidth}
                    \flushright
                    \includegraphics[width=0.95\linewidth]{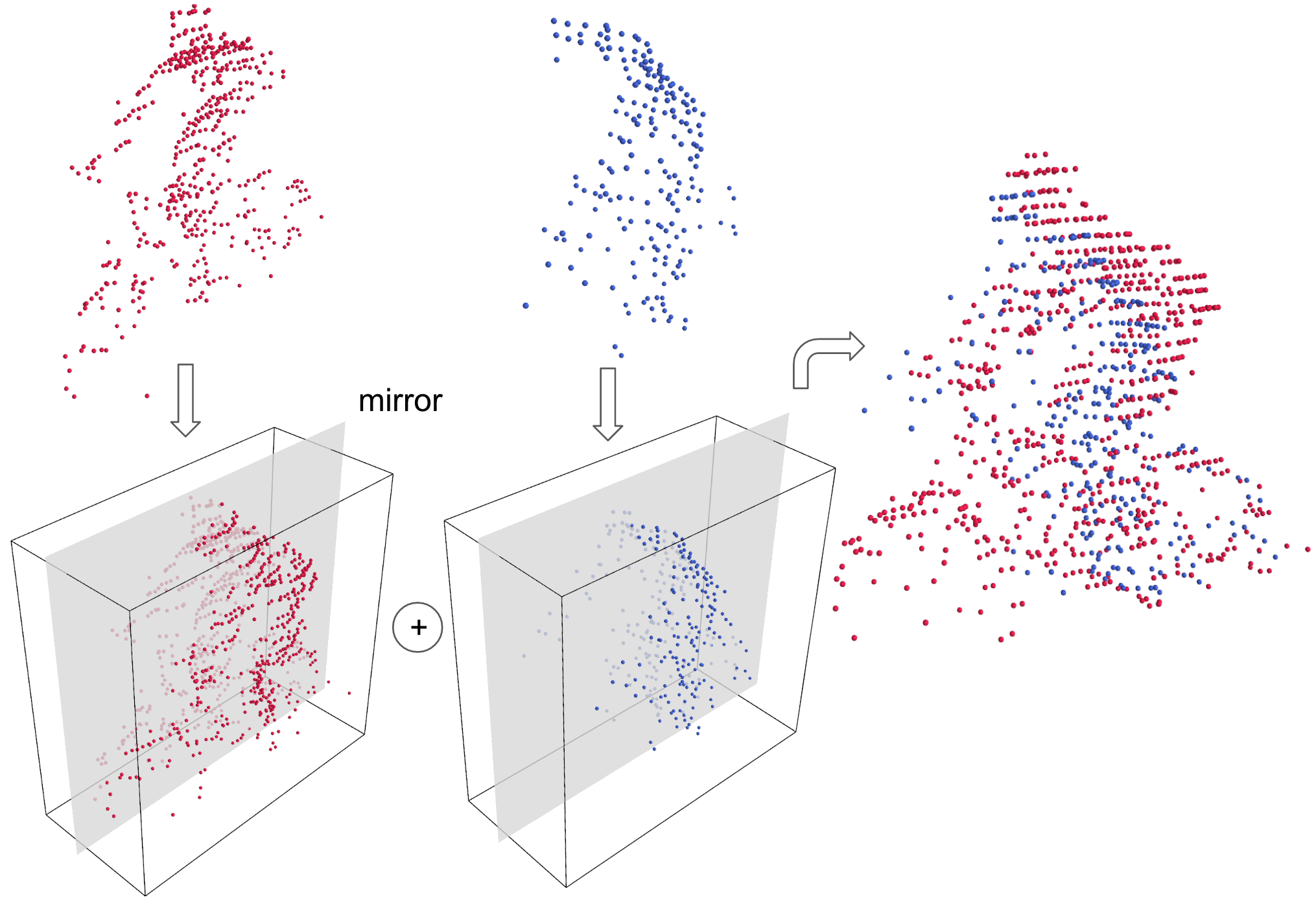}
                \end{subfigure}
                \begin{subfigure}{0.5\linewidth}
                    \flushleft
                    \includegraphics[width=0.95\linewidth]{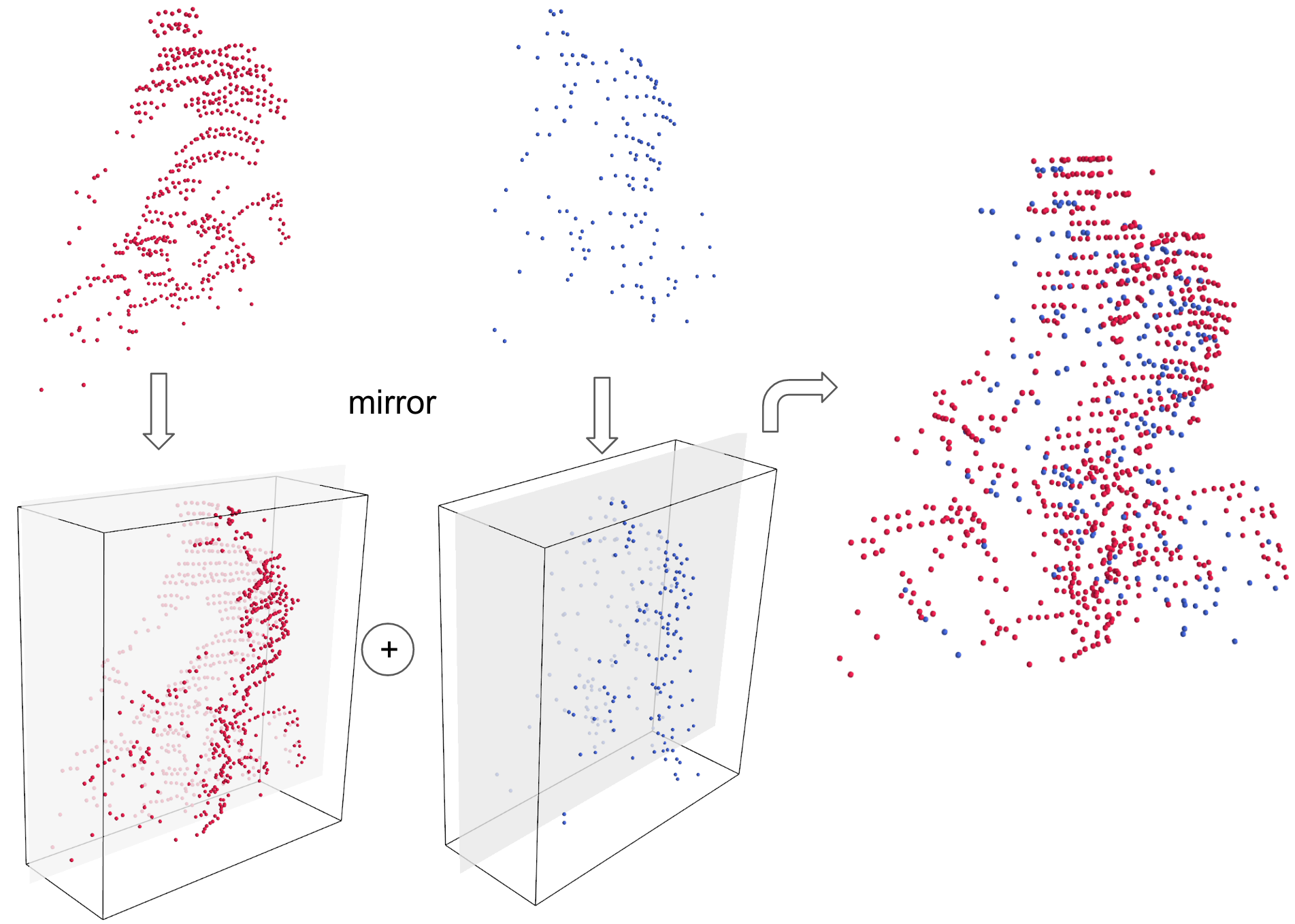}
                \end{subfigure}
                \begin{subfigure}{0.5\linewidth}
                    \flushright
                    \includegraphics[width=0.95\linewidth]{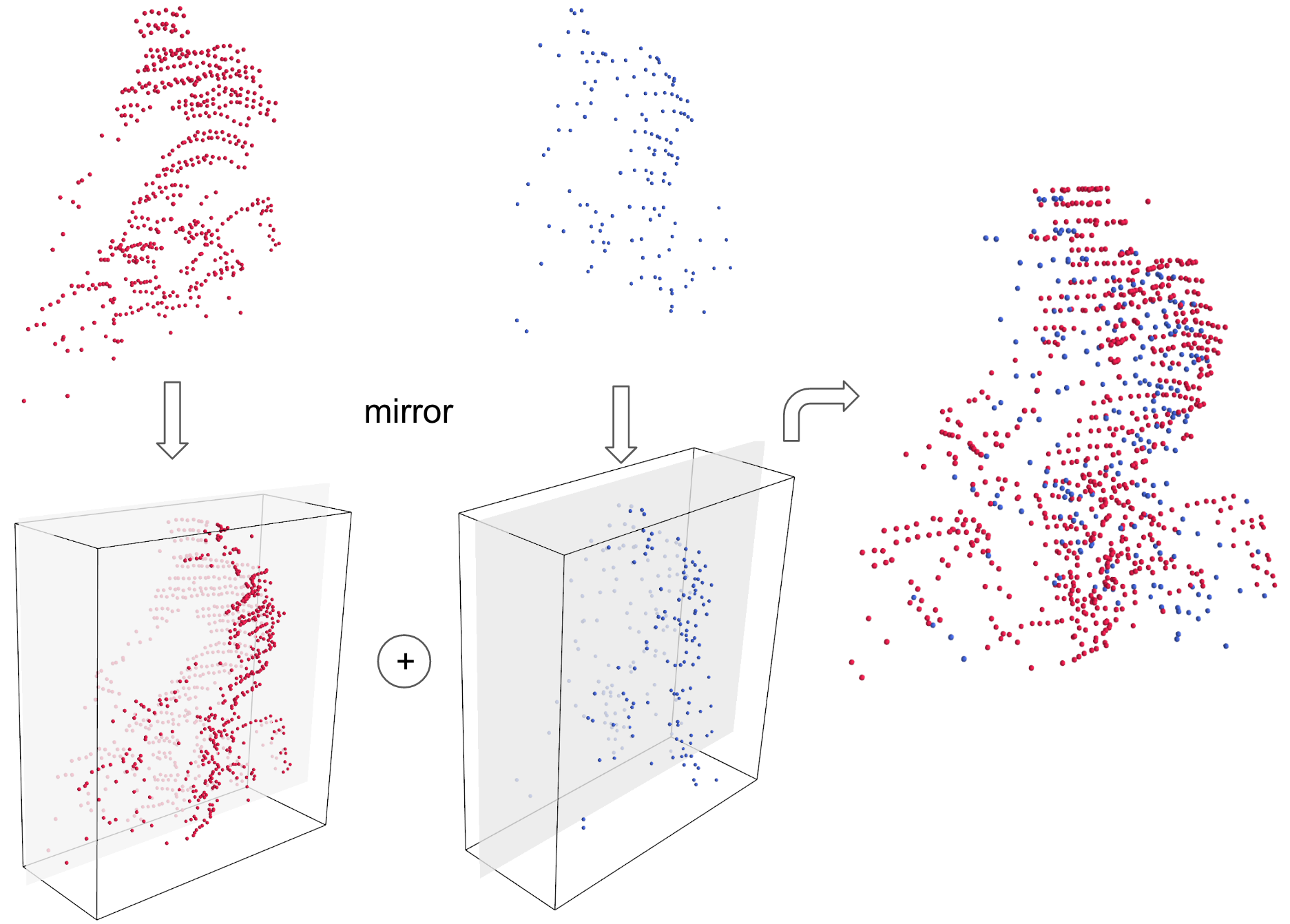}
                \end{subfigure}
                \begin{subfigure}{0.5\linewidth}
                    \flushleft
                    \includegraphics[width=0.95\linewidth]{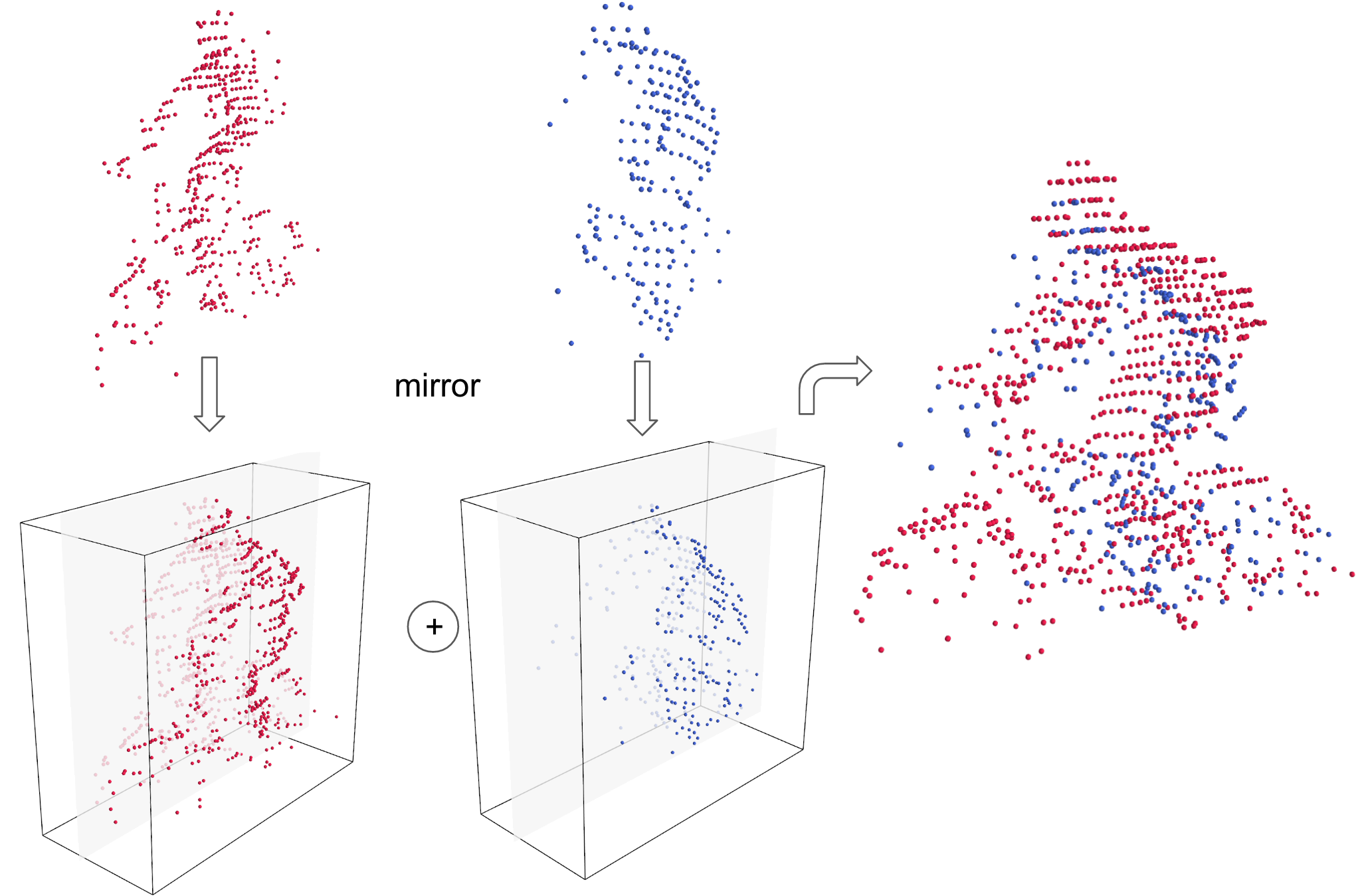}
                \end{subfigure}
                \begin{subfigure}{0.5\linewidth}
                    \flushleft
                    \includegraphics[width=0.95\linewidth]{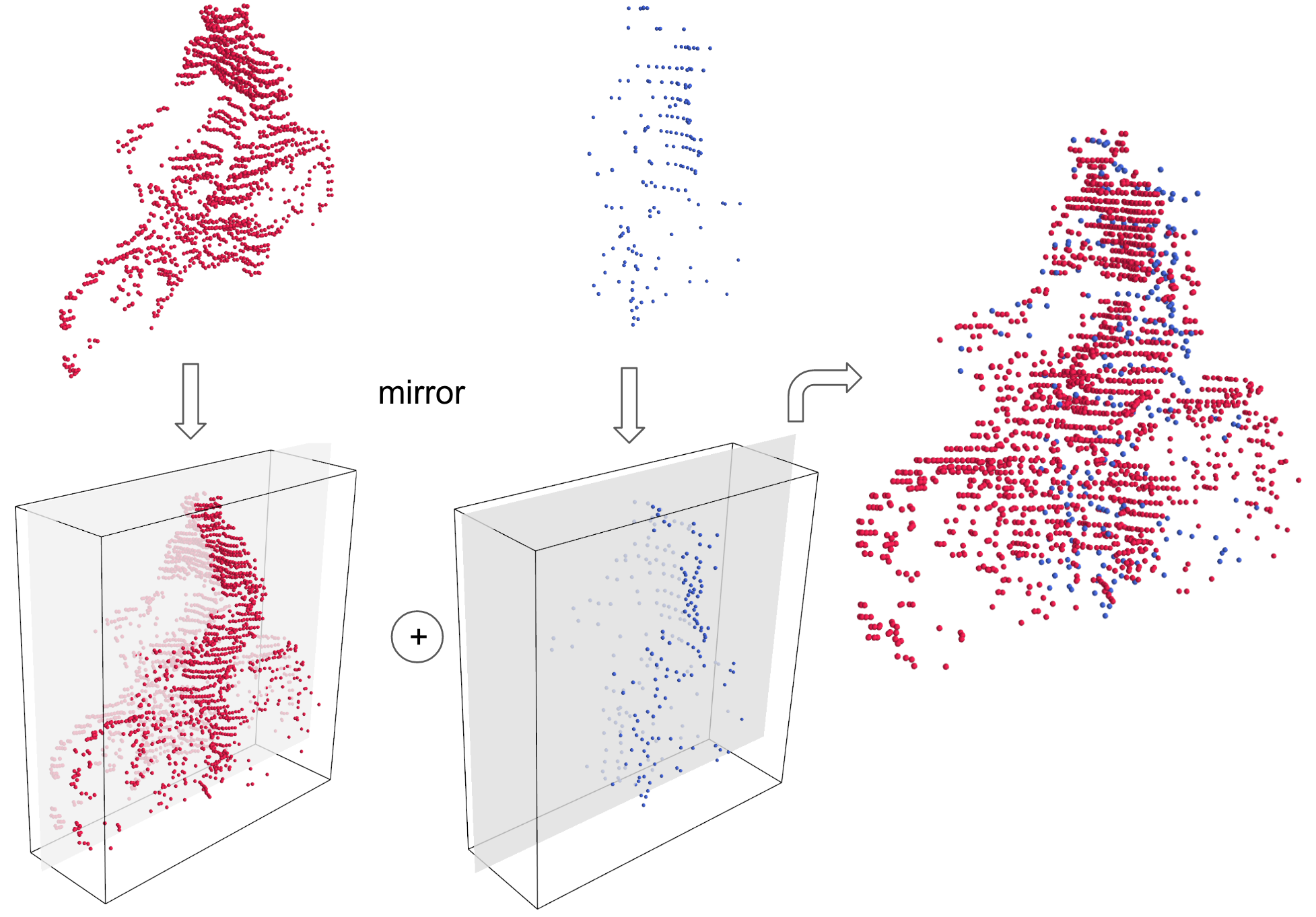}
                \end{subfigure}
            \end{adjustwidth}
            \captionsetup{aboveskip = 20pt}
            \caption{The assembly process to approximate the complete object shapes for cyclists on KITTI \cite{geiger2013vision}. The red points are from the source objects, the blue points are from target objects, the complete shape of the target objects are approximated by borrowing the points from the selected source objects.}
            \label{fig:byc_asb}
        \end{figure*}

\begin{figure*}[!htb]
            \begin{adjustwidth}{-10pt}{-10pt}
                \begin{subfigure}{0.5\linewidth}
                    \flushleft
                    \includegraphics[width=0.9\linewidth]{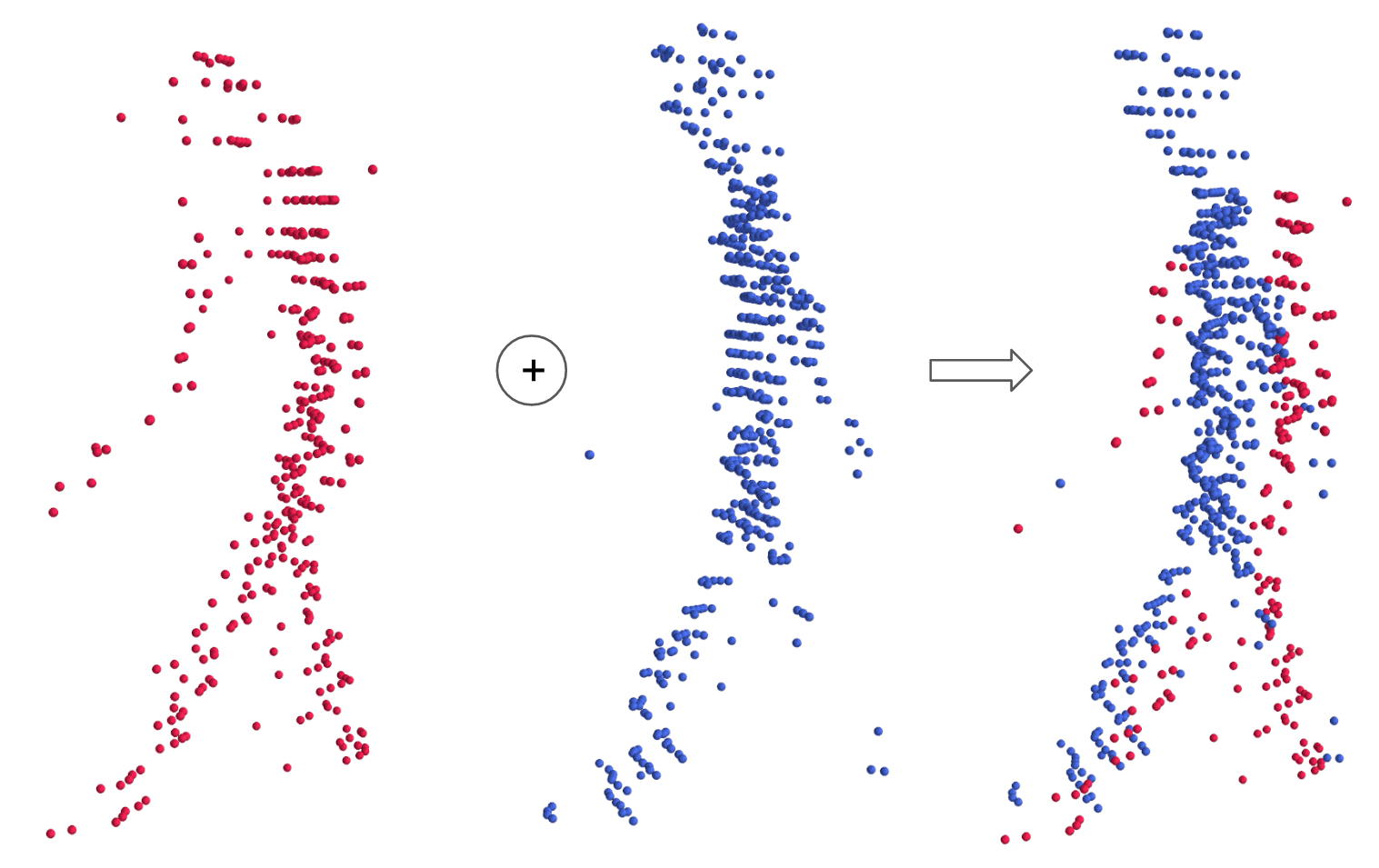}
                \end{subfigure}
                \begin{subfigure}{0.5\linewidth}
                    \flushright
                    \includegraphics[width=0.9\linewidth]{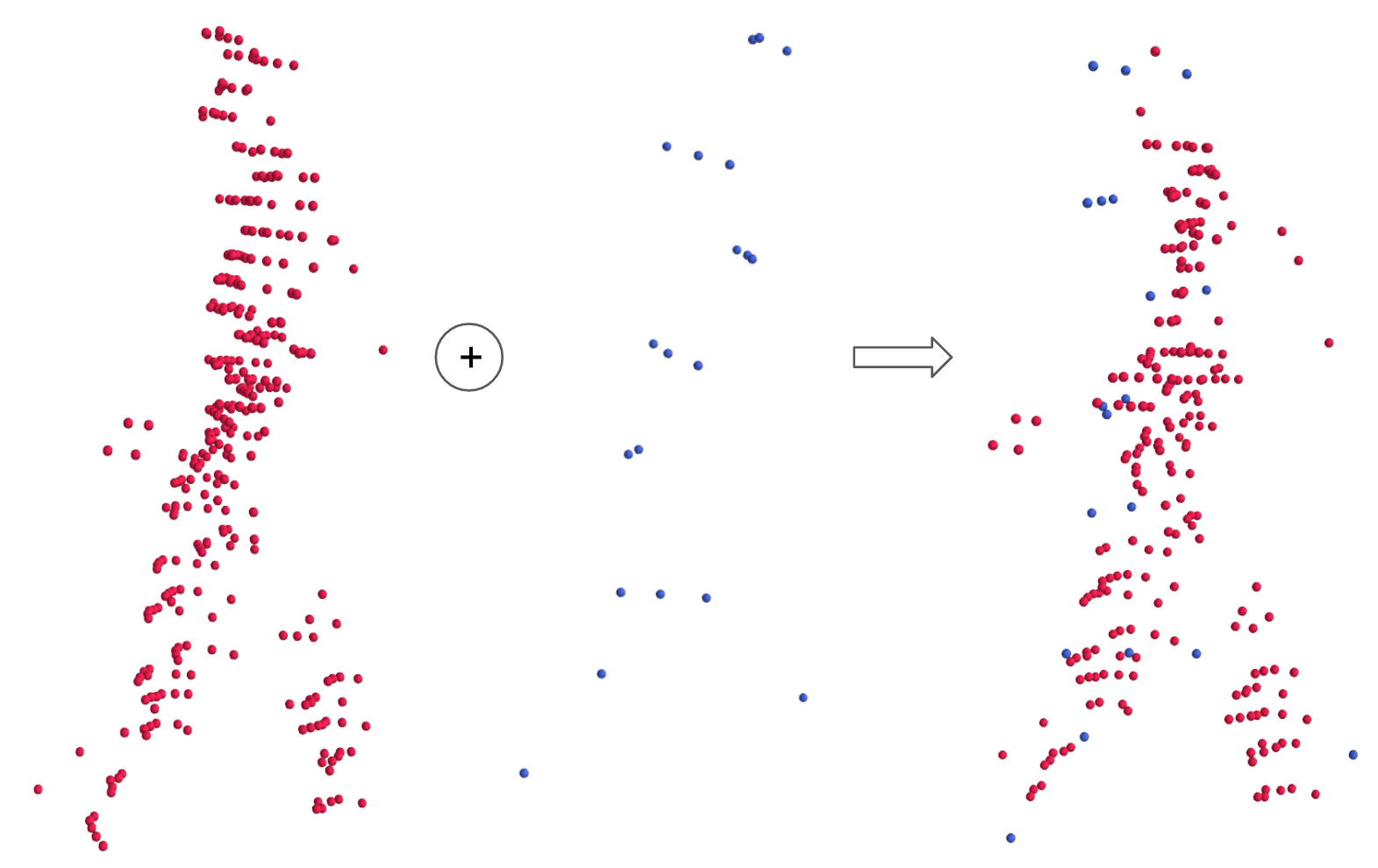}
                \end{subfigure}
                \begin{subfigure}{0.5\linewidth}
                    \flushleft
                    \includegraphics[width=0.9\linewidth]{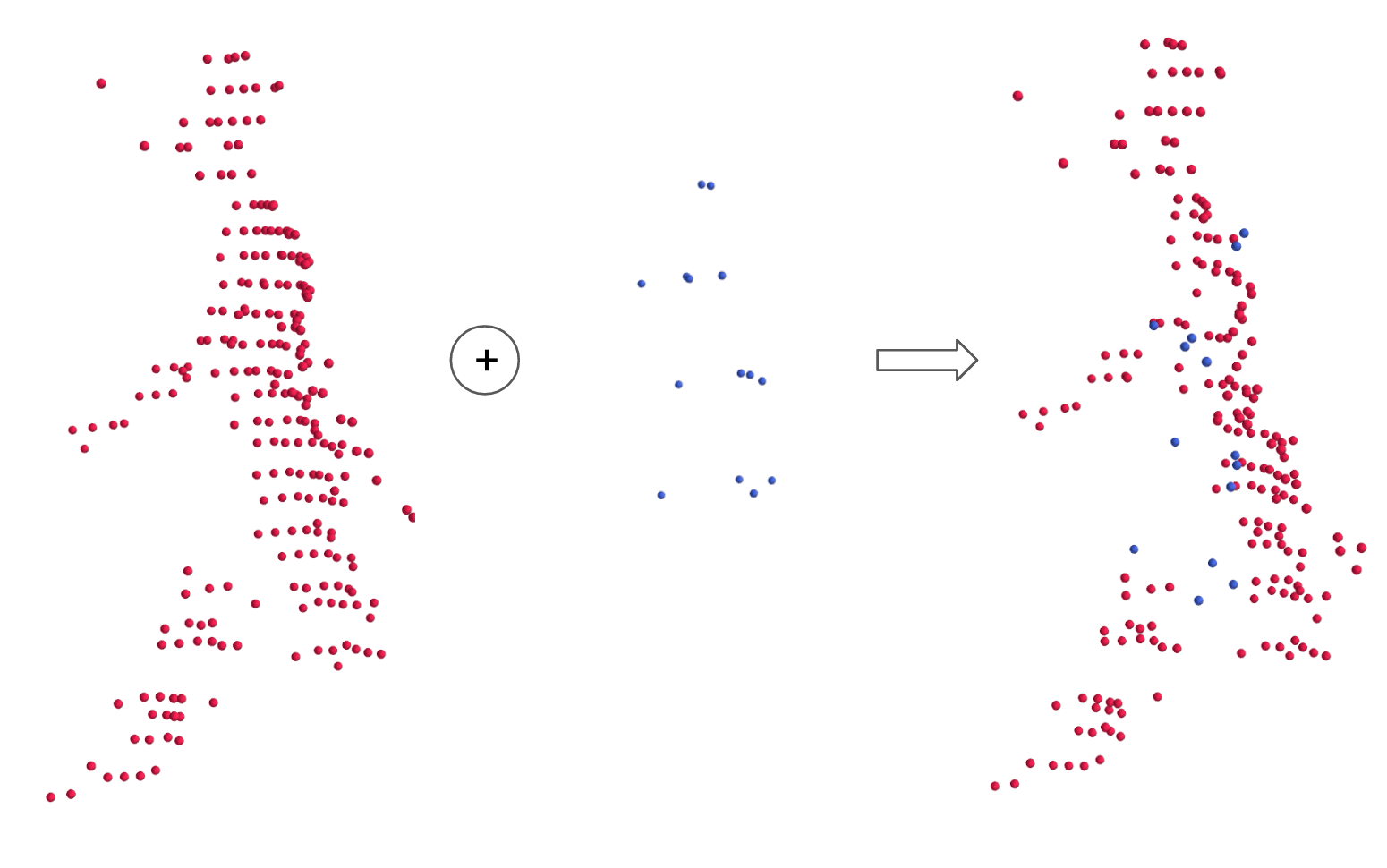}
                \end{subfigure}
                \begin{subfigure}{0.5\linewidth}
                    \flushright
                    \includegraphics[width=0.9\linewidth]{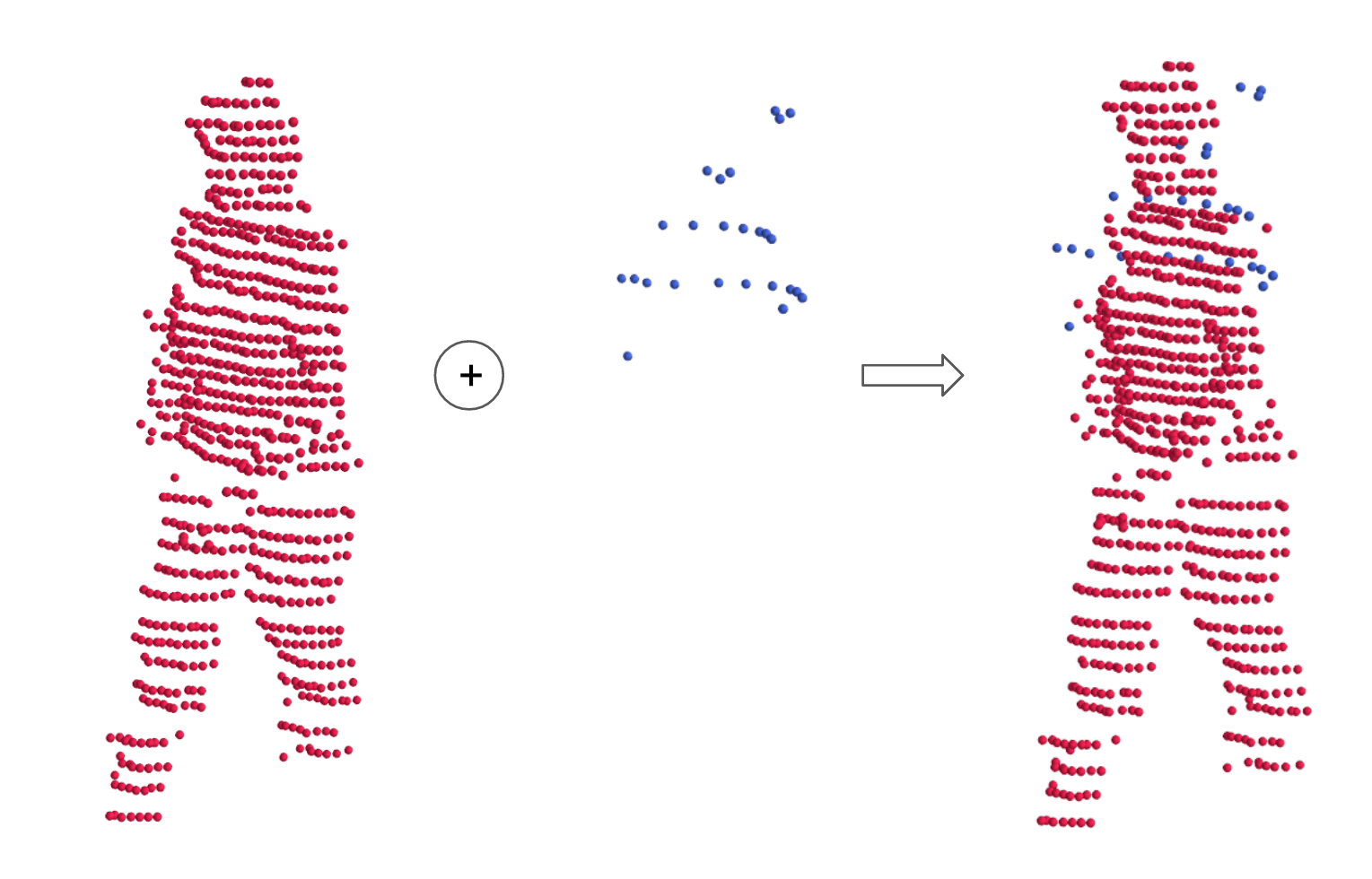}
                \end{subfigure}
                \begin{subfigure}{0.5\linewidth}
                    \flushleft
                    \includegraphics[width=0.9\linewidth]{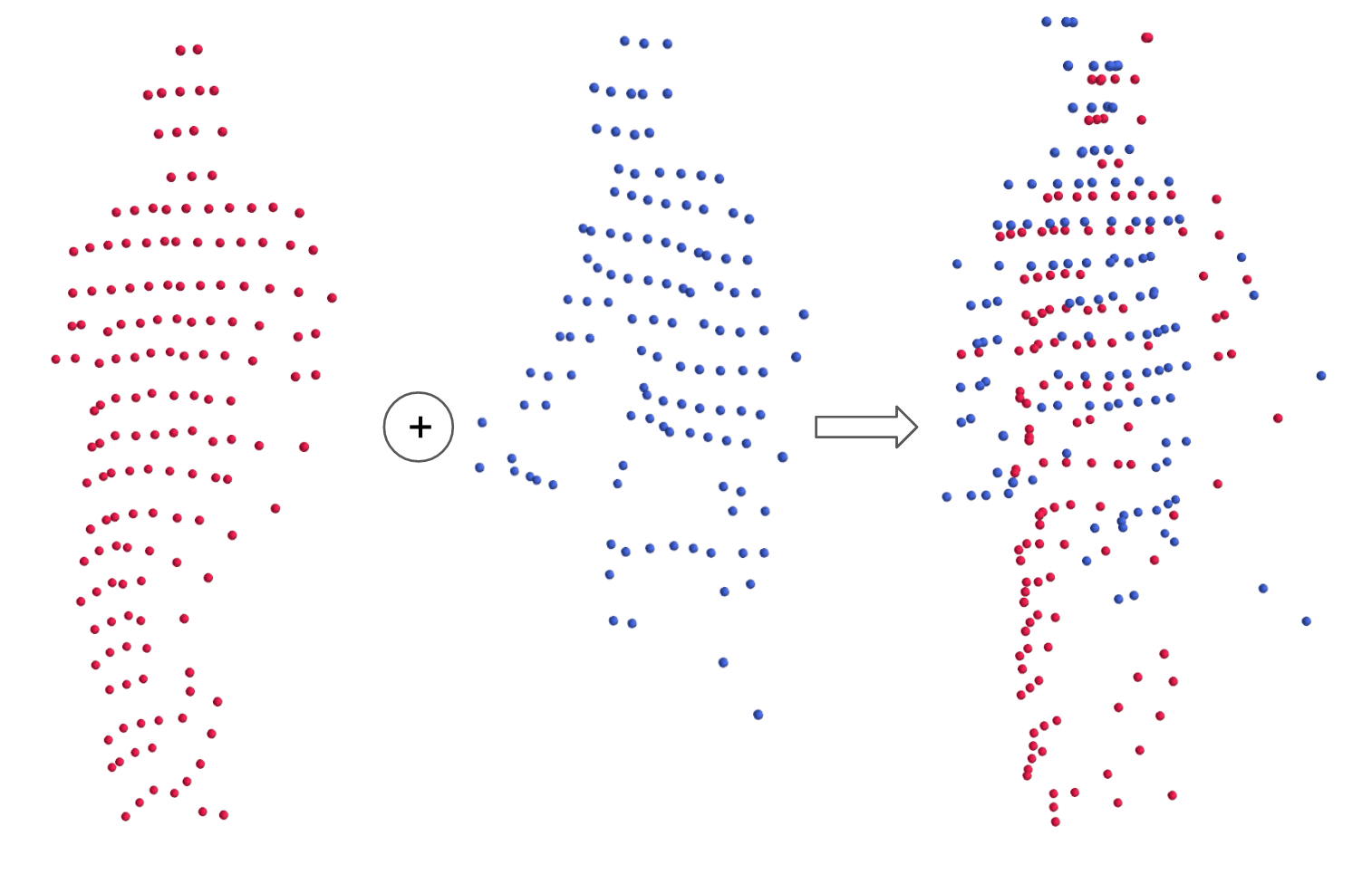}
                \end{subfigure}
                \begin{subfigure}{0.5\linewidth}
                    \flushright
                    \includegraphics[width=0.9\linewidth]{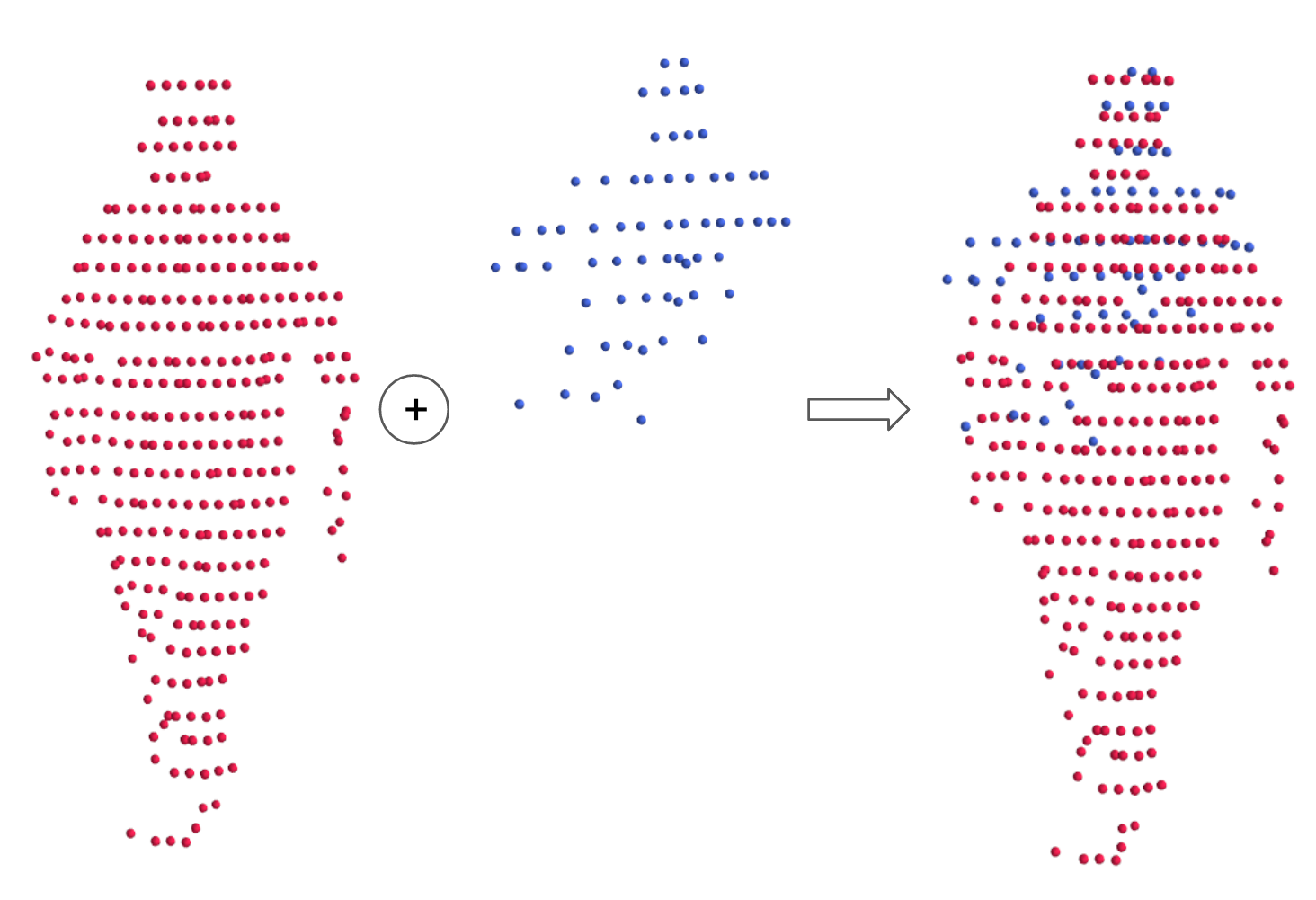}
                \end{subfigure}
                \begin{subfigure}{0.5\linewidth}
                    \flushleft
                    \includegraphics[width=0.9\linewidth]{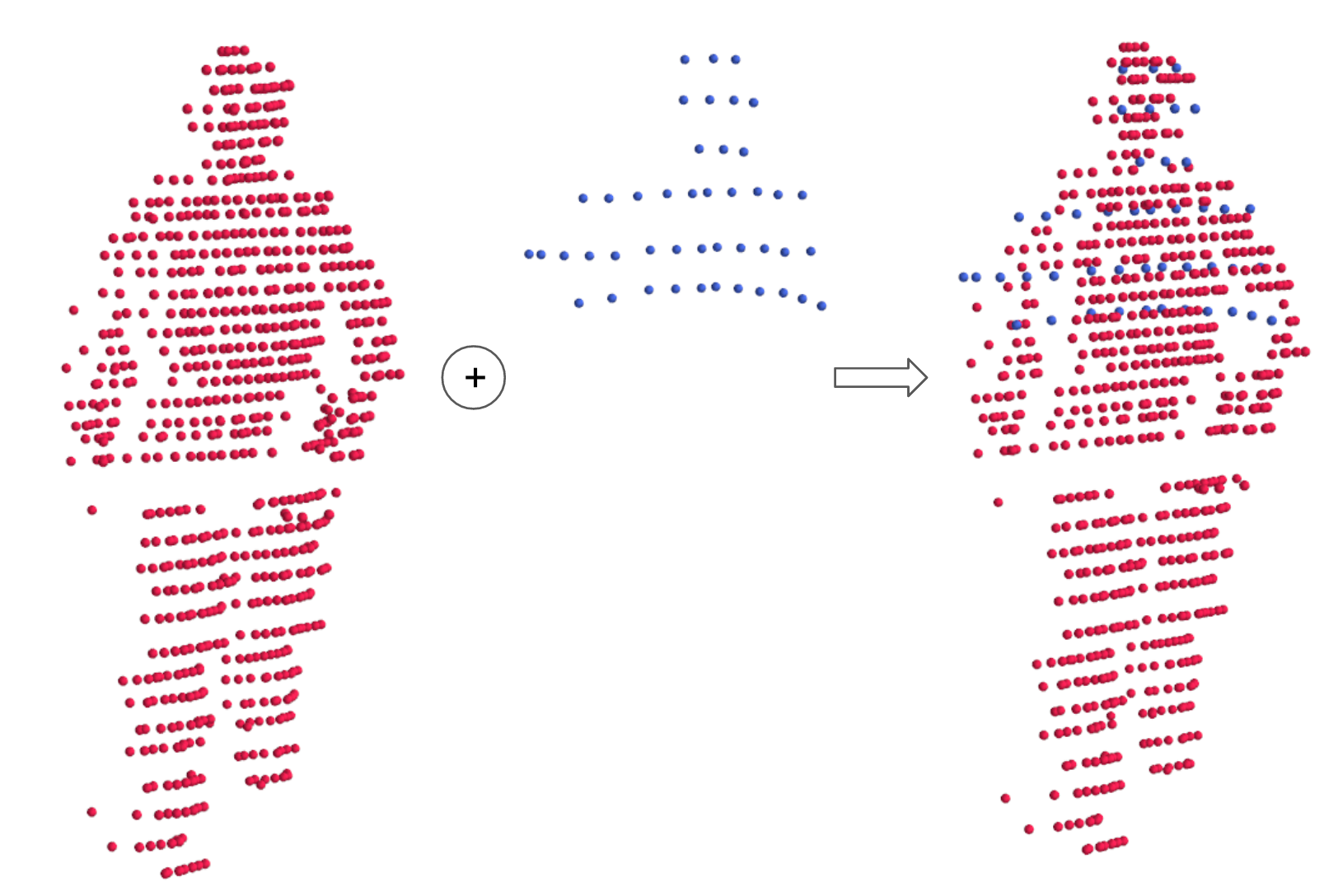}
                \end{subfigure}
                \begin{subfigure}{0.5\linewidth}
                    \flushleft
                    \includegraphics[width=0.9\linewidth]{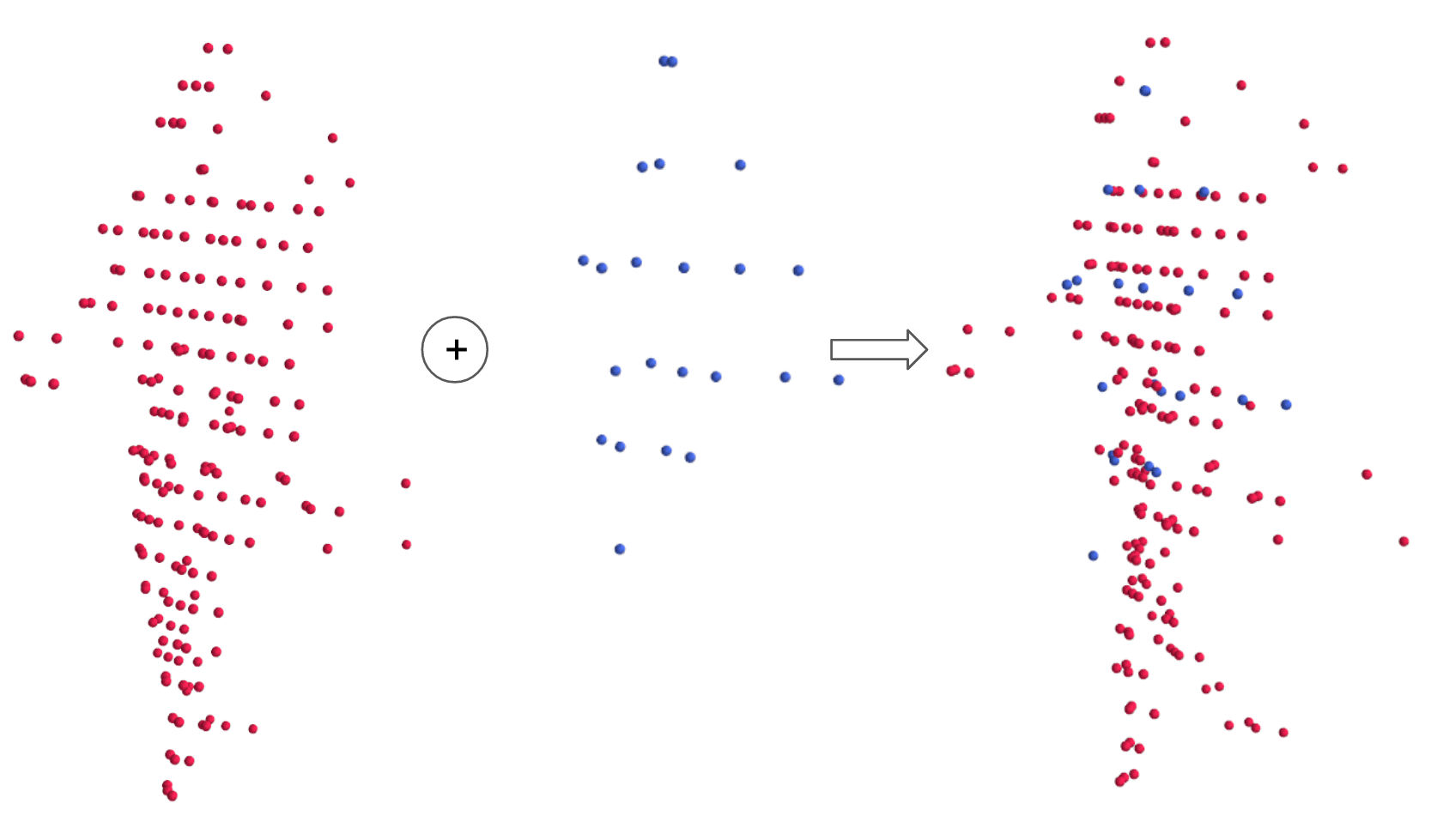}
                \end{subfigure}
            \end{adjustwidth}
            \caption{The assembly process to approximate the complete shapes for pedestrians on KITTI \cite{geiger2013vision}. The red points are from the source objects, the blue points are from target objects, the complete shape of the target objects are approximated by borrowing the points from the selected source objects (red).}
            \label{fig:ped_asb1}
        \end{figure*}

\begin{figure*}[!htb]
            \begin{adjustwidth}{-10pt}{-10pt}
                \begin{subfigure}{0.5\linewidth}
                    \flushleft
                    \includegraphics[width=0.9\linewidth]{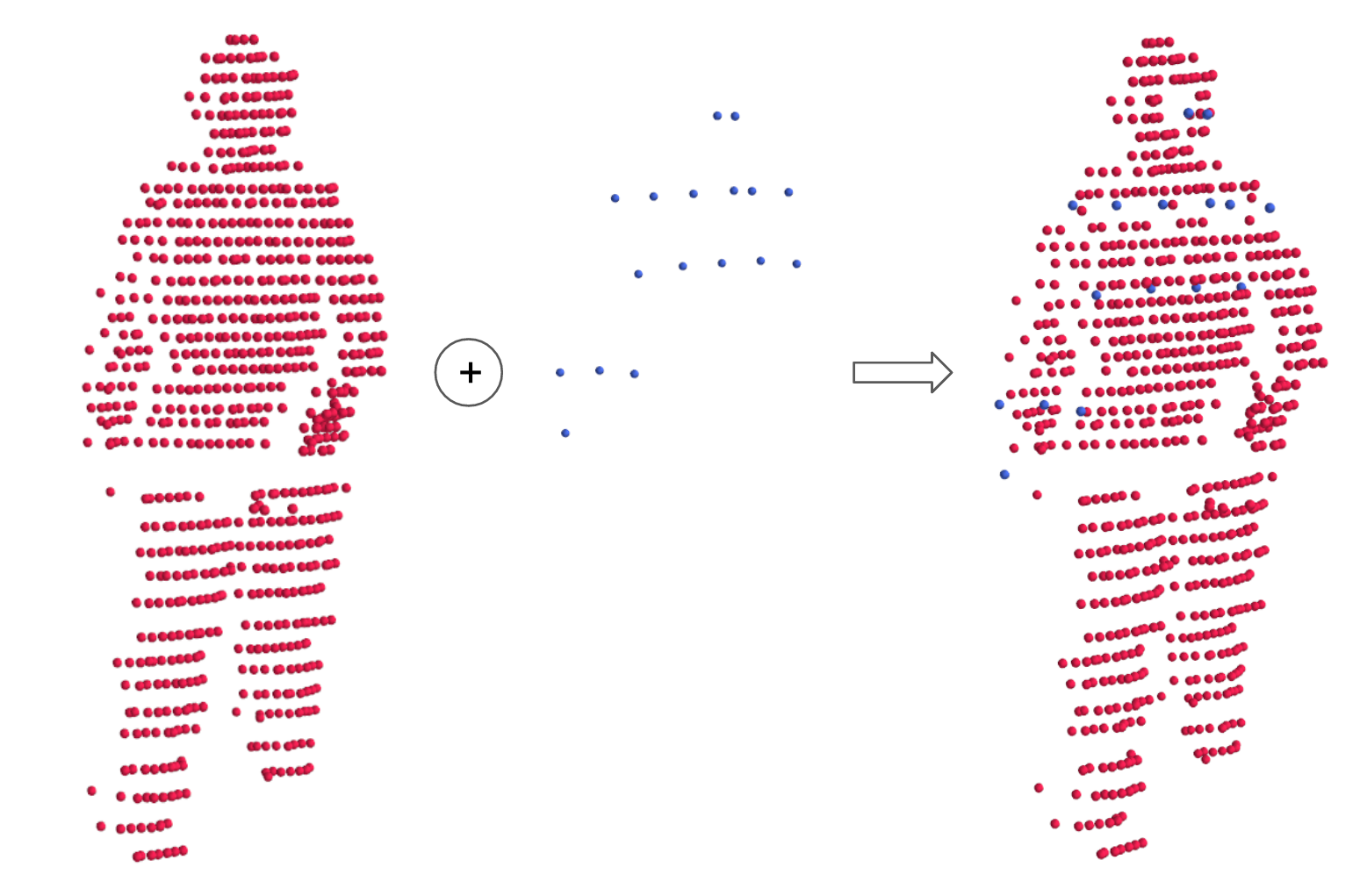}
                \end{subfigure}
                \begin{subfigure}{0.5\linewidth}
                    \flushright
                    \includegraphics[width=0.9\linewidth]{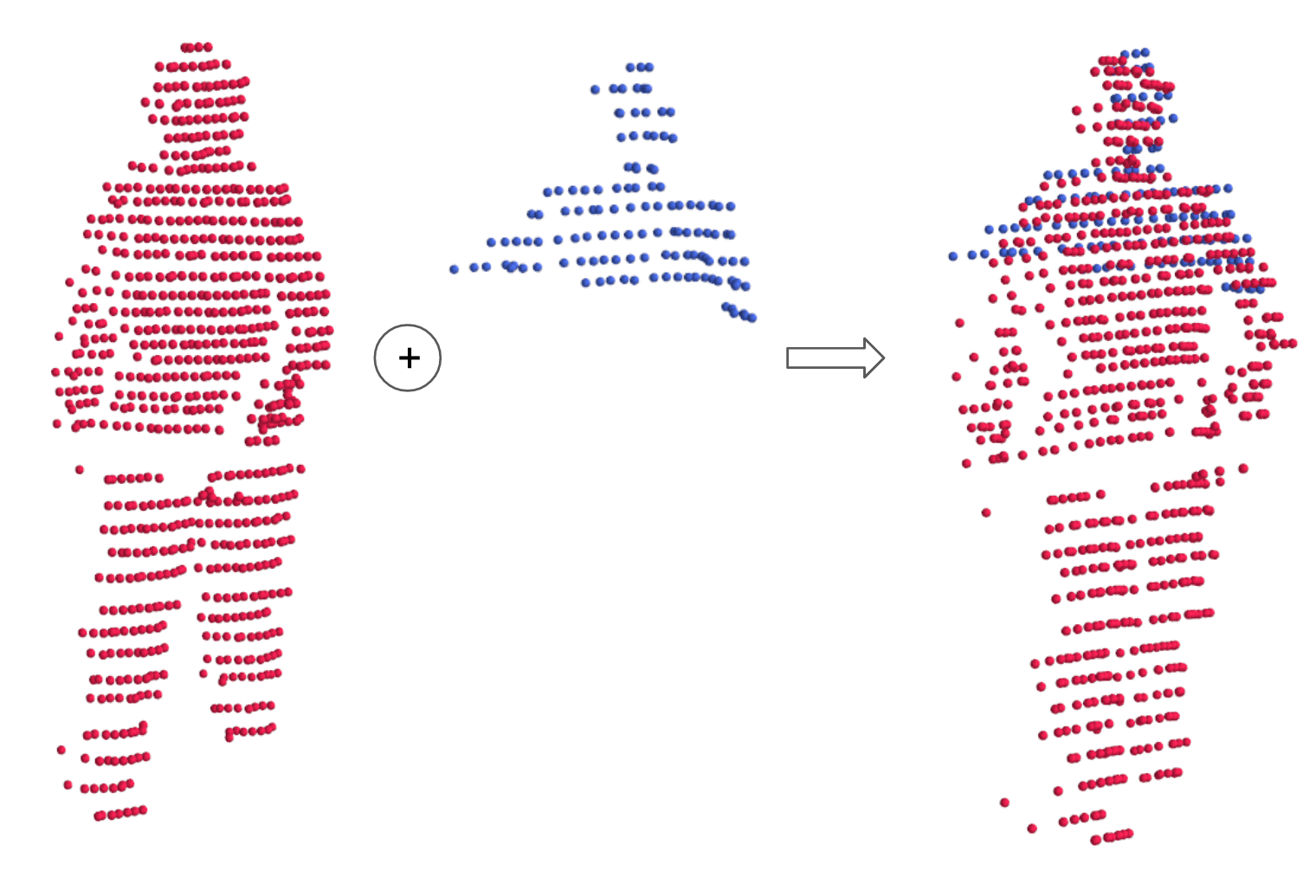}
                \end{subfigure}
                \begin{subfigure}{0.5\linewidth}
                    \flushleft
                    \includegraphics[width=0.9\linewidth]{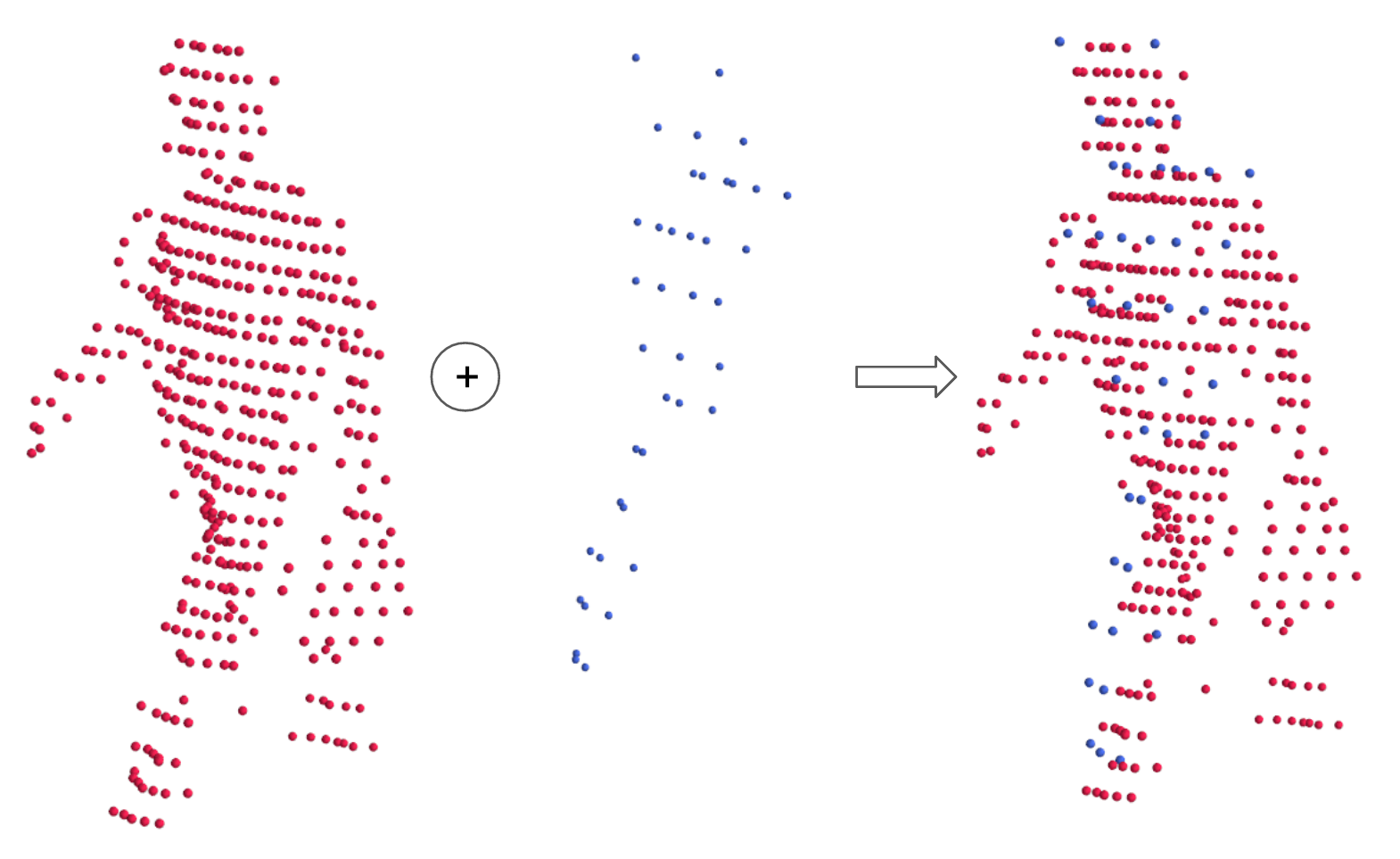}
                \end{subfigure}
                \begin{subfigure}{0.5\linewidth}
                    \flushright
                    \includegraphics[width=0.9\linewidth]{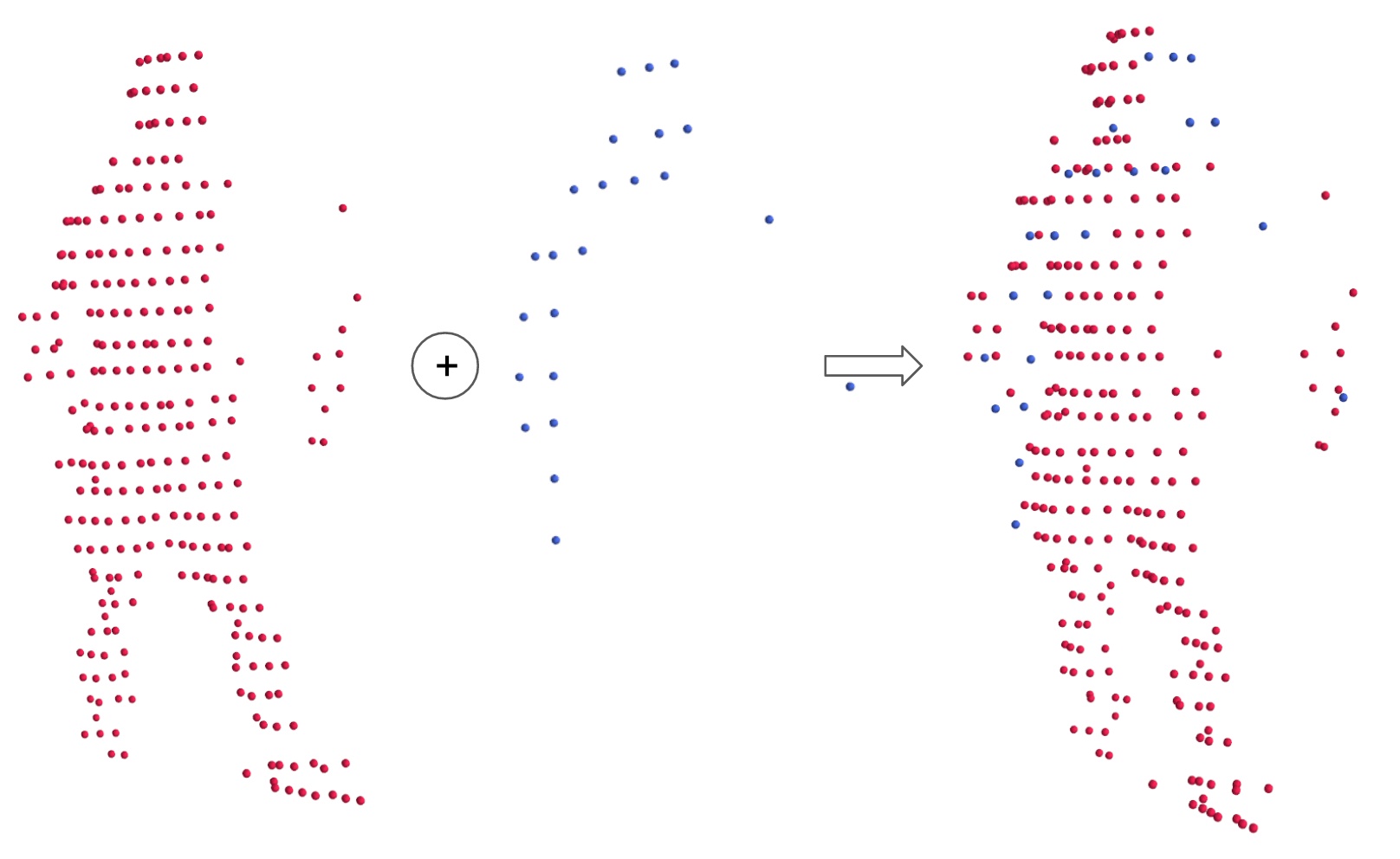}
                \end{subfigure}
                \begin{subfigure}{0.5\linewidth}
                    \flushleft
                    \includegraphics[width=0.9\linewidth]{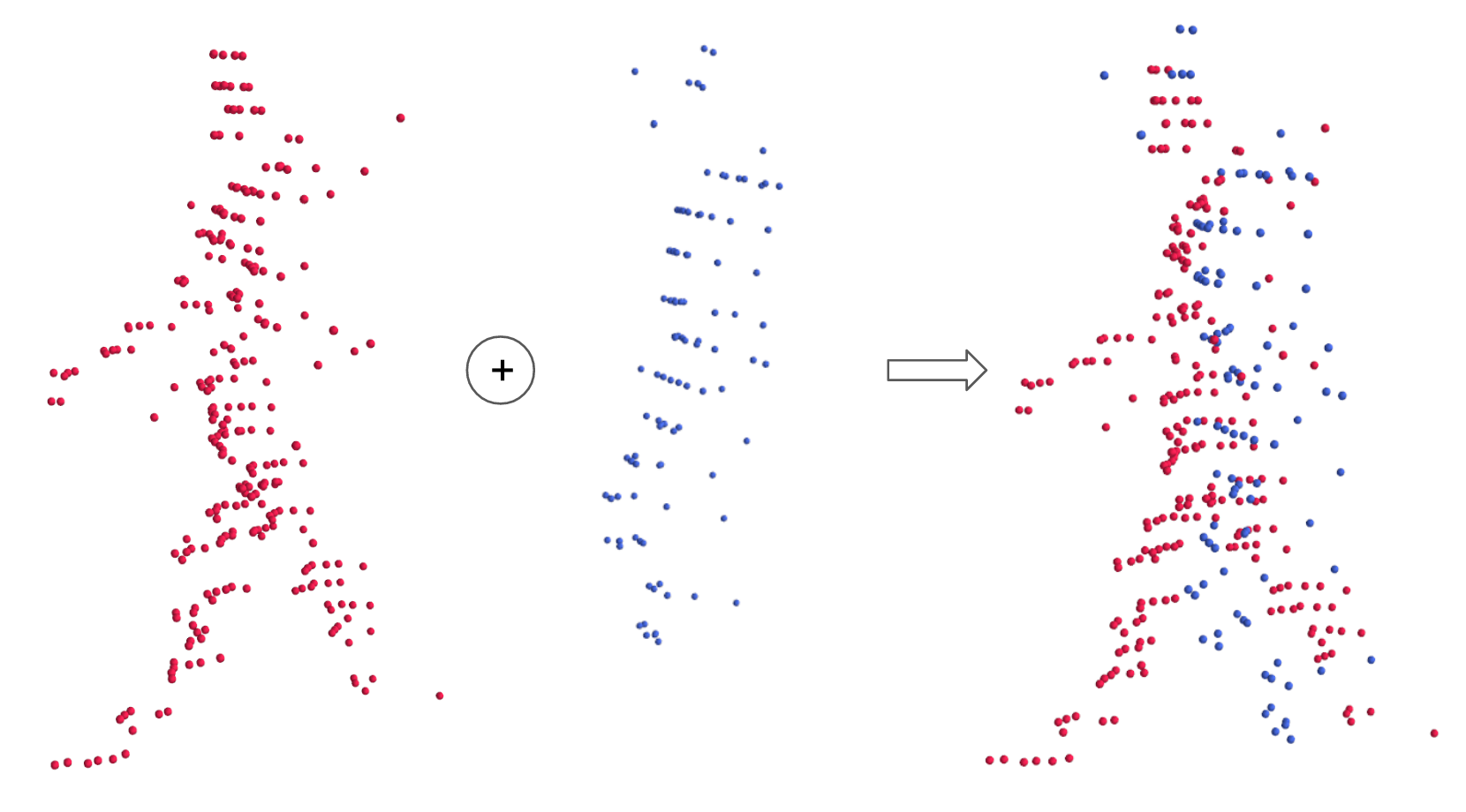}
                \end{subfigure}
                \begin{subfigure}{0.5\linewidth}
                    \flushright
                    \includegraphics[width=0.9\linewidth]{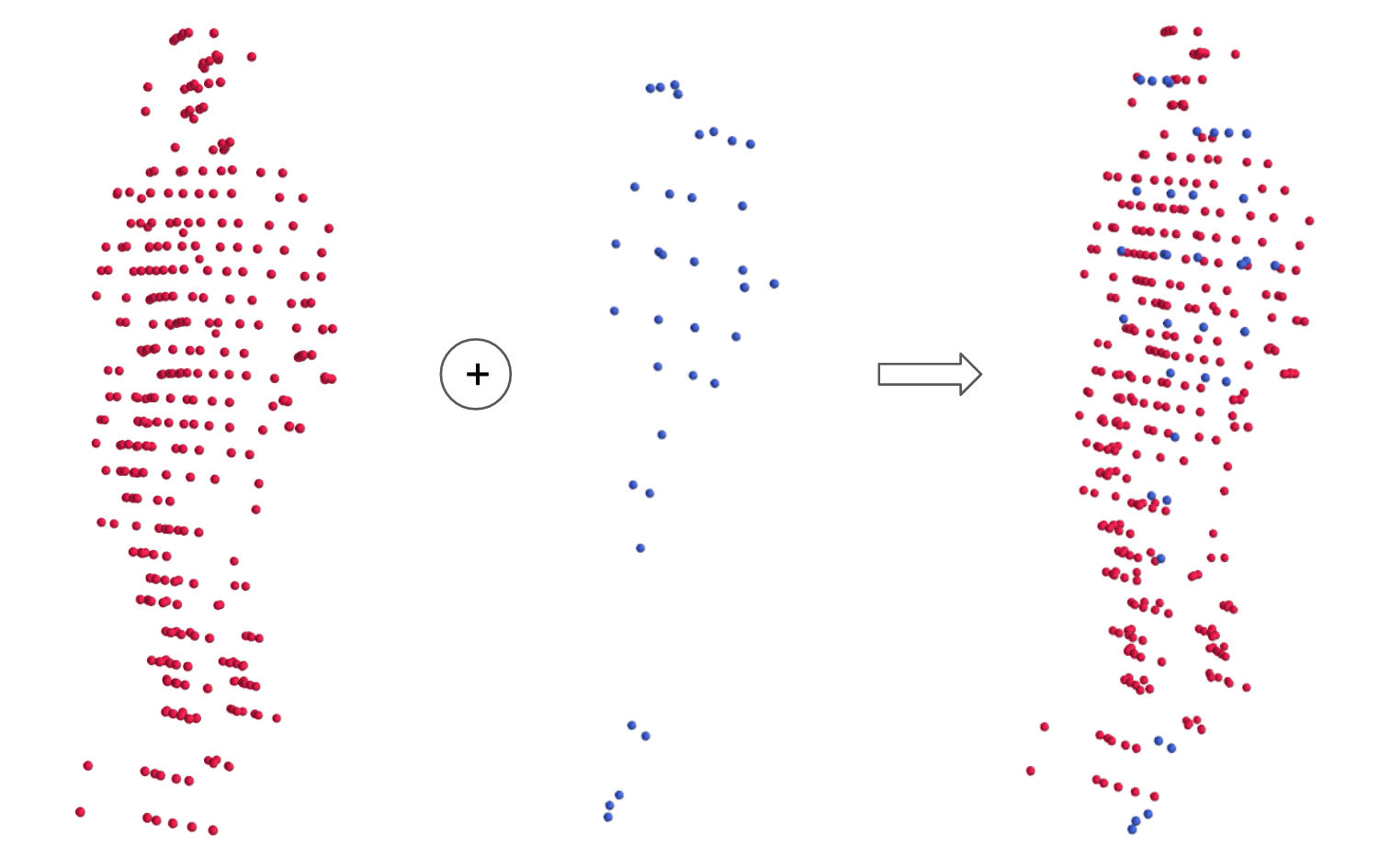}
                \end{subfigure}
                \begin{subfigure}{0.5\linewidth}
                    \flushleft
                    \includegraphics[width=0.9\linewidth]{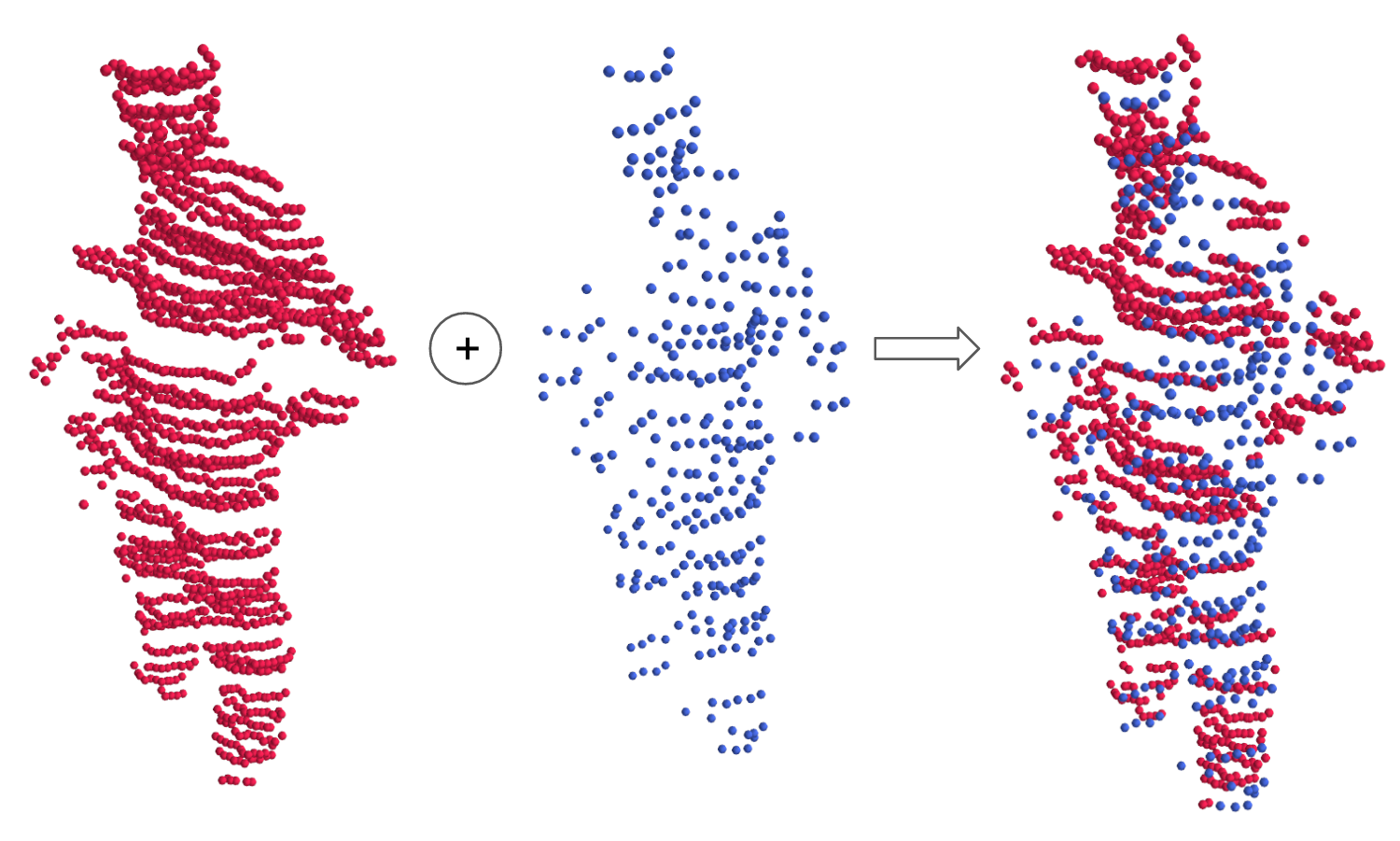}
                \end{subfigure}
            \end{adjustwidth}
            \captionsetup{aboveskip = 20pt}
            \caption{The assembly process to approximate the complete shapes for pedestrians on KITTI \cite{geiger2013vision}. The red points are from the source objects, the blue points are from target objects, the complete shape of the target objects are approximated by borrowing the points from the selected source objects (red).}
            \label{fig:ped_asb2}
        \end{figure*}

        We show more results of the assembled complete object shapes of cyclists and pedestrians in this section. Figure \ref{fig:byc_asb} visualizes the process for cyclists which includes mirroring both source and target objects. Figure \ref{fig:ped_asb1} and \ref{fig:ped_asb2} visualizes the process for pedestrians which does not mirror the objects since pedestrians are less likely to be symmetric. The blue points are the points of the target object and the red points are the points of the source objects. The assembled object faithfully covers the originally partially observed parts of the target objects and provides reasonable recovery points in the shape miss regions of the target objects.
        

    \section{Visualization of the Occupancy Probability \POS}
\begin{figure*}[!htb]
            \begin{adjustwidth}{-10pt}{-10pt}
                \begin{subfigure}{0.5\linewidth}
                    \flushleft
                    \includegraphics[width=0.95\linewidth]{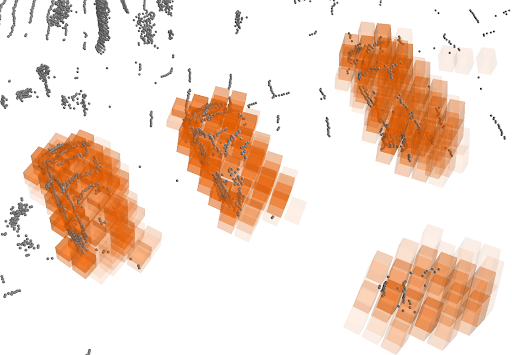}
                \end{subfigure}
                \begin{subfigure}{0.5\linewidth}
                    \flushright
                    \includegraphics[width=0.95\linewidth]{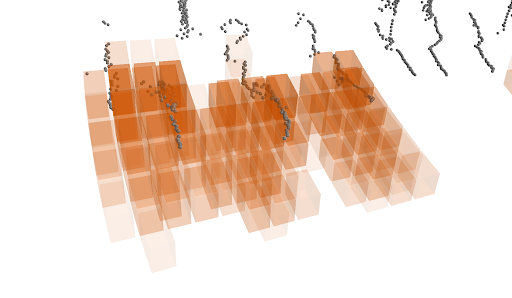}
                \end{subfigure}
                \begin{subfigure}{0.5\linewidth}
                    \flushleft
                    \includegraphics[width=0.95\linewidth]{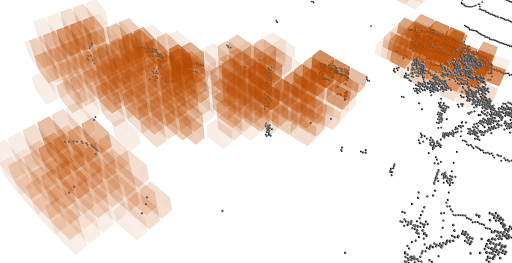}
                \end{subfigure}
                \begin{subfigure}{0.5\linewidth}
                    \flushright
                    \includegraphics[width=0.95\linewidth]{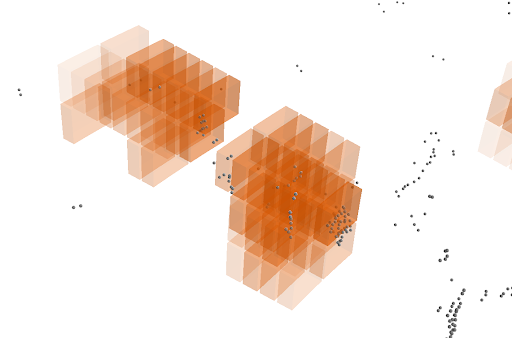}
                \end{subfigure}
                \begin{subfigure}{0.5\linewidth}
                    \flushleft
                    \includegraphics[width=0.95\linewidth]{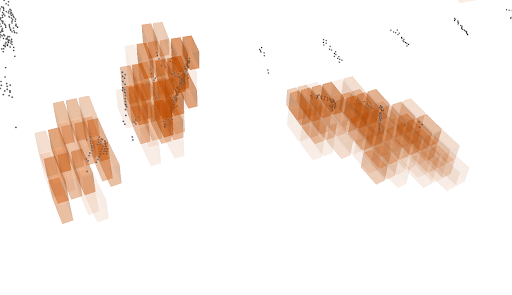}
                \end{subfigure}
                \begin{subfigure}{0.5\linewidth}
                    \flushright
                    \includegraphics[width=0.95\linewidth]{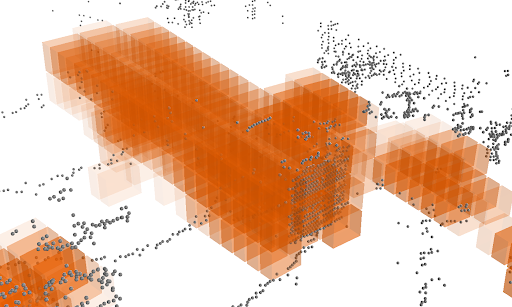}
                \end{subfigure}
                \begin{subfigure}{0.5\linewidth}
                    \flushleft
                    \includegraphics[width=0.95\linewidth]{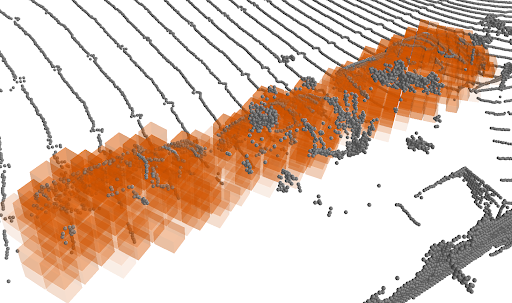}
                \end{subfigure}
                \begin{subfigure}{0.5\linewidth}
                    \flushright
                    \includegraphics[width=0.95\linewidth]{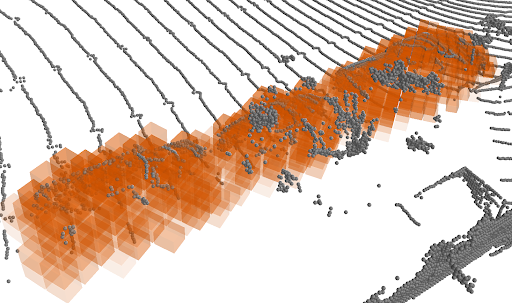}
                \end{subfigure}
            \end{adjustwidth}
            \captionsetup{aboveskip = 20pt}
            \caption{The zoomed in views of the predicted occupancy probability for vehicle objects on the Waymo Open Dataset \cite{sun2019scalability}. The higher probability it is predicted, the larger opacity we apply to the spherical voxel.}
            \label{fig:zoomvis}
        \end{figure*}

\begin{figure*}
            \begin{adjustwidth}{-0pt}{-0pt}
                \centering
                \includegraphics[width=1.0\linewidth]{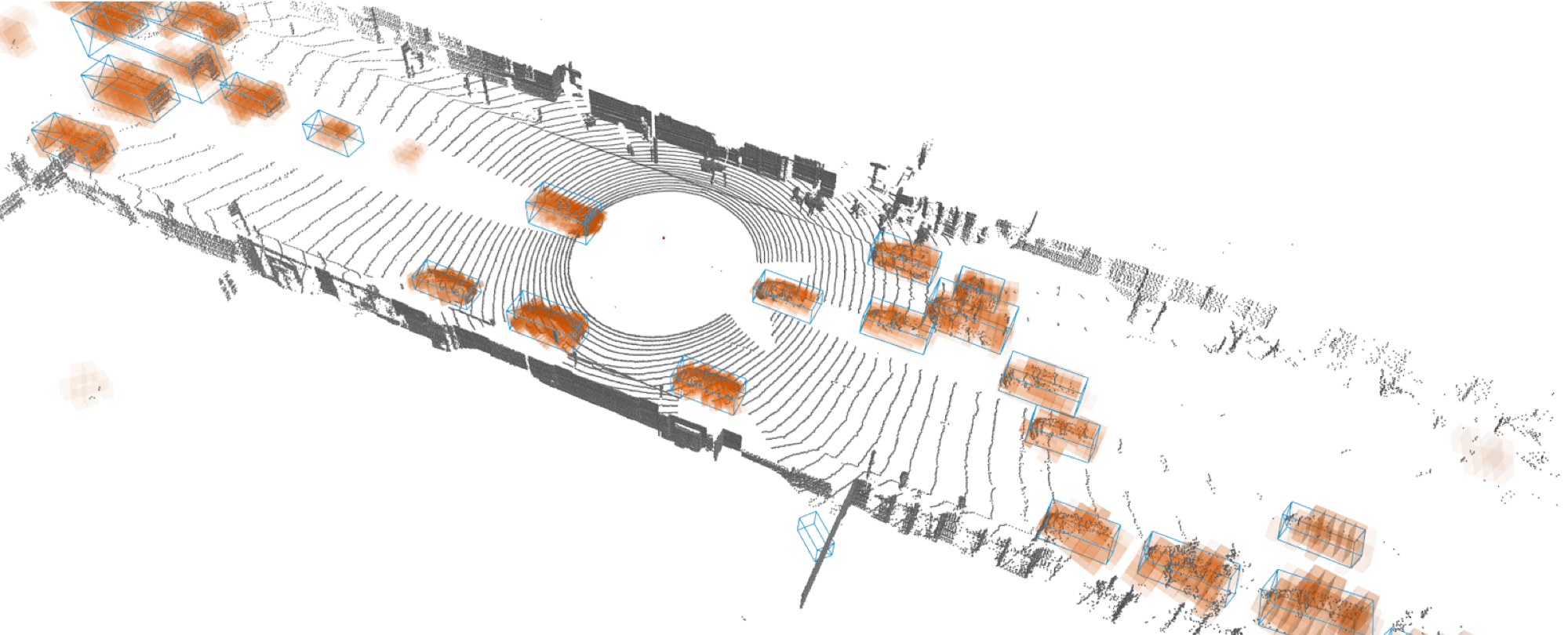}
                \captionsetup{aboveskip = 10pt}
                \captionsetup{belowskip = -10pt}
            \caption{The full scene views of the predicted occupancy probability for vehicle objects on the Waymo Open Dataset \cite{sun2019scalability}. The higher probability it is predicted, the larger opacity we apply to the spherical voxel.}
            \label{fig:scenevis}
            \end{adjustwidth}
        \end{figure*}

        We show the qualitative results of the occupancy probability for vehicle objects on the Waymo Open Dataset \cite{sun2019scalability}. Figure \ref{fig:zoomvis} contains zoomed in views of the occupancy probability while Figure \ref{fig:scenevis} contains full scene views. The higher probability one is estimated, the larger opacity we apply to the spherical voxel. 
        
        \FloatBarrier 
        \begin{figure*}[!ht] 
            \begin{adjustwidth}{-0pt}{-0pt}
                \centering
                \includegraphics[width=1.0\linewidth]{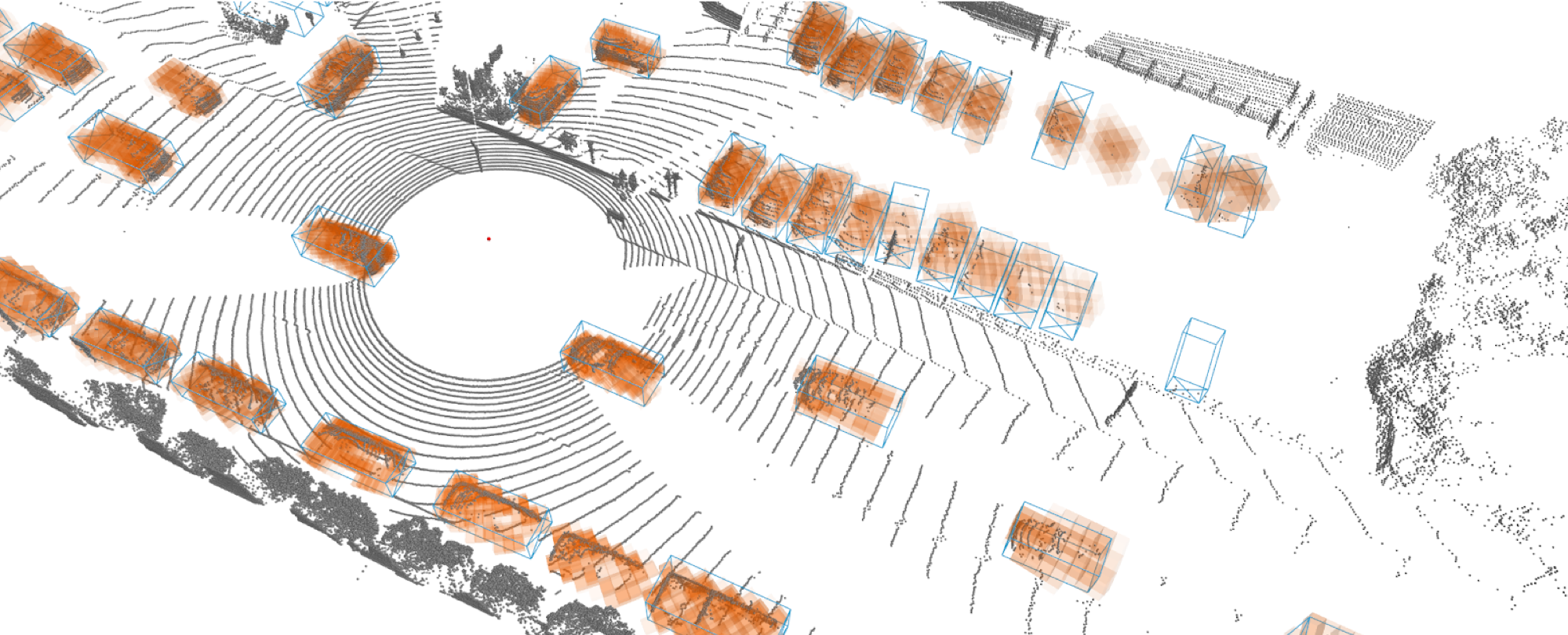}
            \end{adjustwidth}
        \vspace{30pt}
        \end{figure*}
	    \begin{figure*}
            \begin{adjustwidth}{-0pt}{-0pt}
                \centering
                \includegraphics[width=1.0\linewidth]{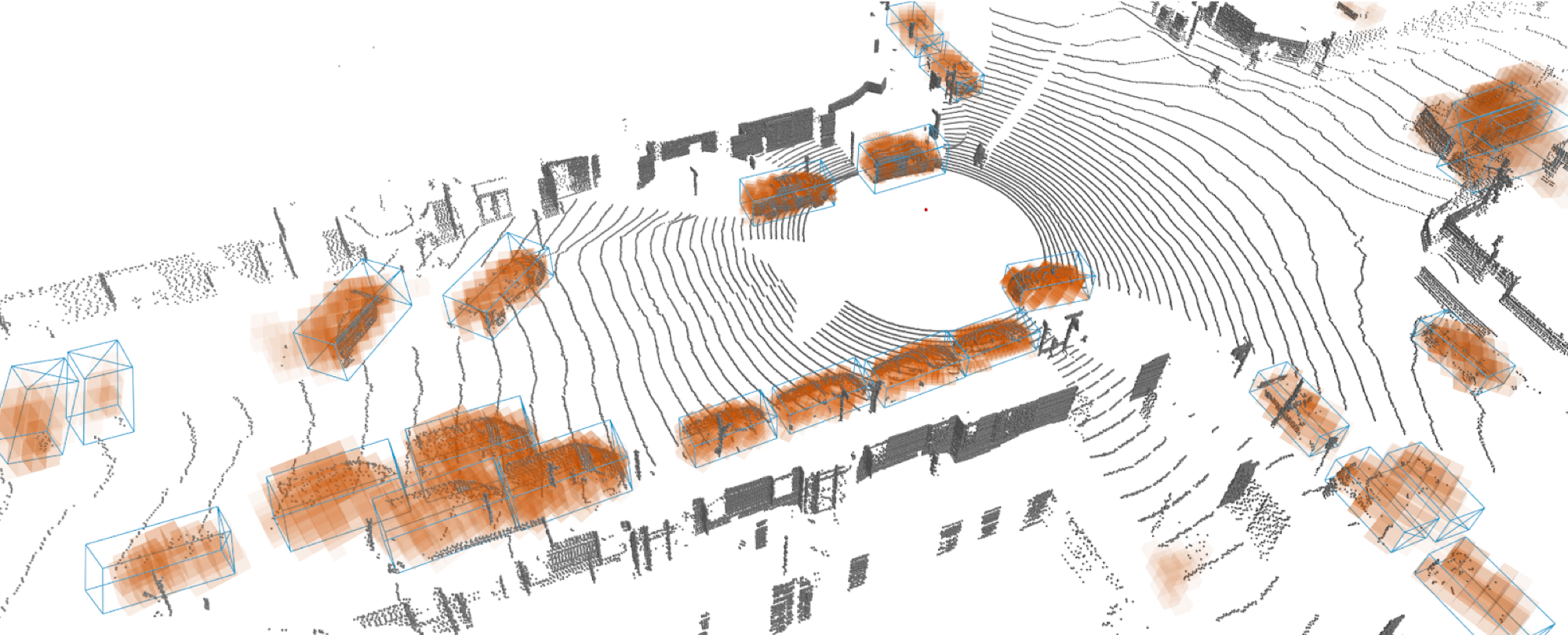}
            \end{adjustwidth}
        \vspace{30pt}
        \end{figure*}
        \begin{figure*}
            \begin{adjustwidth}{-0pt}{-0pt}
                \centering
                \includegraphics[width=1.0\linewidth]{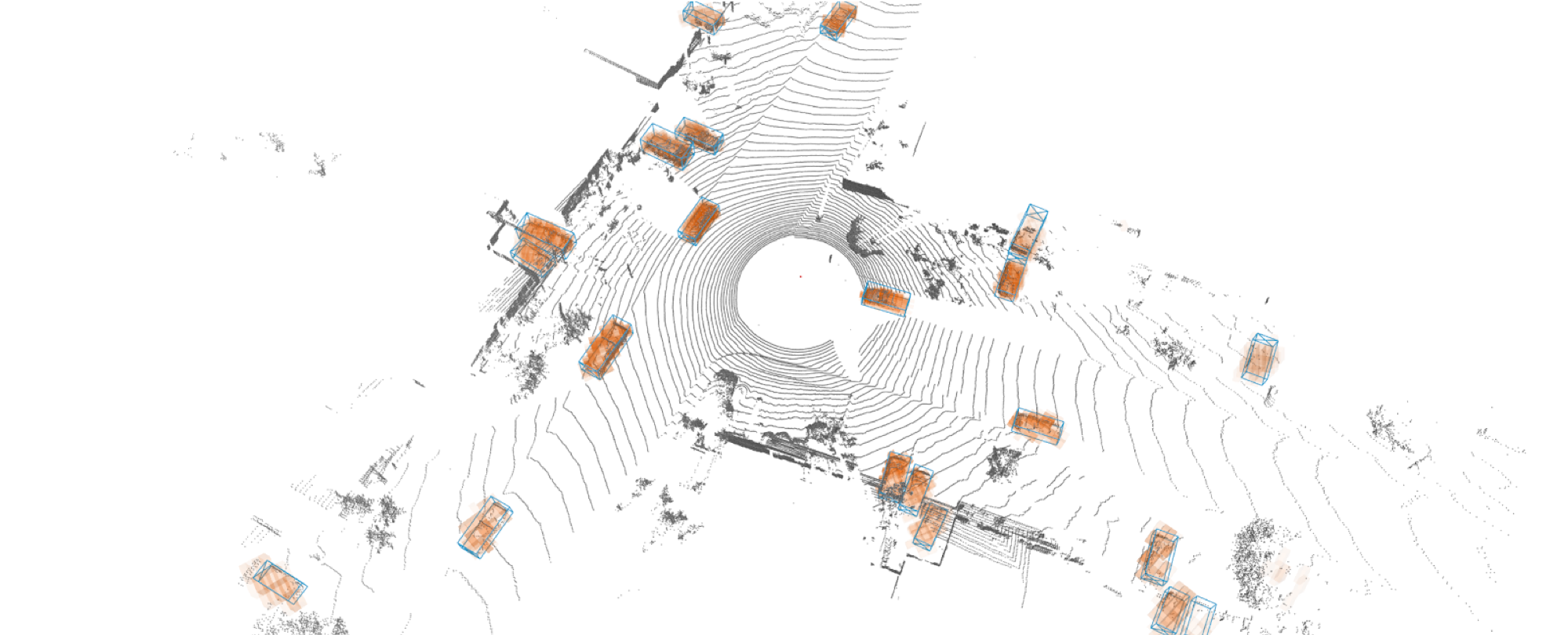}
            \end{adjustwidth}
            \vspace{30pt}
        \end{figure*}
        \begin{figure*}
            \begin{adjustwidth}{-0pt}{-0pt}
                \centering
                \includegraphics[width=1.0\linewidth]{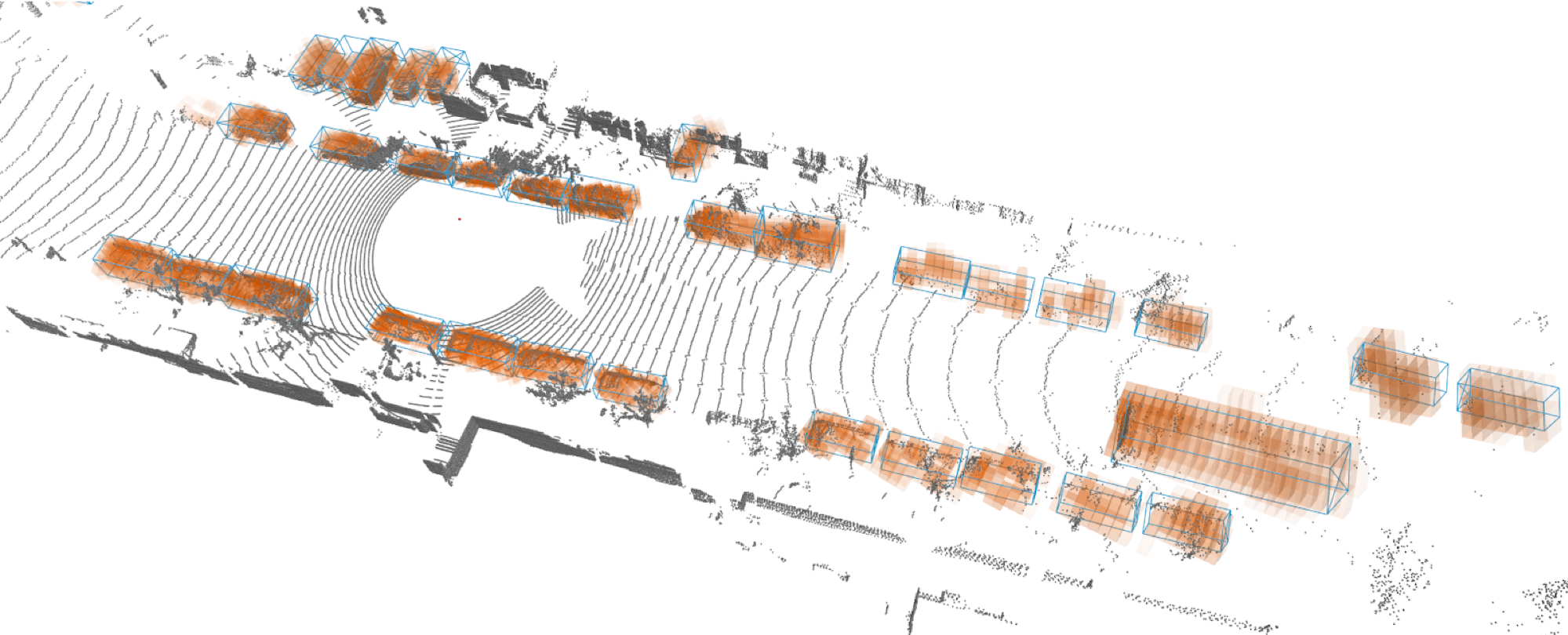}
            \end{adjustwidth}
        \end{figure*}
        \begin{figure*}
            \begin{adjustwidth}{-0pt}{-0pt}
                \centering
                \includegraphics[width=1.0\linewidth]{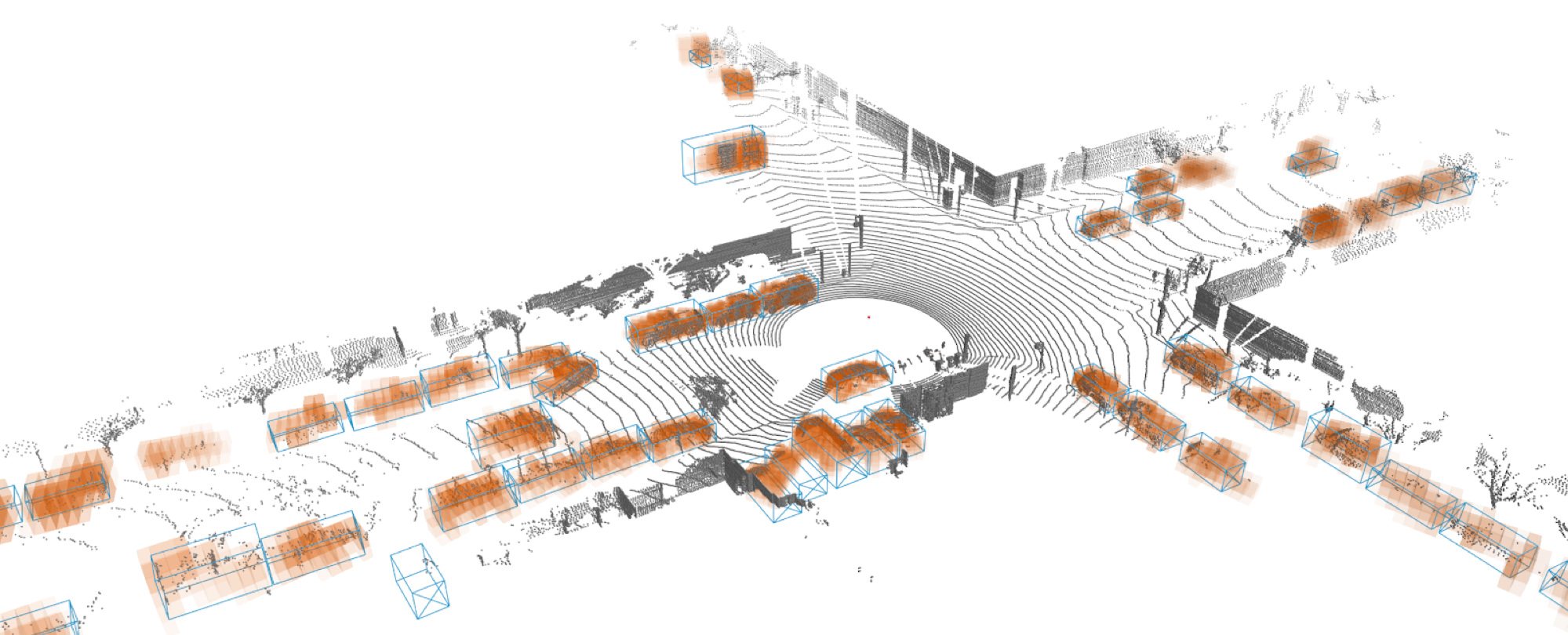}
            \end{adjustwidth}
        \end{figure*}

\end{appendices}
\FloatBarrier 

\bibliography{aaai22}

\begin{thebibliography}{55}
\providecommand{\natexlab}[1]{#1}

\bibitem[{Chen et~al.(2020)Chen, Sun, Wang, Jia, and Yuille}]{chen2020object}
Chen, Q.; Sun, L.; Wang, Z.; Jia, K.; and Yuille, A. 2020.
\newblock Object as hotspots: An anchor-free 3d object detection approach via
  firing of hotspots.
\newblock In \emph{European Conference on Computer Vision}, 68--84. Springer.

\bibitem[{Chen et~al.(2017)Chen, Ma, Wan, Li, and Xia}]{REF:Multiview3D_2017}
Chen, X.; Ma, H.; Wan, J.; Li, B.; and Xia, T. 2017.
\newblock Multi-view 3D Object Detection Network for Autonomous Driving.
\newblock \emph{2017 IEEE Conference on Computer Vision and Pattern Recognition
  (CVPR)}, 6526--6534.

\bibitem[{Chen et~al.(2019)Chen, Liu, Shen, and Jia}]{chen2019fast}
Chen, Y.; Liu, S.; Shen, X.; and Jia, J. 2019.
\newblock Fast point r-cnn.
\newblock In \emph{Proceedings of the IEEE International Conference on Computer
  Vision}, 9775--9784.

\bibitem[{Deng et~al.(2020)Deng, Shi, Li, Zhou, Zhang, and Li}]{deng2020voxel}
Deng, J.; Shi, S.; Li, P.; Zhou, W.; Zhang, Y.; and Li, H. 2020.
\newblock Voxel R-CNN: Towards High Performance Voxel-based 3D Object
  Detection.
\newblock \emph{arXiv:2012.15712}.

\bibitem[{Du et~al.(2020)Du, Ye, Tan, Feng, Xu, Ding, and
  Wen}]{du2020associate}
Du, L.; Ye, X.; Tan, X.; Feng, J.; Xu, Z.; Ding, E.; and Wen, S. 2020.
\newblock Associate-3Ddet: Perceptual-to-Conceptual Association for 3D Point
  Cloud Object Detection.
\newblock In \emph{Proceedings of the IEEE/CVF Conference on Computer Vision
  and Pattern Recognition}, 13329--13338.

\bibitem[{Follmann et~al.(2019)Follmann, K{\"o}nig, H{\"a}rtinger, Klostermann,
  and B{\"o}ttger}]{follmann2019learning}
Follmann, P.; K{\"o}nig, R.; H{\"a}rtinger, P.; Klostermann, M.; and
  B{\"o}ttger, T. 2019.
\newblock Learning to see the invisible: End-to-end trainable amodal instance
  segmentation.
\newblock In \emph{2019 IEEE Winter Conference on Applications of Computer
  Vision (WACV)}, 1328--1336. IEEE.

\bibitem[{Ge et~al.(2020)Ge, Ding, Hu, Wang, Chen, Huang, and Li}]{ge2020afdet}
Ge, R.; Ding, Z.; Hu, Y.; Wang, Y.; Chen, S.; Huang, L.; and Li, Y. 2020.
\newblock Afdet: Anchor free one stage 3d object detection.
\newblock \emph{arXiv preprint arXiv:2006.12671}.

\bibitem[{Geiger et~al.(2013)Geiger, Lenz, Stiller, and
  Urtasun}]{geiger2013vision}
Geiger, A.; Lenz, P.; Stiller, C.; and Urtasun, R. 2013.
\newblock Vision meets robotics: The kitti dataset.
\newblock \emph{The International Journal of Robotics Research}, 32(11):
  1231--1237.

\bibitem[{Graham(2015)}]{graham2015sparse}
Graham, B. 2015.
\newblock Sparse 3D convolutional neural networks.
\newblock \emph{arXiv preprint arXiv:1505.02890}.

\bibitem[{Graham and van~der Maaten(2017)}]{graham2017submanifold}
Graham, B.; and van~der Maaten, L. 2017.
\newblock Submanifold sparse convolutional networks.
\newblock \emph{arXiv preprint arXiv:1706.01307}.

\bibitem[{He et~al.(2020)He, Zeng, Huang, Hua, and Zhang}]{he2020sassd}
He, C.; Zeng, H.; Huang, J.; Hua, X.-S.; and Zhang, L. 2020.
\newblock Structure Aware Single-stage 3D Object Detection from Point Cloud.
\newblock In \emph{Proceedings of the IEEE Conference on Computer Vision and
  Pattern Recognition}.

\bibitem[{Hu et~al.(2020)Hu, Ziglar, Held, and Ramanan}]{Hu_2020_CVPR}
Hu, P.; Ziglar, J.; Held, D.; and Ramanan, D. 2020.
\newblock What You See is What You Get: Exploiting Visibility for 3D Object
  Detection.
\newblock In \emph{Proceedings of the IEEE/CVF Conference on Computer Vision
  and Pattern Recognition (CVPR)}.

\bibitem[{Huang et~al.(2020)Huang, Liu, Chen, and Bai}]{huang2020epnet}
Huang, T.; Liu, Z.; Chen, X.; and Bai, X. 2020.
\newblock Epnet: Enhancing point features with image semantics for 3d object
  detection.
\newblock In \emph{European Conference on Computer Vision}, 35--52. Springer.

\bibitem[{Jiang et~al.(2018)Jiang, Luo, Mao, Xiao, and
  Jiang}]{jiang2018acquisition}
Jiang, B.; Luo, R.; Mao, J.; Xiao, T.; and Jiang, Y. 2018.
\newblock Acquisition of localization confidence for accurate object detection.
\newblock In \emph{Proceedings of the European Conference on Computer Vision
  (ECCV)}, 784--799.

\bibitem[{Kingma and Ba(2014)}]{kingma2014adam}
Kingma, D.~P.; and Ba, J. 2014.
\newblock Adam: A method for stochastic optimization.
\newblock \emph{arXiv preprint arXiv:1412.6980}.

\bibitem[{Ku et~al.(2018)Ku, Mozifian, Lee, Harakeh, and
  Waslander}]{ku2018joint}
Ku, J.; Mozifian, M.; Lee, J.; Harakeh, A.; and Waslander, S.~L. 2018.
\newblock Joint 3d proposal generation and object detection from view
  aggregation.
\newblock In \emph{2018 IEEE/RSJ International Conference on Intelligent Robots
  and Systems (IROS)}, 1--8. IEEE.

\bibitem[{Kuang et~al.(2020)Kuang, Wang, An, Zhang, and Zhang}]{kuang2020voxel}
Kuang, H.; Wang, B.; An, J.; Zhang, M.; and Zhang, Z. 2020.
\newblock Voxel-FPN: Multi-scale voxel feature aggregation for 3D object
  detection from LIDAR point clouds.
\newblock \emph{Sensors}, 20(3): 704.

\bibitem[{Lang et~al.(2019)Lang, Vora, Caesar, Zhou, Yang, and
  Beijbom}]{lang2019pointpillars}
Lang, A.~H.; Vora, S.; Caesar, H.; Zhou, L.; Yang, J.; and Beijbom, O. 2019.
\newblock Pointpillars: Fast encoders for object detection from point clouds.
\newblock In \emph{Proceedings of the IEEE Conference on Computer Vision and
  Pattern Recognition}, 12697--12705.

\bibitem[{Li et~al.(2019)Li, Ouyang, Sheng, Zeng, and Wang}]{li2019gs3d}
Li, B.; Ouyang, W.; Sheng, L.; Zeng, X.; and Wang, X. 2019.
\newblock Gs3d: An efficient 3d object detection framework for autonomous
  driving.
\newblock In \emph{Proceedings of the IEEE Conference on Computer Vision and
  Pattern Recognition}, 1019--1028.

\bibitem[{Li et~al.(2021)Li, Yao, Quan, Yang, and Xie}]{li2021sienet}
Li, Z.; Yao, Y.; Quan, Z.; Yang, W.; and Xie, J. 2021.
\newblock SIENet: Spatial Information Enhancement Network for 3D Object
  Detection from Point Cloud.
\newblock \emph{arXiv preprint arXiv:2103.15396}.

\bibitem[{Liang et~al.(2019)Liang, Yang, Chen, Hu, and
  Urtasun}]{liang2019multi}
Liang, M.; Yang, B.; Chen, Y.; Hu, R.; and Urtasun, R. 2019.
\newblock Multi-task multi-sensor fusion for 3d object detection.
\newblock In \emph{Proceedings of the IEEE Conference on Computer Vision and
  Pattern Recognition}, 7345--7353.

\bibitem[{Lin et~al.(2017)Lin, Goyal, Girshick, He, and
  Doll{\'a}r}]{lin2017focal}
Lin, T.-Y.; Goyal, P.; Girshick, R.; He, K.; and Doll{\'a}r, P. 2017.
\newblock Focal loss for dense object detection.
\newblock In \emph{Proceedings of the IEEE international conference on computer
  vision}, 2980--2988.

\bibitem[{Liu et~al.(2018)Liu, Jing, Nie, Gao, Liu, and Jiang}]{liu2018context}
Liu, Y.; Jing, X.-Y.; Nie, J.; Gao, H.; Liu, J.; and Jiang, G.-P. 2018.
\newblock Context-aware three-dimensional mean-shift with occlusion handling
  for robust object tracking in RGB-D videos.
\newblock \emph{IEEE Transactions on Multimedia}, 21(3): 664--677.

\bibitem[{Liu et~al.(2020)Liu, Zhao, Huang, Hu, Zhou, and Bai}]{liu2020tanet}
Liu, Z.; Zhao, X.; Huang, T.; Hu, R.; Zhou, Y.; and Bai, X. 2020.
\newblock Tanet: Robust 3d object detection from point clouds with triple
  attention.
\newblock In \emph{Proceedings of the AAAI Conference on Artificial
  Intelligence}, volume~34, 11677--11684.

\bibitem[{Najibi et~al.(2020)Najibi, Lai, Kundu, Lu, Rathod, Funkhouser,
  Pantofaru, Ross, Davis, and Fathi}]{najibi2020dops}
Najibi, M.; Lai, G.; Kundu, A.; Lu, Z.; Rathod, V.; Funkhouser, T.; Pantofaru,
  C.; Ross, D.; Davis, L.~S.; and Fathi, A. 2020.
\newblock Dops: Learning to detect 3d objects and predict their 3d shapes.
\newblock In \emph{Proceedings of the IEEE/CVF conference on computer vision
  and pattern recognition}, 11913--11922.

\bibitem[{Pan et~al.(2021)Pan, Xiao, He, Shao, and Li}]{pan2021mulls}
Pan, Y.; Xiao, P.; He, Y.; Shao, Z.; and Li, Z. 2021.
\newblock MULLS: Versatile LiDAR SLAM via Multi-metric Linear Least Square.
\newblock \emph{arXiv preprint arXiv:2102.03771}.

\bibitem[{Pang et~al.(2020)Pang, Morris, and Radha}]{pang2020clocs}
Pang, S.; Morris, D.; and Radha, H. 2020.
\newblock CLOCs: Camera-LiDAR Object Candidates Fusion for 3D Object Detection.
\newblock \emph{arXiv preprint arXiv:2009.00784}.

\bibitem[{Qi et~al.(2019{\natexlab{a}})Qi, Litany, He, and Guibas}]{qi2019deep}
Qi, C.~R.; Litany, O.; He, K.; and Guibas, L.~J. 2019{\natexlab{a}}.
\newblock Deep hough voting for 3d object detection in point clouds.
\newblock In \emph{Proceedings of the IEEE/CVF International Conference on
  Computer Vision}, 9277--9286.

\bibitem[{Qi et~al.(2018)Qi, Liu, Wu, Su, and Guibas}]{qi2018frustum}
Qi, C.~R.; Liu, W.; Wu, C.; Su, H.; and Guibas, L.~J. 2018.
\newblock Frustum pointnets for 3d object detection from rgb-d data.
\newblock In \emph{Proceedings of the IEEE conference on computer vision and
  pattern recognition}, 918--927.

\bibitem[{Qi et~al.(2019{\natexlab{b}})Qi, Jiang, Liu, Shen, and
  Jia}]{qi2019amodal}
Qi, L.; Jiang, L.; Liu, S.; Shen, X.; and Jia, J. 2019{\natexlab{b}}.
\newblock Amodal instance segmentation with kins dataset.
\newblock In \emph{Proceedings of the IEEE/CVF Conference on Computer Vision
  and Pattern Recognition}, 3014--3023.

\bibitem[{Reddy et~al.(2019)Reddy, Vo, and Narasimhan}]{reddy2019occlusion}
Reddy, N.~D.; Vo, M.; and Narasimhan, S.~G. 2019.
\newblock Occlusion-net: 2d/3d occluded keypoint localization using graph
  networks.
\newblock In \emph{Proceedings of the IEEE/CVF Conference on Computer Vision
  and Pattern Recognition}, 7326--7335.

\bibitem[{Saleh et~al.(2021)Saleh, Sz{\'e}n{\'a}si, and
  V{\'a}mossy}]{saleh2021occlusion}
Saleh, K.; Sz{\'e}n{\'a}si, S.; and V{\'a}mossy, Z. 2021.
\newblock Occlusion Handling in Generic Object Detection: A Review.
\newblock In \emph{2021 IEEE 19th World Symposium on Applied Machine
  Intelligence and Informatics (SAMI)}, 000477--000484. IEEE.

\bibitem[{Shi et~al.(2020)Shi, Guo, Jiang, Wang, Shi, Wang, and Li}]{shi2020pv}
Shi, S.; Guo, C.; Jiang, L.; Wang, Z.; Shi, J.; Wang, X.; and Li, H. 2020.
\newblock Pv-rcnn: Point-voxel feature set abstraction for 3d object detection.
\newblock In \emph{Proceedings of the IEEE/CVF Conference on Computer Vision
  and Pattern Recognition}, 10529--10538.

\bibitem[{Shi et~al.(2019{\natexlab{a}})Shi, Wang, and Li}]{shi2019pointrcnn}
Shi, S.; Wang, X.; and Li, H. 2019{\natexlab{a}}.
\newblock Pointrcnn: 3d object proposal generation and detection from point
  cloud.
\newblock In \emph{Proceedings of the IEEE Conference on Computer Vision and
  Pattern Recognition}, 770--779.

\bibitem[{Shi et~al.(2019{\natexlab{b}})Shi, Wang, Shi, Wang, and
  Li}]{shi2019points}
Shi, S.; Wang, Z.; Shi, J.; Wang, X.; and Li, H. 2019{\natexlab{b}}.
\newblock From Points to Parts: 3D Object Detection from Point Cloud with
  Part-aware and Part-aggregation Network.
\newblock \emph{arXiv preprint arXiv:1907.03670}.

\bibitem[{{Shi} et~al.(2020){Shi}, {Wang}, {Shi}, {Wang}, and {Li}}]{9018080}
{Shi}, S.; {Wang}, Z.; {Shi}, J.; {Wang}, X.; and {Li}, H. 2020.
\newblock From Points to Parts: 3D Object Detection from Point Cloud with
  Part-aware and Part-aggregation Network.
\newblock \emph{IEEE Transactions on Pattern Analysis and Machine
  Intelligence}, 1--1.

\bibitem[{Shi and Rajkumar(2020)}]{shi2020point}
Shi, W.; and Rajkumar, R. 2020.
\newblock Point-gnn: Graph neural network for 3d object detection in a point
  cloud.
\newblock In \emph{Proceedings of the IEEE/CVF Conference on Computer Vision
  and Pattern Recognition}, 1711--1719.

\bibitem[{Sun et~al.(2019)Sun, Kretzschmar, Dotiwalla, Chouard, Patnaik, Tsui,
  Guo, Zhou, Chai, Caine, Vasudevan, Han, Ngiam, Zhao, Timofeev, Ettinger,
  Krivokon, Gao, Joshi, Zhang, Shlens, Chen, and Anguelov}]{sun2019scalability}
Sun, P.; Kretzschmar, H.; Dotiwalla, X.; Chouard, A.; Patnaik, V.; Tsui, P.;
  Guo, J.; Zhou, Y.; Chai, Y.; Caine, B.; Vasudevan, V.; Han, W.; Ngiam, J.;
  Zhao, H.; Timofeev, A.; Ettinger, S.; Krivokon, M.; Gao, A.; Joshi, A.;
  Zhang, Y.; Shlens, J.; Chen, Z.; and Anguelov, D. 2019.
\newblock Scalability in Perception for Autonomous Driving: Waymo Open Dataset.
\newblock arXiv:1912.04838.

\bibitem[{Wang et~al.(2020)Wang, Fathi, Kundu, Ross, Pantofaru, Funkhouser, and
  Solomon}]{wang2020pillar}
Wang, Y.; Fathi, A.; Kundu, A.; Ross, D.; Pantofaru, C.; Funkhouser, T.; and
  Solomon, J. 2020.
\newblock Pillar-based object detection for autonomous driving.
\newblock \emph{arXiv preprint arXiv:2007.10323}.

\bibitem[{Wang and Jia(2019)}]{wang2019frustum}
Wang, Z.; and Jia, K. 2019.
\newblock Frustum ConvNet: Sliding Frustums to Aggregate Local Point-Wise
  Features for Amodal 3D Object Detection.
\newblock In \emph{2019 IEEE/RSJ International Conference on Intelligent Robots
  and Systems (IROS)}, 1742--1749. IEEE.

\bibitem[{Xu et~al.(2020)Xu, Sun, Wu, Wang, and Neumann}]{xu2020grid}
Xu, Q.; Sun, X.; Wu, C.-Y.; Wang, P.; and Neumann, U. 2020.
\newblock Grid-gcn for fast and scalable point cloud learning.
\newblock In \emph{Proceedings of the IEEE/CVF Conference on Computer Vision
  and Pattern Recognition}, 5661--5670.

\bibitem[{Xu et~al.(2021)Xu, Zhou, Wang, Qi, and Anguelov}]{xu2021spg}
Xu, Q.; Zhou, Y.; Wang, W.; Qi, C.~R.; and Anguelov, D. 2021.
\newblock Spg: Unsupervised domain adaptation for 3d object detection via
  semantic point generation.
\newblock In \emph{Proceedings of the IEEE/CVF International Conference on
  Computer Vision}, 15446--15456.

\bibitem[{Yan et~al.(2020)Yan, Gao, Li, Zhang, Li, Huang, and
  Cui}]{yan2020sparse}
Yan, X.; Gao, J.; Li, J.; Zhang, R.; Li, Z.; Huang, R.; and Cui, S. 2020.
\newblock Sparse single sweep lidar point cloud segmentation via learning
  contextual shape priors from scene completion.
\newblock \emph{arXiv preprint arXiv:2012.03762}.

\bibitem[{Yan et~al.(2018)Yan, Mao, and Li}]{yan2018second}
Yan, Y.; Mao, Y.; and Li, B. 2018.
\newblock Second: Sparsely embedded convolutional detection.
\newblock \emph{Sensors}, 18(10): 3337.

\bibitem[{Yang et~al.(2020)Yang, Sun, Liu, and Jia}]{yang20203dssd}
Yang, Z.; Sun, Y.; Liu, S.; and Jia, J. 2020.
\newblock 3dssd: Point-based 3d single stage object detector.
\newblock In \emph{Proceedings of the IEEE/CVF Conference on Computer Vision
  and Pattern Recognition}, 11040--11048.

\bibitem[{Yang et~al.(2019)Yang, Sun, Liu, Shen, and Jia}]{yang2019std}
Yang, Z.; Sun, Y.; Liu, S.; Shen, X.; and Jia, J. 2019.
\newblock Std: Sparse-to-dense 3d object detector for point cloud.
\newblock In \emph{Proceedings of the IEEE International Conference on Computer
  Vision}, 1951--1960.

\bibitem[{Ye et~al.(2020)Ye, Xu, and Cao}]{ye2020hvnet}
Ye, M.; Xu, S.; and Cao, T. 2020.
\newblock Hvnet: Hybrid voxel network for lidar based 3d object detection.
\newblock In \emph{Proceedings of the IEEE/CVF conference on computer vision
  and pattern recognition}, 1631--1640.

\bibitem[{Yi et~al.(2020)Yi, Shi, Ding, Sun, Xu, Zhou, Wang, Li, and
  Wang}]{yi2020segvoxelnet}
Yi, H.; Shi, S.; Ding, M.; Sun, J.; Xu, K.; Zhou, H.; Wang, Z.; Li, S.; and
  Wang, G. 2020.
\newblock Segvoxelnet: Exploring semantic context and depth-aware features for
  3d vehicle detection from point cloud.
\newblock In \emph{2020 IEEE International Conference on Robotics and
  Automation (ICRA)}, 2274--2280. IEEE.

\bibitem[{Yoo et~al.(2020)Yoo, Kim, Kim, and Choi}]{yoo20203d}
Yoo, J.~H.; Kim, Y.; Kim, J.~S.; and Choi, J.~W. 2020.
\newblock 3d-cvf: Generating joint camera and lidar features using cross-view
  spatial feature fusion for 3d object detection.
\newblock \emph{arXiv preprint arXiv:2004.12636}, 3.

\bibitem[{Zhang et~al.(2018)Zhang, Wen, Bian, Lei, and Li}]{zhang2018occlusion}
Zhang, S.; Wen, L.; Bian, X.; Lei, Z.; and Li, S.~Z. 2018.
\newblock Occlusion-aware R-CNN: detecting pedestrians in a crowd.
\newblock In \emph{Proceedings of the European Conference on Computer Vision
  (ECCV)}, 637--653.

\bibitem[{Zheng et~al.(2021)Zheng, Tang, Chen, Jiang, and Fu}]{zheng2020ciassd}
Zheng, W.; Tang, W.; Chen, S.; Jiang, L.; and Fu, C.-W. 2021.
\newblock CIA-SSD: Confident IoU-Aware Single-Stage Object Detector From Point
  Cloud.
\newblock In \emph{AAAI}.

\bibitem[{Zhou et~al.(2019)Zhou, Fang, Song, Guan, Yin, Dai, and
  Yang}]{zhou2019iou}
Zhou, D.; Fang, J.; Song, X.; Guan, C.; Yin, J.; Dai, Y.; and Yang, R. 2019.
\newblock Iou loss for 2d/3d object detection.
\newblock In \emph{2019 International Conference on 3D Vision (3DV)}, 85--94.
  IEEE.

\bibitem[{Zhou et~al.(2020{\natexlab{a}})Zhou, Fang, Song, Liu, Yin, Dai, Li,
  and Yang}]{Zhou_2020_CVPR}
Zhou, D.; Fang, J.; Song, X.; Liu, L.; Yin, J.; Dai, Y.; Li, H.; and Yang, R.
  2020{\natexlab{a}}.
\newblock Joint 3D Instance Segmentation and Object Detection for Autonomous
  Driving.
\newblock In \emph{Proceedings of the IEEE/CVF Conference on Computer Vision
  and Pattern Recognition (CVPR)}.

\bibitem[{Zhou et~al.(2020{\natexlab{b}})Zhou, Sun, Zhang, Anguelov, Gao,
  Ouyang, Guo, Ngiam, and Vasudevan}]{zhou2020end}
Zhou, Y.; Sun, P.; Zhang, Y.; Anguelov, D.; Gao, J.; Ouyang, T.; Guo, J.;
  Ngiam, J.; and Vasudevan, V. 2020{\natexlab{b}}.
\newblock End-to-end multi-view fusion for 3d object detection in lidar point
  clouds.
\newblock In \emph{Conference on Robot Learning}, 923--932.

\bibitem[{Zhou and Tuzel(2018)}]{zhou2018voxelnet}
Zhou, Y.; and Tuzel, O. 2018.
\newblock Voxelnet: End-to-end learning for point cloud based 3d object
  detection.
\newblock In \emph{Proceedings of the IEEE Conference on Computer Vision and
  Pattern Recognition}, 4490--4499.

\end{thebibliography}
\end{document}